\documentclass{article}
\usepackage[preprint]{tmlr}

\usepackage[utf8]{inputenc}
\usepackage[english]{babel}

\usepackage{amsmath,amssymb,amsthm}
\usepackage{mathtools}

\usepackage{booktabs}
\usepackage{array}
\usepackage{multirow}
\usepackage{tabularx}

\usepackage{graphicx}
\graphicspath{{figures/submission/}}
\usepackage{caption}
\usepackage{subcaption}
\usepackage{float}

\usepackage{xcolor}

\usepackage{tikz}
\usetikzlibrary{shapes.geometric, shapes.symbols, arrows.meta, positioning, shadows, calc, fit, backgrounds, decorations.markings}

\usepackage{algorithm}
\usepackage{algpseudocode}
\algrenewcommand\algorithmicrequire{\textbf{Input:}}
\algrenewcommand\algorithmicensure{\textbf{Output:}}
\algnewcommand{\LineComment}[1]{\State \(\triangleright\) \textit{#1}}

\usepackage{listings}
\lstdefinestyle{pseudocode}{
  basicstyle=\small\ttfamily,
  frame=none,
  numbers=left,
  numberstyle=\tiny\color{gray},
  numbersep=6pt,
  xleftmargin=2em,
  xrightmargin=0.5em,
  backgroundcolor=,
  tabsize=4,
  morekeywords={function,end,if,then,else,for,do,while,return,Input,Output,Source,Constants,Global,State,true,false,and,or,not,in,is},
  keywordstyle=\bfseries,
  commentstyle=\itshape\color{gray!70!black},
  morecomment=[l]{//},
  literate={<-}{{$\gets$}}2
           {->}{{$\rightarrow$}}2
           {>=}{{$\geq$}}1
           {<=}{{$\leq$}}1
           {!=}{{$\neq$}}1,
}

\usepackage[colorlinks=true,linkcolor=blue!60!black,citecolor=blue!60!black,urlcolor=blue!60!black]{hyperref}

\usepackage{enumitem}
\usepackage{adjustbox}
\usepackage{microtype}
\usepackage{xspace}

\title{PRECEPT: Planning Resilience via Experience, Context Engineering \& Probing Trajectories\\[0.5em]
\large A Unified Framework for Test-Time Adaptation with Compositional Rule Learning and Pareto-Guided Prompt Evolution}

\author{Arash Shahmansoori}

\date{}

\begin{document}

\maketitle

\begin{abstract}

LLM agents that store knowledge as natural language suffer steep retrieval degradation as condition count grows, often struggle to compose learned rules reliably, and typically lack explicit mechanisms to detect stale or adversarial knowledge.  We introduce \textbf{PRECEPT}, a unified framework for test-time adaptation with three tightly coupled components: \textbf{(1)} deterministic exact-match rule retrieval over structured condition keys, \textbf{(2)} conflict-aware memory with Bayesian source reliability and threshold-based rule invalidation, and \textbf{(3)} \textbf{COMPASS}, a Pareto-guided prompt-evolution outer loop.  Exact retrieval eliminates partial-match interpretation errors on the deterministic path (0\% by construction, vs 94.4\% under Theorem~B.6's independence model at $N{=}10$) and supports compositional stacking through a semantic tier hierarchy; conflict-aware memory resolves static--dynamic disagreements and supports drift adaptation; COMPASS evaluates prompts through the same end-to-end execution pipeline.

\textbf{Results} (9--10 seeds): PRECEPT achieves a +41.1pp first-try advantage over Full Reflexion ($d{>}1.9$), +33.3pp compositional generalization ($d{=}1.55$), 100\% $P_1$ on 2-way logistics compositions ($d{=}2.64$), +40--55pp continuous learning gains, strong eventual robustness under adversarial static knowledge (100\% logistics with adversarial SK active; partial recovery on integration), +55.0pp drift recovery ($d{=}0.95$, $p{=}0.031$), and 61\% fewer steps.  Core comparisons are statistically significant, often at $p{<}0.001$.

\end{abstract}

\section{Introduction}

Deploying LLM agents in real-world applications demands learning from limited examples, composing atomic rules into complex policies, adapting to environmental changes, and making deterministic decisions without interpretation errors.  In our evaluated setting, current approaches do not jointly address all four axes.

\textbf{Verbal reflection} \citep{shinn2023,zhao2023} stores knowledge as natural language, requiring LLM interpretation at retrieval time.  Under the independence model of Theorem~B.6, this interpretation degrades exponentially with condition count (94.4\% partial-match error at $N{=}10$).  \textbf{Reinforcement learning} suffers from prohibitive sample complexity ($\beta{=}100{+}$), cannot adapt without retraining, and offers no compositional generalization.  The \textbf{Digital Red Queen (DRQ)} framework \citep{kumar2026} reveals a deeper issue: agents trained via static optimization fail 72\% against novel dynamics, motivating a shift to \textit{survivable systems} with growing adversarial histories.

Four fundamental limitations affect all prior approaches: \textbf{(L1)~Compositional explosion}---$2^N$ condition combinations require $O(2^N)$ training; \textbf{(L2)~Interpretation degradation}---verbal retrieval accuracy degrades exponentially with $N$; \textbf{(L3)~Drift blindness}---stale rules persist indefinitely, and RL requires complete retraining; \textbf{(L4)~Sample inefficiency}---RL needs $\beta{=}100{+}$, verbal methods $\beta{=}5{+}$, while deployment demands $\beta{\leq}3$.

\textbf{Our Solution.}  PRECEPT addresses all four limitations through three tightly coupled ideas.  \textbf{Deterministic retrieval} ($O(1)$ exact-match via structured condition keys) removes interpretation from matched-rule application and enables compositional generalization via a semantic tier hierarchy (L1, L2).  \textbf{Evo-Memory with Bayesian conflict resolution}---analogous to DRQ's opponent history---detects and overrides adversarial static knowledge (Type~I) and environmental drift (Type~II), with a $64\times$ model-based resilience bound under Corollary~B.5 (L3).  \textbf{COMPASS} optimizes prompts through the full pipeline with Pareto selection and MAP-Elites diversity.  These layers are interdependent: removing any one degrades the others.

\subsection{Contributions}

PRECEPT is a \textit{unified} framework rather than a bundle of independent tricks: deterministic retrieval enables compositional rule use, structured rule memory enables conflict resolution, and conflict resolution enables reliable drift adaptation.  We therefore state the contributions as three tightly coupled capabilities plus one validation summary:

\textbf{Contribution 1: Compositional rule learning with deterministic retrieval} (Sections 2--3).
PRECEPT introduces $O(1)$ exact-match retrieval over structured condition keys, eliminating interpretation errors on the exact-match path.  This deterministic substrate supports atomic constraint stacking through a semantic tier hierarchy, yielding conditional $2^N{-}1$ compositional coverage from $N$ learned atomic rules (Theorem~3.1), while the \texttt{RefineInterceptor} guarantees zero repeated failed actions in default pruning mode (Theorem~B.7).

\textbf{Contribution 2: Unified conflict resolution and drift adaptation} (Section 4).
PRECEPT maintains a growing failure-constraint history (Evo-Memory) and treats two deployment-time conflict types in one framework: \textit{Type~I} static-vs-dynamic source conflict, handled by a six-signal ensemble detector with Bayesian source reliability and Thompson Sampling; and \textit{Type~II} environmental drift, handled by confidence decay and threshold-based rule invalidation.  In the modeled stationary-segment analysis, this yields a $64\times$ drift-resilience ratio (Corollary~B.5).

\textbf{Contribution 3: COMPASS, a dual-frequency adaptation layer} (Section 5).
COMPASS combines a \textit{high-frequency} runtime layer for per-step action/error monitoring with a \textit{low-frequency} event-triggered prompt-evolution layer.  It extends GEPA \citep{agrawal2025} with ML-based complexity estimation, rollout allocation, MAP-Elites-style diversity maintenance \citep{mouret2015}, and bi-objective selection over success and efficiency.  Crucially, candidate prompts are evaluated through PRECEPT's full retrieval-and-adaptation pipeline rather than by heuristic prompt scoring alone.

\textbf{Validation} (Sections 6--7).
Closed-form analysis predicts a $22.6\times$ advantage at $N{=}10$ conditions (Corollary~B.3) and a $64\times$ model-based drift-resilience ratio (Corollary~B.5).  Seven core experiments across three domains support these trends empirically, including +41.1pp first-try advantage, +33.3pp compositional generalization, 100\% $P_1$ on 2-way logistics compositions, +40--55pp continuous learning, strong rule persistence, robust eventual behavior under adversarial static knowledge, and +55.0pp drift recovery.  Additional architectural stress tests and COMPASS ablations characterize where the outer loop matters most.

\section{PRECEPT Framework Overview}

\subsection{System Architecture}

\begin{figure}[htbp]
  \centering
  \resizebox{0.98\textwidth}{!}{\begin{tikzpicture}[
    >=Stealth,
    font=\small,
    module/.style={
        rectangle, rounded corners=3pt,
        draw=#1!55, fill=#1!6,
        line width=0.7pt,
        text width=4.0cm, minimum height=0.95cm,
        align=center, inner sep=3.5pt
    },
    terminal/.style={
        rectangle, rounded corners=3pt,
        draw=gray!55, fill=gray!4,
        line width=0.7pt,
        minimum width=2.2cm, minimum height=0.80cm,
        align=center, inner sep=3pt
    },
    lane/.style={
        rectangle, rounded corners=6pt,
        draw=#1!45, fill=none,
        line width=0.7pt
    },
    arr/.style={->, line width=0.7pt, draw=black!65},
    biarr/.style={<->, line width=0.8pt, draw=black!65},
    fbarr/.style={->, line width=0.7pt, dashed, draw=violet!55},
    lbl/.style={font=\scriptsize, text=black!55, fill=white, inner sep=1.5pt}
]

\node[module=blue]          (c1)  at (-3.2,  4.0)
    {\textbf{Task Orchestration}\\[-1pt]{\scriptsize hybrid parse + complexity analysis}};
\node[module=teal]          (c2)  at (-3.2,  2.0)
    {\textbf{COMPASS Hi-Freq Monitor}\\[-1pt]{\scriptsize evaluate\_action/error, learn\_pattern}};
\node[module=teal!70!violet](c2t) at (-3.2,  0.0)
    {\textbf{Trigger Events}\\[-1pt]{\scriptsize new rule $|$ goal failure $|$ phase change}};
\node[module=orange]        (c3)  at (-3.2, -2.0)
    {\textbf{Retrieval \& Decision}\\[-1pt]{\scriptsize exact, compositional, hybrid, LLM}\\[-1pt]{\scriptsize + RefineInterceptor}};
\node[module=violet]        (c4)  at (-3.2, -4.0)
    {\textbf{Learning \& Adaptation}\\[-1pt]{\scriptsize confidence, rule/procedure writes}};

\node[module=green!60!black] (s1)  at (4.2,  4.0)
    {\textbf{MCP Tool Gateway}\\[-1pt]{\scriptsize stdio / JSON-RPC dispatch}};
\node[module=cyan!60!black]  (s2)  at (4.2,  2.0)
    {\textbf{Knowledge \& Conflict}\\[-1pt]{\scriptsize dual-mode retrieval, conflict resolution}};
\node[module=teal!50!orange] (s2l) at (4.2,  0.0)
    {\textbf{COMPASS Lo-Freq Architect}\\[-1pt]{\scriptsize GEPAEvolutionEngine, Pareto, MAP-Elites}};
\node[module=lime!60!black]  (s3)  at (4.2, -2.0)
    {\textbf{Domain Executors}\\[-1pt]{\scriptsize strategy-driven tool actions}};
\node[module=green!50!black] (s4)  at (4.2, -4.0)
    {\textbf{Persistent Stores}\\[-1pt]{\scriptsize static KB, dynamic, episodic, rules}};

\node[lane=blue, fit=(c1)(c2)(c2t)(c3)(c4), inner sep=0.28cm] (CL) {};
\node[lane=green!60!black, fit=(s1)(s2)(s2l)(s3)(s4), inner sep=0.28cm] (SL) {};

\node[font=\small\bfseries, text=black!75, anchor=south]
    at ([yshift=2pt]CL.north) {\strut Client: \texttt{precept\_agent.py}};
\node[font=\small\bfseries, text=black!75, anchor=south]
    at ([yshift=2pt]SL.north) {\strut Server: \texttt{precept\_mcp\_server.py}};

\node[terminal] (task)   at (-7.6,  4.0)  {Task string};
\node[terminal] (result) at (-7.6, -4.0)  {Execution result};

\draw[arr] (task.east)  -- (c1.west);
\draw[arr] (c1.south)   -- (c2.north);
\draw[arr] (c2.south)   -- (c2t.north);
\draw[arr] (c2t.south)  -- (c3.north);
\draw[arr] (c3.south)   -- (c4.north);
\draw[arr] (c4.west)    -- (result.east);

\draw[arr] (s1.south)   -- (s2.north);
\draw[arr] (s2.south)   -- (s2l.north);
\draw[arr] (s2l.south)  -- (s3.north);
\draw[arr] (s3.south)   -- (s4.north);

\draw[biarr] ([yshift=-8pt]CL.south east) --
    node[lbl, below] {\textit{stdio / JSON-RPC: tool calls, retrieval, execution, knowledge writes}}
    ([yshift=-8pt]SL.south west);

\draw[arr] (c2t.east) -- node[lbl, above] {\scriptsize trigger via MCP} (s2l.west);

\draw[fbarr]
    ([yshift=5pt]s2l.west) -- ++(-0.55, 0)
    |- node[lbl, pos=0.22, left, align=right] {evolved\\prompt}
    (c2.east);

\end{tikzpicture}}
  \caption{PRECEPT architecture overview.  The client handles orchestration, high-frequency monitoring, retrieval-time decision support, and learning updates; the server handles MCP dispatch, conflict-aware retrieval, low-frequency COMPASS evolution, domain execution, and persistent memory.  The dashed arrow denotes the evolved prompt flowing from the server-side architect back to the client monitor.}
  \label{fig:mcp_arch}
\end{figure}
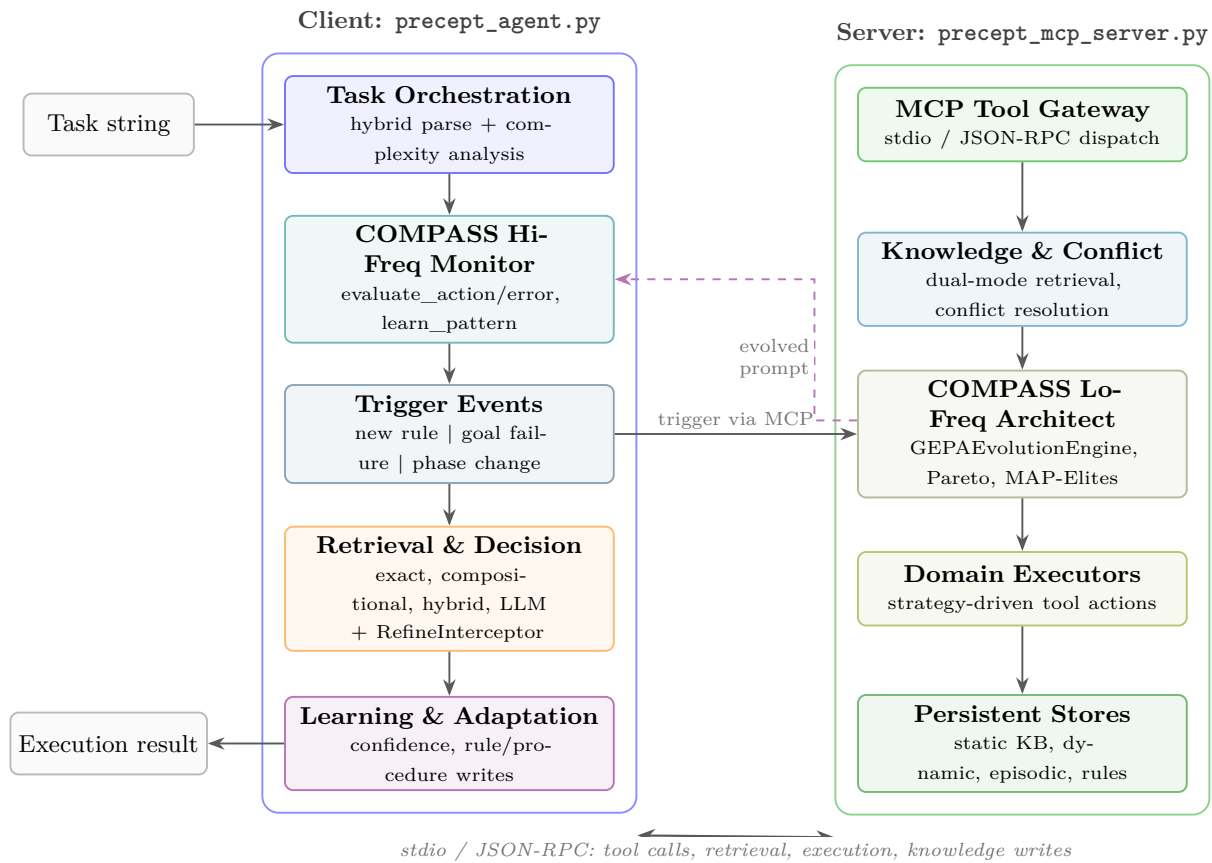

While theoretically complex, PRECEPT's operational overhead is minimized via encapsulation: the heavy lifting (Bayesian tracking, Evo-Memory, 6-method ensemble) is abstracted behind an MCP server with domain-agnostic default hyperparameter configurations, allowing the client-agent loop to remain lightweight.  Figure~\ref{fig:mcp_arch} provides a high-level overview.  On the \textbf{client}, the runtime processes each task through orchestration, the COMPASS high-frequency monitor ($O(1)$ constraint checks, error evaluation, pattern learning at every step), and trigger events that activate the server-side low-frequency path.  Retrieval and decision uses exact-match, compositional, hybrid, or LLM reasoning, guarded by the client-side \texttt{RefineInterceptor}.  On the \textbf{server}, the MCP gateway dispatches tool calls; the knowledge and conflict engine provides dual-mode retrieval with ensemble conflict resolution (Bayesian posteriors, Thompson Sampling); the COMPASS low-frequency architect (\texttt{GEPAEvolutionEngine}, Pareto selection, MAP-Elites) executes prompt evolution when triggered by the client; domain executors run strategy-driven actions; and persistent stores hold static KB, dynamic experience, episodic memory, and learned rules.  The evolved prompt feeds back from the server to the client hi-freq monitor (dashed arrow).  The full COMPASS architecture is detailed in Figures~\ref{fig:compass} and~\ref{fig:dual_freq}.

\subsection{Theoretical Foundations: Code-to-Theory Mapping}

\textbf{Theoretical Foundations for PRECEPT's Key Features.}

\begin{table}[htbp]
\centering
\small
\begin{tabularx}{\textwidth}{>{\raggedright\arraybackslash}p{0.16\textwidth}|>{\raggedright\arraybackslash}X|>{\raggedright\arraybackslash}X|>{\raggedright\arraybackslash}p{0.11\textwidth}}
\toprule
Feature & Code Implementation & Theoretical Foundation & Section \\
\midrule
\textbf{Dual-Mode Retrieval} & \texttt{fetch\_context*()} + server retrieval tools:\newline \texttt{retrieve\_memories()}, \texttt{get\_rule\_hybrid()},\newline \texttt{retrieve\_atomic\_precepts()}; unified interface:\newline \texttt{retrieve\_with\_dual\_mode()} & Definition 2.2 + retrieval-tier specification (\S{}2.3.1) & \S{}2.3.1 \\
\textbf{Type I Conflict (Static vs Dynamic)} & \texttt{EnsembleConflictDetector.detect()} & Ensemble weight specification + Definition 4.1 + Algorithm 4.1a & \S{}4.1 \\
\textbf{Evo-Memory} & \texttt{partial\_progress}, \texttt{failed\_options} & Definition 4.0, Theorem 4.0 (Eliminates Cyclic Failures) & \S{}4.0 \\
\textbf{Deterministic Pruning} & \texttt{RefineInterceptor.is\_forbidden()}, \texttt{add\_constraint()} & Theorem 4.5 (P(repeat\_fail)=0), Corollary 4.5.1 & \S{}4.5 \\
\textbf{Thompson Sampling} & \texttt{BetaDistribution.sample()} & Definition 4.1, Theorem 4.1 (Local stationary-segment bound) & \S{}4.1 \\
\textbf{Bayesian Conflict Resolution} & \texttt{BetaDistribution.update(success)} & Definition 4.1, Algorithm 4.1a & \S{}4.1 \\
\textbf{MAP-Elites / Pareto Diversity} & \texttt{pareto\_front}, \texttt{diversity\_threshold},\newline \texttt{diversity\_rollouts} & Definition 5.1, 5.2 (Topological Distinctness) & \S{}5.1.1 \\
\textbf{Epistemic Probing} & \texttt{\_DIAGNOSTIC\_PROBES}, \texttt{enable\_epistemic\_probing} & Definition 2.1, Algorithm 2.1 & \S{}2.5 \\
\textbf{Compositional Stacking} & \texttt{retrieve\_atomic\_precepts()}, tier sorting & Theorem 3.1 ($2^N-1$ Coverage) & \S{}3 \\
\bottomrule
\end{tabularx}
\caption{Code-to-theory mapping for PRECEPT's major components.}
\end{table}

\textit{All implementations verified in source code; theoretical proofs in respective sections.  Appendix~\ref{app:agent_capabilities} distinguishes the main evaluated runtime path from supporting server-side tools and auxiliary utilities.}

\subsection{PRECEPTAgent Detailed Execution Flow}

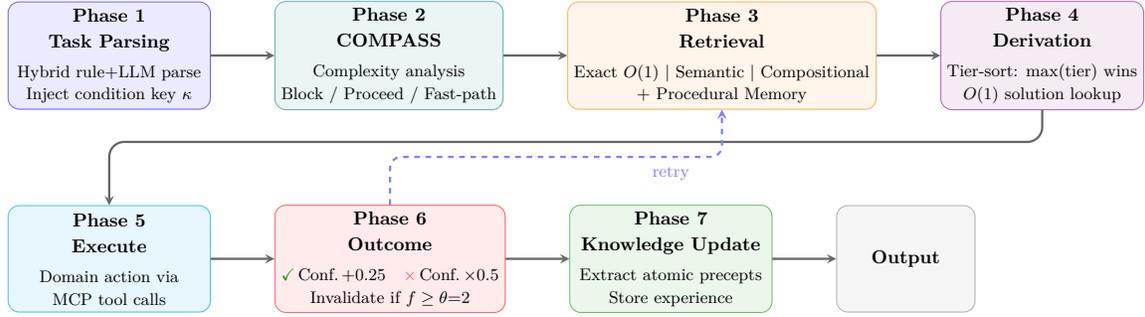
\begin{figure*}[t]
\centering
\resizebox{0.92\textwidth}{!}{%
\begin{tikzpicture}[
    >=stealth,
    node distance=0.9cm and 1.2cm,
    every node/.style={font=\normalsize},
    phase/.style={rectangle, rounded corners=6pt, draw=#1!60, fill=#1!8,
        minimum width=3.8cm, minimum height=2cm, align=center, font=\normalsize,
        drop shadow={shadow xshift=0.5pt, shadow yshift=-0.5pt, opacity=0.12}},
    arrow/.style={->, very thick, color=black!60},
    elabel/.style={font=\small, color=black!55, align=center},
]


\node[phase=blue] (p1) {\textbf{Phase 1}\\[2pt]\textbf{Task Parsing}\\[4pt]{\small Hybrid rule+LLM parse}\\{\small Inject condition key $\kappa$}};

\node[phase=teal, right=of p1] (p2) {\textbf{Phase 2}\\[2pt]\textbf{COMPASS}\\[4pt]{\small Complexity analysis}\\{\small Block / Proceed / Fast-path}};

\node[phase=orange, right=of p2] (p3) {\textbf{Phase 3}\\[2pt]\textbf{Retrieval}\\[4pt]{\small Exact $O(1)$ $\mid$ Semantic $\mid$ Compositional}\\{\small + Procedural Memory}};

\node[phase=violet, right=of p3] (p4) {\textbf{Phase 4}\\[2pt]\textbf{Derivation}\\[4pt]{\small Tier-sort: $\max(\text{tier})$ wins}\\{\small $O(1)$ solution lookup}};

\draw[arrow] (p1) -- (p2);
\draw[arrow] (p2) -- (p3);
\draw[arrow] (p3) -- (p4);


\node[phase=cyan, below=1.8cm of p1] (p5) {\textbf{Phase 5}\\[2pt]\textbf{Execute}\\[4pt]{\small Domain action via}\\{\small MCP tool calls}};

\node[phase=red, right=of p5] (p6) {\textbf{Phase 6}\\[2pt]\textbf{Outcome}\\[4pt]{\small \textcolor{green!60!black}{\checkmark}\,Conf.\,+0.25\quad \textcolor{red!70}{$\times$}\,Conf.\,$\times$0.5}\\{\small Invalidate if $f \geq \theta{=}2$}};

\node[phase=green!60!black, right=of p6] (p7) {\textbf{Phase 7}\\[2pt]\textbf{Knowledge Update}\\[4pt]{\small Extract atomic precepts}\\{\small Store experience}};

\node[phase=gray, right=of p7, minimum width=2.6cm] (out) {\textbf{Output}};

\draw[arrow, rounded corners=8pt] (p4.south) -- ++(0,-0.6) -| (p5.north);

\draw[arrow] (p5) -- (p6);
\draw[arrow] (p6) -- (p7);
\draw[arrow] (p7) -- (out);

\draw[arrow, dashed, color=blue!50, rounded corners=6pt]
    (p6.north) -- ++(0,0.9) -| (p3.south);
\node[elabel, color=blue!50] at ([yshift=0.6cm]p7.north) {retry};

\end{tikzpicture}
}%
\caption{Complete seven-phase execution flow of the PRECEPT agent: (1)~Task parsing with hybrid rule+LLM fallback, (2)~COMPASS complexity evaluation with block/proceed/fast-path decisions, (3)~Three-mode context retrieval (compositional, hybrid, semantic), (4)~Solution derivation with tier-sorted priority, (5)~Domain action execution, (6)~Outcome processing with threshold-based invalidation ($\theta{=}2$), and (7)~Knowledge update via atomic precept extraction. Dashed arrow indicates the retry loop.}
\label{fig:exec_flow}
\end{figure*}

Figure~\ref{fig:exec_flow} presents the complete seven-phase execution flow of \texttt{PRECEPTAgent.run\_task()}, illustrating how PRECEPT orchestrates task parsing, knowledge retrieval, decision-making, and adaptive learning within a single unified pipeline.

The seven phases span task parsing (rule-based by default, with optional hybrid rule-based + LLM fallback), COMPASS evaluation, compositional + hybrid retrieval, direct resolution or LLM reasoning, domain-specific execution, success/failure handling with confidence updates, and atomic precept learning.  Implementation details are in Appendix~\ref{app:agent_capabilities}.

\subsection{Knowledge Layer Architecture}

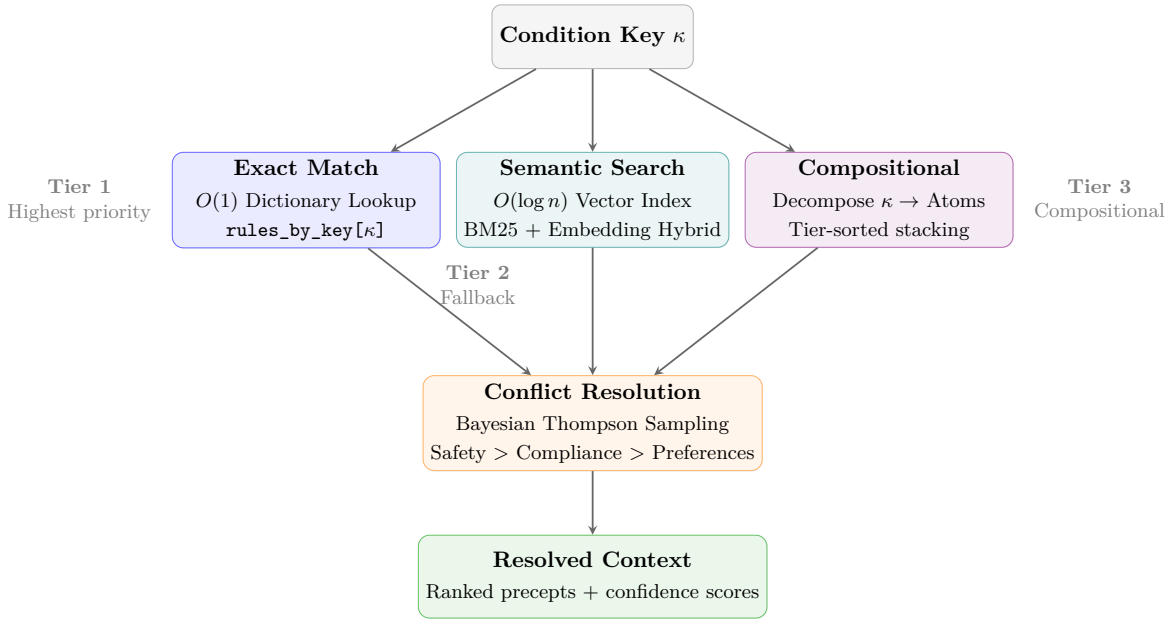
\begin{figure}[t]
\centering
\resizebox{0.95\textwidth}{!}{%
\begin{tikzpicture}[
    >=stealth,
    node distance=1cm and 1.5cm,
    every node/.style={font=\normalsize},
    tier/.style={rectangle, rounded corners=5pt, draw=#1!60, fill=#1!8,
        minimum width=4.2cm, minimum height=1.3cm, align=center, font=\normalsize,
        drop shadow={shadow xshift=0.4pt, shadow yshift=-0.4pt, opacity=0.12}},
    query/.style={rectangle, rounded corners=4pt, draw=gray!60, fill=gray!8,
        minimum width=3cm, minimum height=1cm, align=center, font=\normalsize},
    arrow/.style={->, thick, color=black!60},
    annot/.style={font=\small, color=black!50, align=center},
]

\node[query] (q) {\textbf{Condition Key} $\kappa$};

\node[tier=blue, below left=1.3cm and 0.8cm of q] (exact) {\textbf{Exact Match}\\[2pt]{\small$O(1)$ Dictionary Lookup}\\[1pt]{\small\texttt{rules\_by\_key[$\kappa$]}}};

\node[tier=teal, below=1.3cm of q] (semantic) {\textbf{Semantic Search}\\[2pt]{\small$O(\log n)$ Vector Index}\\[1pt]{\small BM25 + Embedding Hybrid}};

\node[tier=violet, below right=1.3cm and 0.8cm of q] (comp) {\textbf{Compositional}\\[2pt]{\small Decompose $\kappa \to$ Atoms}\\[1pt]{\small Tier-sorted stacking}};

\draw[arrow] (q) -- (exact);
\draw[arrow] (q) -- (semantic);
\draw[arrow] (q) -- (comp);

\node[annot, left=0.2cm of exact] {\textbf{Tier 1}\\Highest priority};
\node[annot, below=0.15cm of semantic, xshift=-1.8cm] {\textbf{Tier 2}\\Fallback};
\node[annot, right=0.2cm of comp] {\textbf{Tier 3}\\Compositional};

\node[tier=orange, below=2cm of semantic] (merge) {\textbf{Conflict Resolution}\\[2pt]{\small Bayesian Thompson Sampling}\\[1pt]{\small Safety $>$ Compliance $>$ Preferences}};

\draw[arrow] (exact) -- (merge);
\draw[arrow] (semantic) -- (merge);
\draw[arrow] (comp) -- (merge);

\node[tier=green!60!black, below=1cm of merge] (out) {\textbf{Resolved Context}\\[2pt]{\small Ranked precepts + confidence scores}};

\draw[arrow] (merge) -- (out);

\end{tikzpicture}
}%
\caption{Knowledge Layer with three retrieval modes: exact-match $O(1)$ via dictionary lookup (highest priority), semantic similarity $O(\log n)$ via hybrid BM25+embedding search, and compositional retrieval via atomic precept decomposition with tier-sorted stacking. Conflicts are resolved through Bayesian Thompson sampling respecting the Safety~$>$~Compliance~$>$~Preferences hierarchy.}
\label{fig:knowledge_layer}
\end{figure}

Figure~\ref{fig:knowledge_layer} illustrates PRECEPT's four-tier knowledge layer: static KB (vector DB), dynamic experience (vector DB), episodic memory (trajectory store), and learned rules (hash table).  Each tier has a distinct confidence prior and retrieval mechanism; the layers interact through the dual-mode retrieval interface described below.

\subsection{Dual-Mode Retrieval with Conflict Resolution}

PRECEPT provides a unified server-side retrieval interface (\texttt{retrieve\_with\_dual\_mode()}) that queries all knowledge sources in parallel and automatically resolves conflicts.  In the evaluated \texttt{PRECEPTAgent} runtime, client-side retrieval is orchestrated through \texttt{fetch\_context()}, \texttt{fetch\_context\_with\_hybrid()}, and \texttt{fetch\_context\_compositional()}.  These helpers call server-side tools including \texttt{retrieve\_memories()}, \texttt{get\_rule\_hybrid()}, and \texttt{retrieve\_atomic\_precepts()}.  In addition, PRECEPT implements a \textbf{3-Tier Hybrid Retrieval} strategy (\texttt{get\_rule\_hybrid()}): \textbf{Tier~1} provides $O(1)$ exact hash lookup for deterministic rule application; \textbf{Tier~2} uses vector similarity (ChromaDB cosine search, optionally combined with BM25 via an ensemble retriever) for semantic matching over unseen or continuous representations; and \textbf{Tier~3} applies Jaccard similarity over condition codes as a structural fallback.  This cascading design ensures that while $O(1)$ hashes guarantee deterministic retrieval for discrete constraints, the architecture natively handles continuous, high-cardinality, and out-of-vocabulary state spaces through dense embedding fallback.  The full algorithm is in Appendix~\ref{app:agent_capabilities}.

\begin{table}[htbp]
\centering
\footnotesize
\resizebox{\textwidth}{!}{%
\begin{tabular}{l|l|l|l|l|l}
\toprule
Tier & Storage & Source & Confidence Prior & Retrieval & Use Case \\
\midrule
\textbf{Static KB} & Vector DB & Pre-deployment (PDFs, docs) & 0.9 (high) & Semantic similarity & Factual grounding \\
\textbf{Dynamic Experience} & Vector DB & Runtime (web, APIs) & 0.8 (moderate) & Semantic + metadata filter & Current information \\
\textbf{Episodic Memory} & MemoryStore & Task executions & Task-dependent & Trajectory matching & Experience replay \\
\textbf{Learned Rules} & Hash Table & Successful solutions & 1.0 (deterministic) & $O(1)$ exact match & Direct application \\
\bottomrule
\end{tabular}
}
\caption{PRECEPT's four knowledge tiers, their storage backends, priors, and retrieval roles.}
\end{table}

\textbf{Phase~1 (Parallel Retrieval)} queries three tiers: static KB via semantic similarity, dynamic experiences via semantic+metadata filtering, and episodic memory via trajectory matching.  \textbf{Phase~2 (Conflict Detection)} applies the six-method ensemble detector (ensemble-weight specification in \S{}4.1) to all static--dynamic pairs, triggering resolution when $\text{weighted\_conflict} \geq \theta{=}0.30$ or when any single method votes conflict with confidence $\geq 0.60$ (circuit breaker).  Detected conflicts invoke the Type~I resolution mechanism (Section~4.1): Thompson Sampling from Beta posteriors, with posterior updates based on resolution outcomes.  This pipeline makes PRECEPT a \textbf{conflict-aware RAG system}---the key differentiator from standard RAG, which returns results without detecting or resolving contradictions.

\subsection{Simplified Execution Pipeline}

\begin{figure*}[t]
\centering
\resizebox{0.88\textwidth}{!}{%
\begin{tikzpicture}[
    >=stealth,
    every node/.style={font=\normalsize},
    process/.style={rectangle, rounded corners=5pt, draw=black!60, fill=white,
        minimum width=3.2cm, minimum height=1.5cm, align=center,
        drop shadow={shadow xshift=0.4pt, shadow yshift=-0.4pt, opacity=0.12}},
    monitor/.style={process, fill=blue!8, draw=blue!50},
    success/.style={process, fill=green!8, draw=green!50!black!40},
    fail/.style={process, fill=red!8, draw=red!40},
    retrieval/.style={process, fill=orange!8, draw=orange!50},
    decision/.style={diamond, draw=black!60, fill=yellow!6, aspect=2.2,
        minimum width=1.8cm, inner sep=3pt, align=center, font=\normalsize},
    arrow/.style={->, very thick, color=black!60},
    darrow/.style={->, thick, dashed, color=blue!50},
    elabel/.style={font=\small, color=black!55},
]

\node[process] (input) at (0,0)
    {\textbf{Task Input}\\[2pt]{\small Parse with hybrid}\\{\small rule + LLM}};

\node[monitor] (compass) at (4.8,0)
    {\textbf{COMPASS Monitor}\\[2pt]{\small $O(1)$ constraint check}\\{\small Block / Proceed / Fast}};

\node[retrieval] (retrieve) at (9.6,0)
    {\textbf{3-Mode Retrieval}\\[2pt]{\small Exact $O(1)$ $\mid$ Semantic}\\{\small $\mid$ Compositional stacking}};

\node[process] (compose) at (14.4,0)
    {\textbf{Compose Solution}\\[2pt]{\small Max tier wins}\\{\small Conflict? $\to$ Thompson}};

\draw[arrow] (input) -- (compass);
\draw[arrow] (compass) -- (retrieve);
\draw[arrow] (retrieve) -- (compose);

\node[process] (execute) at (0,-3.8)
    {\textbf{Execute Action}\\[2pt]{\small Domain-specific}\\{\small MCP tool calls}};

\node[decision] (result) at (4.8,-3.8) {\textbf{Result?}};

\node[success] (win) at (9.6,-3.8)
    {\textbf{Success Path}\\[2pt]{\small Conf.\,+0.25}\\{\small Record rule $\to$ Learn}};

\node[fail] (lose) at (14.4,-3.8)
    {\textbf{Failure Path}\\[2pt]{\small Conf.\,$\times$0.5}\\{\small $f{\geq}2$? Invalidate : Retry}};

\draw[arrow, rounded corners=6pt]
    (compose.south) -- ++(0,-0.6) -| (execute.north);
\node[elabel] at (7.2,-1.2) {execute};

\draw[arrow] (execute) -- (result);
\draw[arrow] (result) -- node[elabel, above] {\checkmark} (win);
\draw[arrow, rounded corners=6pt]
    (result.south) -- ++(0,-1.0) -| (lose);
\node[elabel] at (4.2,-4.6) {$\times$\,fail};

\node[monitor, minimum width=4.2cm] (architect) at (4.8,-7.5)
    {\textbf{COMPASS Architect}\\[2pt]{\small Low-frequency prompt evolution}\\{\small + Pareto-optimal selection}};

\node[success, minimum width=2.6cm, minimum height=1.2cm] (done) at (9.6,-7.5)
    {\textbf{Done}\\{\small Rule persisted}};

\draw[arrow, rounded corners=6pt]
    (win.south) -- ++(0,-1.0) -| (done.north);
\node[elabel] at (11.2,-5.2) {persist rule};

\draw[darrow, rounded corners=6pt, color=teal!60]
    ([xshift=-4pt]win.south) -- ++(0,-1.8) -| (architect.north);
\node[elabel, color=teal!60] at (3.0,-5.8) {\textit{state-change trigger}};

\draw[arrow, color=teal!60] (architect) -- (done)
    node[elabel, midway, above] {evolved};

\draw[->, very thick, dashed, color=red!50, rounded corners=8pt]
    (lose.north) -- (14.4,-2.0) -- (4.8,-2.0) -- (compass.south);
\node[elabel, color=red!60] at (9.6,-1.7) {\textit{retry}};

\end{tikzpicture}
}%
\caption{Complete PRECEPT execution pipeline. Tasks flow through the COMPASS Monitor ($O(1)$ constraint check), three-mode retrieval (exact-match, semantic, compositional), Bayesian conflict resolution via Thompson sampling, execution with deterministic pruning via RefineInterceptor, and threshold-based rule invalidation ($\theta{=}2$). On success, rules are persisted; on state-change events, the COMPASS Architect (low-frequency loop, dashed teal) triggers prompt evolution and Pareto selection. On failure, the retry loop (dashed red) returns to the Monitor with the failed option pruned.}
\label{fig:pipeline}
\end{figure*}
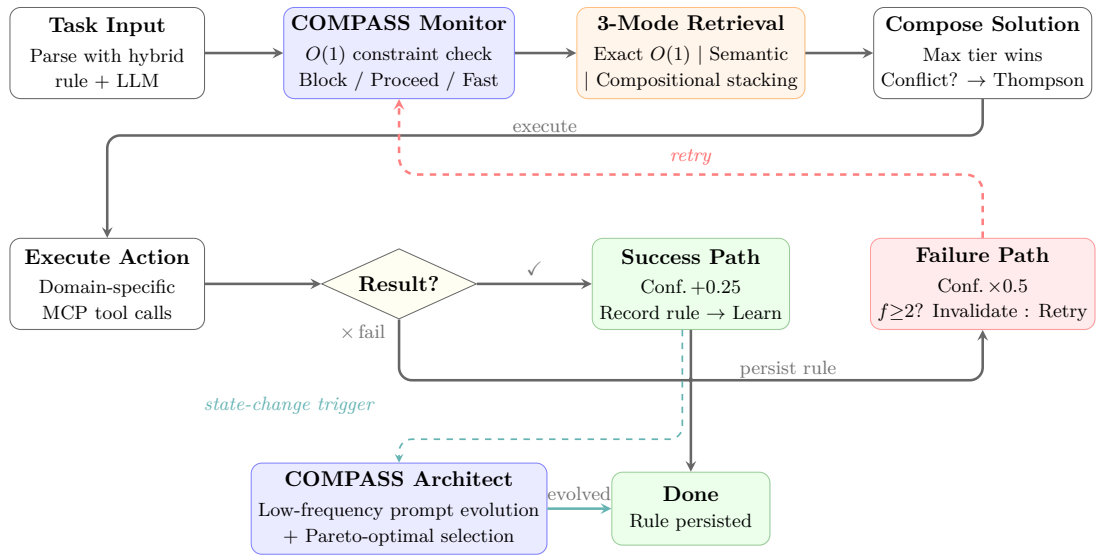

Figure~\ref{fig:pipeline} presents the simplified six-stage execution pipeline, mapping each stage to its client/server location and the corresponding MCP tool.

\textbf{Execution Pipeline Stage Details with MCP Tool Mapping.}

\begin{table}[htbp]
\centering
\small
\begin{tabularx}{\textwidth}{>{\raggedright\arraybackslash}p{0.1\textwidth}|>{\raggedright\arraybackslash}p{0.14\textwidth}|>{\raggedright\arraybackslash}X|>{\raggedright\arraybackslash}X}
\toprule
Stage & Client/Server & MCP Tool / Function & Key Operation \\
\midrule
(1) Parse & Client & \texttt{PRECEPTAgent.run\_task()} & Parse task; inject condition metadata when available \\
(2) Retrieve & Client$\rightarrow$Server & \texttt{fetch\_context*()} $\rightarrow$ \texttt{retrieve\_memories()}, \texttt{get\_rule\_hybrid()}, \texttt{retrieve\_atomic\_precepts()} & Orchestrate retrieval for the reported runtime path \\
(3) Compose & Client + Server & Tier sorting + constraint-stack construction & Use deterministic highest-tier fast path when directly applicable; otherwise pass stacked constraints to synthesis \\
(4) Detect & Server & \texttt{ConflictManager.detect()} + atomic-precept conflict checks & Resolve static--dynamic conflicts; compositional retrieval also checks atomic-precept conflicts by default \\
(5) Execute & Server & Domain tools (\texttt{book\_shipment}, etc.) & Apply selected or synthesized solution via domain-specific tool \\
(6) Handle & Client$\rightarrow$Server & \texttt{report\_rule\_failure($\kappa$)} / \texttt{record\_rule\_success($\kappa$)} & Decay $\times$0.5, restore +0.25, invalidate at $\theta$=2 \\
\bottomrule
\end{tabularx}
\caption{Simplified client--server execution pipeline with MCP tool mapping.}
\end{table}

\subsection{Additional Agent Capabilities}

Beyond the core mechanisms described above, PRECEPT includes several supporting capabilities that contribute to its robustness (detailed in Appendix~\ref{app:agent_capabilities}):

\textbf{Epistemic Probing.}  Rather than passively logging errors, PRECEPT actively probes the environment to discover hidden constraints before committing to plans---operationalizing the adversarial dynamics from evolutionary computing \citep{kumar2026}.  Formally, given environment $\mathcal{E}$ with hidden constraints $\mathcal{C}(\mathcal{E}) = \{c_1, \ldots, c_n\}$, epistemic probing maps targeted queries to constraint evidence: $\text{Probe}: \mathcal{E} \times q \rightarrow \text{Evidence}(c_i)$.  Discovered constraints are immediately added to the \texttt{RefineInterceptor}'s forbidden set, transforming error handling from reactive to proactive.

\textbf{Structured Outputs and Hybrid Parsing.}  LLM responses are parsed through Pydantic-validated structured outputs (\texttt{ReasoningResponse}) ensuring guaranteed schema compliance.  The reported experiments retain the default rule-based task parser; when enabled, PRECEPT also supports a two-stage hybrid approach with LLM fallback when rule-based confidence $<$~0.8, combining speed with robustness.

\textbf{Cross-Episode Learning.}  Failed options persist across episodes via \texttt{context.failed\_options}, enabling resumable exploration.  Successful error recoveries are stored as procedural memory (\texttt{store\_procedure()}) for future reuse.  A Strategy Pattern provides domain-specific behavior for the three evaluated domains (logistics, booking, integration) through a common interface.

\section{Atomic Constraint Stacking}

\subsection{Problem: Compositional Explosion}

Given N atomic conditions, there are $2^N$ possible combinations. Training on all combinations is infeasible.

\textbf{Example:}
\begin{itemize}
  \item Illustrative 6-atomics subset: {SAFE, ASIA, EURO, FAST, ECON, BULK}
  \item $2^6$ = 64 possible combinations
  \item Training all 64 requires $O(64 \times \beta)$ = $O(192)$ tasks at $\beta$=3
\end{itemize}

\textit{Implementation note (Q2).}  The six-condition set above is illustrative.  In the evaluated compositional runs, atomic condition vocabularies are domain-provided by scenario generators/configs (e.g., logistics semantic mode uses 8 codes: ASIA, EURO, AMER, INTL, FAST, ECON, SAFE, BULK).  PRECEPT learns rule/precept associations from execution outcomes over this provided symbolic vocabulary; it does not discover arbitrary new symbolic tokens from raw unstructured text.

\textbf{PRECEPT Solution:} Learn N atomic precepts, compose at test time via tier-based resolution using \texttt{SEMANTIC\_CONDITION\_TIERS}.

\subsection{Semantic Tier Hierarchy}

\textbf{Semantic Tier Hierarchy from \texttt{SEMANTIC\_CONDITION\_TIERS} in \texttt{precept\_mcp\_\allowbreak{}server.py}.}

\begin{table}[htbp]
\centering
\footnotesize
\resizebox{\textwidth}{!}{%
\begin{tabular}{c|c|l|l|l}
\toprule
Tier & Priority & Category & Conditions & Dominance Rule \\
\midrule
\textbf{3} & Highest & Safety & SAFE, SECURE, RISK, CANCEL, AUTH & Always wins \\
\textbf{2} & Medium & Compliance & ASIA, EURO, INTL, HIPAA, AUDIT & Wins over Tier 1 \\
\textbf{1} & Lowest & Preferences & FAST, ECON, BULK, SPEED, COST & Default fallback \\
\bottomrule
\end{tabular}
}
\caption{Semantic tier hierarchy used for atomic constraint stacking.}
\end{table}

\textbf{Resolution Rule:} Given conditions {A, B, C}, solution $\leftarrow$ precept(argmax{tier(A), tier(B), tier(C)})


\subsection{Composition Algorithm}

Algorithm~\ref{alg:atomic_constraint_stacking} details the deterministic direct-application fast path used when compositional retrieval returns multiple tier-annotated atomic precepts and the runtime can directly apply the highest-priority returned precept.  Given a set of retrieved atomic precepts, it sorts them by tier in descending order and extracts the solution hint from the highest-priority precept.  The algorithm handles both direct solutions and exploration paths (e.g., \texttt{LLM$\to$origin$\to$dest}), parsing each format deterministically.  The overall complexity is $O(N \log N)$ for $N$ precepts (dominated by the sort).  In the full evaluated pipeline, this fast path is complemented by constraint-stack construction and LLM synthesis whenever direct application is not taken.

\begin{algorithm}[htbp]
\caption{Atomic Constraint Stacking}
\label{alg:atomic_constraint_stacking}
\begin{lstlisting}[style=pseudocode]
Input: compositional_context from fetch_context_compositional()
Output: compositional_direct_solution (if direct fast path is taken)

function COMPOSE(compositional_context):
    precepts_found <- compositional_context.precepts_found
    // Guard: only for multi-constraint scenarios
    if |precepts_found| <= 1 then
        return ?
    end if
    // Sort by tier (descending) - highest priority first
    sorted_precepts <- sorted(precepts_found,
    key=lambdap: p.get("tier", 1),
    reverse=True)
    // Get solution from highest-tier precept
    highest_precept <- sorted_precepts[0]
    solution_hint <- highest_precept.get("solution_hint", "")
    // Parse solution_hint format: "solution:value" or "solution:LLM->value"
    if ":" in solution_hint then
        raw <- solution_hint.split(":", 1)[1]
        if "->" in raw then
            // Handle exploration paths like "LLM->hamburg->singapore"
            for part in raw.split("->") do
                if part.lower() != "llm" and part.strip() then
                    return part.strip()
                end if
            end for
        else
            return raw.strip()
        end if
    else
        return solution_hint.strip() if solution_hint else ?
    end if
end function
----------------------------------------------------------------------
Complexity: O(N log N) for N precepts
Property: Deterministic--no LLM interpretation required for tier resolution
\end{lstlisting}
\end{algorithm}

\subsection{Compositional Generalization Property}

\textbf{Theorem 3.1 (Conditional Compositional Coverage).} \textit{Given N atomic precepts learned from N training scenarios, PRECEPT can construct a resolution candidate for up to $2^N - 1$ non-empty composite test scenarios.  When the retrieved precepts yield a unique tier-maximal resolution, this candidate is obtained deterministically by highest-tier selection.  When residual same-tier ambiguity remains, the evaluated runtime falls back to constraint-stack synthesis.  If the semantic tier hierarchy faithfully matches the domain's true constraint priorities and the synthesis step succeeds on tied cases, these resolutions are correct.}

\textbf{Proof.} Let A = {a$\_1$, ..., $a\_n$} be the set of N atomic conditions, each with an associated precept $p\_i$ $\in$ P stored in \texttt{precepts\_found}, where each $p\_i$ has a tier $\tau$$_i$ $\in$ {1, 2, 3}.

Consider any non-empty composite scenario S $\subseteq$ A. We show PRECEPT constructs a resolution candidate for S, and is correct under the tier-ordering and synthesis assumptions:

\begin{enumerate}
  \item \textbf{Decomposition}: \texttt{retrieve\_atomic\_precepts} decomposes S into constituent atoms. For each $a\_i$ $\in$ S, the corresponding precept $p\_i$ is retrieved via exact-match lookup. Since each $a\_i$ was learned during training, retrieval succeeds with probability 1.
  \item \textbf{Composition}: If a unique highest-tier precept exists, Algorithm~1 computes argmax{$\tau$$_i$ : $a\_i$ $\in$ S} and returns that direct solution deterministically.  If multiple active precepts remain at the same effective priority after retrieval-time conflict handling, PRECEPT forwards their stacked constraints to synthesis rather than claiming a purely deterministic tie-break.
  \item \textbf{Conditional correctness}: If the semantic tier hierarchy faithfully reflects the domain's true priority order, then the highest-tier direct resolution is correct whenever a unique maximizer exists.  In tied cases, correctness additionally depends on successful synthesis over the stacked active constraints.
  \item \textbf{Determinism}: The highest-tier fast path is deterministic whenever the maximizer is unique; otherwise residual ambiguity is delegated to synthesis.
  \item \textbf{Coverage}: The number of non-empty subsets of A is |$\mathcal{P}$(A) \ $\emptyset$| = $2^N$ - 1.  Each such subset can be routed either through the deterministic highest-tier fast path or through the synthesis path described above.
\end{enumerate}

Therefore, N atomic precepts yield coverage of $2^N$ - 1 composite scenarios, with deterministic direct resolution whenever a unique highest-tier precept is available and synthesis handling the remaining tied cases.  Correctness is inherited from the tier-ordering assumption together with successful synthesis on residual ambiguities. $\blacksquare$

\textbf{Remark.} The guarantee is conditional and path-dependent rather than universal.  For composites whose retrieved precepts induce a unique highest tier, PRECEPT uses the deterministic fast path described above.  In the evaluated implementation, \texttt{retrieve\_atomic\_precepts()} also performs atomic-precept conflict detection by default before building the constraint stack.  When multiple active constraints of comparable priority remain, PRECEPT switches to \textit{LLM Constraint Synthesis} (stacking the competing atomic constraints into the context), rather than claiming that every compositional case is resolved by a purely deterministic shortcut.

Deterministic retrieval and compositional stacking (Sections~2--3) guarantee correct rule application \textit{given correct rules}.  The next section addresses the complementary challenge: what happens when rules conflict or become invalid.

\section{Dual Conflict Resolution and Adversarial Adaptation}

\subsection{Theoretical Foundation: Evo-Memory and the Red Queen Principle}

PRECEPT's conflict resolution draws theoretical inspiration from the \textbf{Red Queen hypothesis} in evolutionary computing. The Digital Red Queen (DRQ) framework \citep{kumar2026} demonstrates that agents optimized against static objectives become brittle, failing 72\% against novel adversarial dynamics. DRQ overcomes this by maintaining a ''growing history of opponents'' that forces continual adaptation.

\textbf{PRECEPT's Evo-Memory Architecture} operationalizes this principle for rule-governed agents:

\begin{lstlisting}
DRQ: Optimize(Warrior) against History[Opponent_1, Opponent_2, ..., Opponent_n]
PRECEPT: Optimize(Plan) against History[Constraint_1, Constraint_2, ..., Constraint_n]
\end{lstlisting}

\textbf{Definition 4.0 (Evo-Memory).} \textit{Let $H(t) = \{c_1, c_2, \ldots, c_k\}$ be the accumulated constraint history at time step $t$, where each $c_i$ is a constraint (a failed option or violated condition) discovered during episodes $1, \ldots, t$. PRECEPT's Evo-Memory comprises three components:}

\begin{enumerate}
  \item \textbf{In-Episode Memory} (\texttt{RefineInterceptor.forbidden\_set}): The set of constraints $F_{\text{episode}} \subseteq H(t)$ discovered during the current task episode.
  \item \textbf{Cross-Episode Memory} (\texttt{partial\_progress.failed\_options}): The set of constraints $F_{\text{cross}} \subseteq H(t)$ persisted from previous episodes, loaded at startup.
  \item \textbf{Rule Confidence} (\texttt{rule\_confidence[k]}): A scalar $c_k \in [0,1]$ tracking the reliability of learned rule $k$, decayed on failure ($c_k \leftarrow c_k \times 0.5$) and restored on success ($c_k \leftarrow \min(1.0, c_k + 0.25)$).
\end{enumerate}

\textit{The complete constraint history is $H(t) = F_{\text{episode}} \cup F_{\text{cross}}$.}

\textbf{Theorem 4.0 (Evo-Memory Eliminates Cyclic Failures).} \textit{For any candidate plan $P$ and accumulated constraint set $H(t)$:}
$$P(\text{repeat}(c_i) \mid \text{EvoMemory}(H(t))) = 0, \quad \forall c_i \in H(t)$$

\textit{where $\text{repeat}(c_i)$ denotes the event that the agent selects an option that violates previously-discovered constraint $c_i$.}

\textit{Proof}: The \texttt{RefineInterceptor.is\_forbidden()} check deterministically rejects any option in the forbidden set. Cross-episode failures are loaded from \texttt{partial\_progress.json} at startup and injected into context via \texttt{forbidden\_injection}. Any plan violating accumulated constraints is pruned before execution. $\blacksquare$

This architectural parallel to DRQ explains why PRECEPT achieves P(repeat\_fail) = 0---the same mechanism that enables DRQ's warriors to defeat all previous opponents enables PRECEPT's agents to satisfy all accumulated constraints.

\subsection{Type I: Static vs Dynamic Knowledge Conflict}

\textbf{Ensemble Conflict Detection Weights (\texttt{EnsembleConflictDetector}).}

\begin{table}[htbp]
\centering
\footnotesize
\resizebox{\textwidth}{!}{%
\begin{tabular}{l|c|c|l}
\toprule
Method & Raw Weight & Normalized Weight & Description \\
\midrule
NLI Classifier & 0.30 & 0.200 & Natural language inference \\
Semantic Patterns & 0.30 & 0.200 & Keyword contradiction detection \\
Temporal Analysis & 0.15 & 0.100 & Recency-based staleness \\
Evidence Strength & 0.15 & 0.100 & Confirmation/failure counts \\
Recommendation Conflict & 0.50 & 0.333 & Dismissive-vs-active strategy detection \\
LLM Vote & 0.10 & 0.067 & Optional LLM adjudication \\
\bottomrule
\end{tabular}
}%
\caption{Six-method ensemble conflict detector weights used for Type~I conflict detection.}
\end{table}

\textbf{Detection:} \texttt{weighted\_conflict = $\Sigma$(weight $\times$ is\_conflict $\times$ confidence) / $\Sigma$weight} $\rightarrow$ trigger if $\geq 0.30$, \textbf{or} if any single method votes conflict with confidence $\geq 0.60$ (circuit breaker).  Raw weights are scoring coefficients and are explicitly normalized in the denominator; they are not required to sum to 1.

The \textbf{Recommendation Conflict} method detects three patterns: (i)~dismissive static claims (e.g., ``can be ignored'', ``proceed normally'') paired with active dynamic strategies (e.g., ``strategy is X'', ``solution: Y''); (ii)~overlapping condition codes with divergent recommendations; (iii)~explicitly different solution suggestions.  Its higher per-method weight reflects that this signal captures action-level contradiction patterns that NLI/semantic similarity often miss in short operational texts.  Importantly, it does not dominate the ensemble: the combined NLI+semantic mass is still larger (0.60 raw), and the circuit breaker is used only for single high-confidence conflict evidence.

\textbf{Resolution:} Thompson Sampling from Beta posteriors: Static \texttt{Beta(5,5)}, Dynamic \texttt{Beta(5,3)}

\textbf{Definition 4.1 (Bayesian Source Reliability Model).}
\textit{Let $r_s, r_d \in [0,1]$ denote the unknown reliability of the static and dynamic sources, respectively, modeled as:}
$$r_s \sim \text{Beta}(\alpha_s, \beta_s), \quad \alpha_s^{(0)}{=}5, \; \beta_s^{(0)}{=}5 \qquad r_d \sim \text{Beta}(\alpha_d, \beta_d), \quad \alpha_d^{(0)}{=}5, \; \beta_d^{(0)}{=}3$$

The Beta distribution is the conjugate prior for Bernoulli observations, so each correct/incorrect outcome updates the posterior analytically: $\alpha \leftarrow \alpha{+}1$ (correct) or $\beta \leftarrow \beta{+}1$ (incorrect).  The prior $\text{Beta}(5,5)$ for static knowledge encodes maximum uncertainty (mean $0.50$), reflecting that static knowledge is unverified external input whose reliability must be earned.  The prior $\text{Beta}(5,3)$ for dynamic experience encodes moderate confidence (mean $0.625$), reflecting that first-hand execution outcomes are inherently more trustworthy.

\textbf{Resolution Mechanism.}  The \texttt{ConflictResolver} auto-selects among four strategies: anomaly detection (outlier dynamic items), recency (stale static $>$30 days), evidence strength (sufficient data $\geq$3 observations), or Bayesian reliability (default, scoring $r_s \cdot \text{confidence}_s$ vs $r_d \cdot \text{confidence}_d$).  After resolution, posteriors are updated: the winner's $\alpha$ increments; the loser's $\beta$ increments.  Thompson Sampling drives active exploration---random draws $\theta_s \sim \text{Beta}(\alpha_s, \beta_s)$, $\theta_d \sim \text{Beta}(\alpha_d, \beta_d)$ balance exploitation with re-verification of potentially stale static knowledge.  Full algorithms are in Appendix~\ref{app:agent_capabilities}.

\textbf{Theorem 4.1 (Stationary-Segment Regret Bound).} \textit{Under locally stationary Bernoulli rewards for source reliability within a segment, Thompson Sampling has expected regret $O(\sqrt{T \log T})$ over $T$ exploration decisions \citep{agrawal2012}, compared to $O(T)$ for naive heuristics.}  This is a local bound: PRECEPT does not claim the same global regret guarantee across non-stationary drift segments.

\subsection{Type II: Rule Drift Adaptation}

\textbf{Rule Drift Parameters (\texttt{precept\_mcp\_server.py}).}

\begin{table}[htbp]
\centering
\footnotesize
\resizebox{\textwidth}{!}{%
\begin{tabular}{l|c|l}
\toprule
Parameter & Value & Operation \\
\midrule
\texttt{UNLEARN\_FAILURE\_THRESHOLD} & 2 & Invalidate after N consecutive failures \\
Confidence decay ($\delta$) & $\times$0.5 & \texttt{confidence *= 0.5} per failure \\
Confidence restore & +0.25 & \texttt{confidence = min(1.0, c + 0.25)} on success \\
\bottomrule
\end{tabular}
}%
\caption{Rule invalidation hyperparameters used for Type~II drift adaptation.}
\end{table}

\begin{figure}[t]
\centering
\resizebox{0.92\textwidth}{!}{%
\begin{tikzpicture}[
    >=stealth,
    node distance=1.2cm and 1.8cm,
    every node/.style={font=\small},
    process/.style={rectangle, rounded corners=4pt, draw=#1!70, fill=#1!10,
        minimum width=2.8cm, minimum height=0.9cm, align=center, line width=0.8pt,
        drop shadow={shadow xshift=0.5pt, shadow yshift=-0.5pt, opacity=0.12}},
    decision/.style={diamond, draw=#1!70, fill=#1!10, aspect=1.6,
        minimum width=2cm, inner sep=2pt, align=center, line width=0.8pt},
    arrow/.style={->, thick, color=black!65},
    greenarrow/.style={->, thick, color=green!60!black},
    redarrow/.style={->, thick, color=red!70!black},
    label/.style={font=\scriptsize, color=black!55},
    update/.style={rectangle, rounded corners=3pt, draw=#1!60, fill=#1!8,
        minimum width=2.4cm, minimum height=0.75cm, align=center, font=\small},
]

\node[process=blue] (apply) {\textbf{Apply Rule}\\[-1pt]\scriptsize\texttt{evaluate\_action()}};

\node[decision=gray, right=2cm of apply] (ok) {\textbf{OK?}};
\draw[arrow] (apply) -- (ok);

\node[update=green!60!black, above right=0.8cm and 2cm of ok] (success) {
    $f \leftarrow 0$\\[-2pt]
    \scriptsize $c \mathrel{+}= 0.25$
};
\draw[greenarrow] (ok) -- node[above left, label] {\textbf{\color{green!50!black}\checkmark} Pass} (success);

\node[update=red!70!black, below right=0.8cm and 2cm of ok] (fail) {
    $f \mathrel{+}= 1$\\[-2pt]
    \scriptsize $c \mathrel{\times}= 0.5$
};
\draw[redarrow] (ok) -- node[below left, label] {\textbf{\color{red!70!black}\texttimes} Fail} (fail);

\node[decision=orange, right=2cm of fail] (thresh) {$f \geq \theta$?};
\draw[redarrow] (fail) -- (thresh);

\node[process=red, above right=0.8cm and 1.8cm of thresh] (invalidate) {
    \textbf{Invalidate Rule}\\[-1pt]
    \scriptsize\texttt{record\_rule\_failure()}
};
\draw[redarrow] (thresh) -- node[above left, label] {\textbf{Yes}} (invalidate);

\node[process=orange, below right=0.8cm and 1.8cm of thresh] (retry) {
    \textbf{Retry}\\[-1pt]
    \scriptsize Decayed confidence
};
\draw[arrow, color=orange!70!black] (thresh) -- node[below left, label] {\textbf{No}} (retry);

\draw[arrow, color=orange!70!black, dashed] (retry.south) -- ++(0,-0.6) -| (apply.south)
    node[pos=0.25, below, label] {Next attempt};

\draw[greenarrow, dashed] (success.north) -- ++(0,0.5) -| (apply.north)
    node[pos=0.25, above, label] {Reinforced rule};

\node[below=0.3cm of apply, font=\scriptsize, color=blue!60] {$\theta = 2$ (default)};

\end{tikzpicture}
}%
\caption{Evo-Memory lifecycle for Type~II (rule drift) handling. On success, the failure counter resets and confidence increases; on failure, confidence decays by half ($c \times 0.5$) and the failure counter increments. When $f \geq \theta$ ($\theta{=}2$ by default), the rule is invalidated via \texttt{record\_rule\_failure()}, triggering re-learning. This yields stale-rule persistence probability $(1{-}d)^\theta \leq 0.0025$ for PRECEPT (Corollary~6.7).}
\label{fig:evo_memory}
\end{figure}
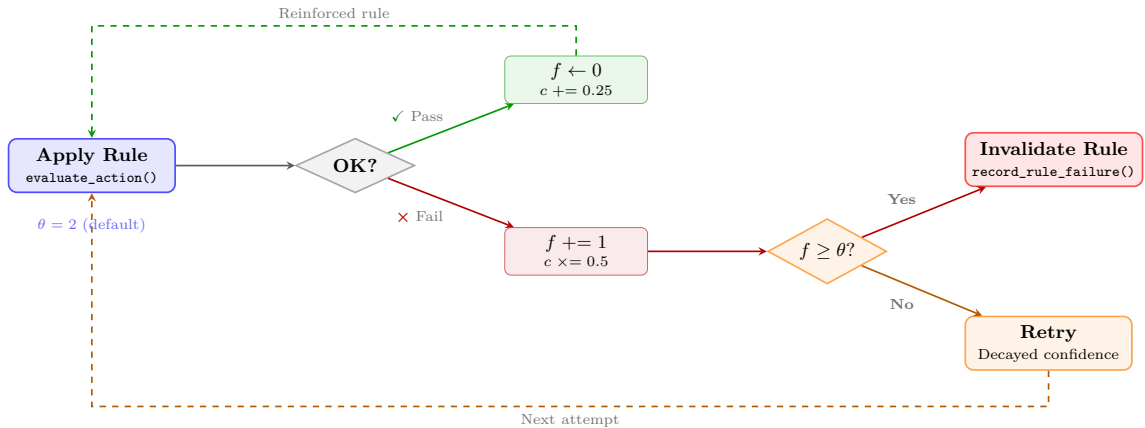

Figure~\ref{fig:evo_memory} depicts the Evo-Memory lifecycle: how partial progress and failed options accumulate across episodes, feeding back into the RefineInterceptor's constraint set.

\subsection{Rule Invalidation Algorithm}

Algorithm~\ref{alg:threshold_based_rule_invalidat} specifies the threshold-based rule invalidation mechanism.  When a learned rule produces an incorrect outcome, the failure count for that condition key is incremented and its confidence is halved (soft decay).  Once the failure count reaches the threshold $\theta{=}2$, the rule is permanently deleted from the rule store and the deletion is persisted to disk.  Conversely, a successful application resets the failure counter and restores confidence by $+0.25$ (capped at $1.0$), preventing transient failures from triggering premature invalidation.

\begin{algorithm}[htbp]
\caption{Threshold-Based Rule Invalidation}
\label{alg:threshold_based_rule_invalidat}
\begin{lstlisting}[style=pseudocode]
Constants:
UNLEARN_FAILURE_THRESHOLD = 2
Global State:
learned_rules: Dict[str, str]
rule_failure_counts: Dict[str, int]
rule_confidence: Dict[str, float]
\Function{record_rule_failure}{condition_key}
global learned_rules, rule_failure_counts, rule_confidence
// Only track for existing rules
if condition_key not in learned_rules then
    return None
end if
// Increment failure count
rule_failure_counts[condition_key] <- rule_failure_counts.get(condition_key, 0 ...
current_failures <- rule_failure_counts[condition_key]
// Soft decay: multiply by 0.5
current_conf <- rule_confidence.get(condition_key, 1.0)
new_conf <- current_conf x 0.5
rule_confidence[condition_key] <- new_conf
// Check threshold
if current_failures >= UNLEARN_FAILURE_THRESHOLD then
    old_rule <- learned_rules.pop(condition_key, None)
    rule_failure_counts.pop(condition_key, None)
    rule_confidence.pop(condition_key, None)
    save_rules() // Persist deletion
    return "Rule invalidated"
end if
return None
end function
\Function{record_rule_success}{condition_key}
global rule_failure_counts, rule_confidence
// Reset failure count
if condition_key in rule_failure_counts then
    rule_failure_counts.pop(condition_key, 0)
end if
// Restore confidence: add 0.25, cap at 1.0
if condition_key in rule_confidence then
    old_conf <- rule_confidence[condition_key]
    rule_confidence[condition_key] <- min(1.0, old_conf + 0.25)
end if
end function
\end{lstlisting}
\end{algorithm}

\textbf{Prompt self-sanitization.}  A critical design property ensures that invalidated rules are also removed from COMPASS-evolved instructions.  The system prompt is never a static artifact.  On every call to \texttt{get\_evolved\_prompt(include\_rules=True)}, the evolved base prompt is rebuilt and the \textit{current} \texttt{learned\_rules} dict is appended dynamically.  When \texttt{record\_rule\_failure()} deletes an invalidated rule from \texttt{learned\_rules} (line~543 of Algorithm~\ref{alg:threshold_based_rule_invalidat}), the next \texttt{refresh\_evolved\_prompt()} call on the client automatically produces a prompt without the stale rule; no separate sanitization step is required.  This self-sanitizing architecture ensures that drift-invalidated knowledge is purged from both the rule store and the active instructions simultaneously.

\subsection{Smart Pivot Error Recovery}

When initial execution fails, PRECEPT enters a bounded retry loop (\texttt{max\_retries} pivots) with five key properties: (1)~\textbf{deterministic pruning}---in the default pruning configuration (\texttt{enable\_random\_fallback=False}, \texttt{soft\_constraints\_retriable=False}), the \texttt{RefineInterceptor} guarantees failed options are never retried; (2)~\textbf{cross-episode memory}---\texttt{context.failed\_options} persists across episodes for resumable exploration; (3)~\textbf{validation filter}---all LLM suggestions are validated against domain-valid options before execution, falling back to remaining untried options when necessary; (4)~\textbf{COMPASS integration}---error evaluation classifies constraint tiers and may trigger epistemic probes or block actions; (5)~\textbf{learning on recovery}---successful recoveries are stored as rules, procedures, and atomic precepts.  The full algorithm is detailed in Appendix~\ref{app:agent_capabilities}.

\subsection{Deterministic Pruning via Constraint Classification}

The Digital Red Queen (DRQ) identifies ``cyclic dynamics'' as a fundamental failure mode in static optimization.  PRECEPT eliminates this through the \texttt{RefineInterceptor}, which classifies errors into three constraint types---\textbf{HARD} (physical/logical impossibility: closures, strikes), \textbf{SOFT} (configurable: busy, congestion), and \textbf{TRANSIENT} (retryable: timeouts)---and maintains a forbidden set with $O(1)$ lookup.

\textbf{Theorem 4.5 (Elimination of Cyclic Failures).} \textit{Let $F = \{f_1, \ldots, f_k\}$ be the set of failed options. Under default pruning mode (\texttt{enable\_random\_fallback=False}, \texttt{soft\_constraints\_retriable=False}), the RefineInterceptor guarantees:}
$$P(\text{select}(f_i) \mid \text{RefineInterceptor}(F)) = 0, \quad \forall f_i \in F$$

\textit{Proof:} \texttt{is\_forbidden()} returns True for any option in the accumulated HARD and SOFT constraint sets; the inline validation-and-filter stage in \texttt{PRECEPTAgent.run\_task()} removes invalid or forbidden options before execution; with random fallback disabled, forbidden options are never reintroduced after exhaustion. $\blacksquare$

\textbf{Corollary 4.5.1.}  The RefineInterceptor serves the same mathematical function as DRQ's opponent history---both prevent revisiting previously-failed states, enabling monotonic progress.

The HARD/SOFT distinction enables guaranteed termination (finite option space + HARD pruning $\to$ bounded exploration), configurable exhaustiveness, and diagnostic probe triggering when all options are exhausted.  The full constraint classification algorithm is in Appendix~\ref{app:agent_capabilities}.

\section{COMPASS: Complexity-Optimized Multi-strategy Pareto Adaptive Search}

Sections~2--4 established PRECEPT's core learning pipeline: deterministic retrieval, compositional stacking, conflict resolution, and drift adaptation.  These mechanisms govern \textit{what} PRECEPT learns and \textit{how} it resolves knowledge conflicts.  COMPASS addresses the complementary question of \textit{how the agent's prompt evolves} to better leverage this knowledge.  Specifically, COMPASS controls the outer loop of prompt optimization---selecting which system prompts most effectively utilize PRECEPT's learned rules, while PRECEPT's inner loop handles per-task retrieval and adaptation.  The two systems are tightly coupled: COMPASS evaluates candidate prompts by executing them through PRECEPT's full pipeline (including retrieval, conflict resolution, and pruning), ensuring that prompt evolution is grounded in the same deterministic guarantees that govern task execution.

COMPASS should be understood as a \textit{dual-frequency} control layer rather than a single monolithic module.  Its \textbf{high-frequency} path operates at every step through lightweight action/error evaluation and pattern monitoring; because this path is active in the full PRECEPT runtime, it is exercised throughout the main experiments (Experiments~1--7).  Its \textbf{low-frequency} path performs heavier event-triggered prompt evolution and rollout shaping, and is isolated more directly in Experiments~8--9.  The evidence therefore supports a differentiated claim: high-frequency COMPASS is broadly validated as part of PRECEPT's core execution loop, while low-frequency COMPASS shows more regime-dependent gains, clearest in the OOD semantic setting.

\subsection{Architecture Overview}

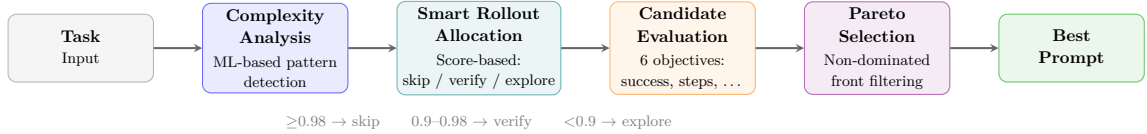
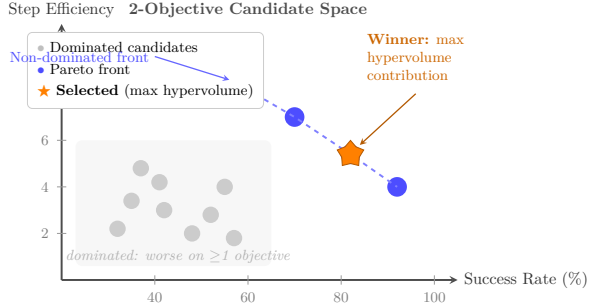
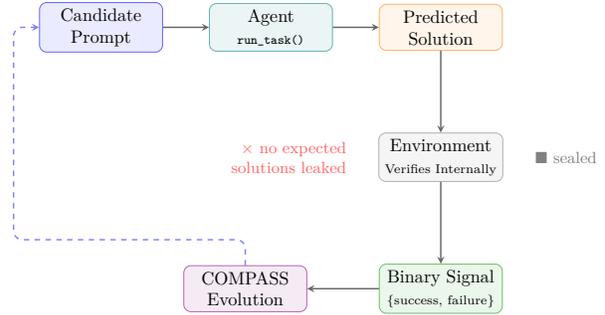
\begin{figure*}[t]
\centering

\begin{subfigure}[b]{\textwidth}
\centering
\resizebox{0.92\textwidth}{!}{%
\begin{tikzpicture}[
    >=stealth,
    node distance=0.6cm and 1cm,
    every node/.style={font=\normalsize},
    stage/.style={rectangle, rounded corners=5pt, draw=#1!60, fill=#1!8,
        minimum width=3cm, minimum height=1.4cm, align=center,
        drop shadow={shadow xshift=0.4pt, shadow yshift=-0.4pt, opacity=0.12}},
    arrow/.style={->, very thick, color=black!60},
    annot/.style={font=\small, color=black!50, align=center},
]

\node[stage=gray] (task) {\textbf{Task}\\{\small Input}};
\node[stage=blue, right=of task] (complex) {\textbf{Complexity}\\[1pt]\textbf{Analysis}\\[2pt]{\small ML-based pattern}\\{\small detection}};
\node[stage=teal, right=of complex] (rollout) {\textbf{Smart Rollout}\\[1pt]\textbf{Allocation}\\[2pt]{\small Score-based:}\\{\small skip / verify / explore}};
\node[stage=orange, right=of rollout] (eval) {\textbf{Candidate}\\[1pt]\textbf{Evaluation}\\[2pt]{\small 6 objectives:}\\{\small success, steps, \ldots}};
\node[stage=violet, right=of eval] (pareto) {\textbf{Pareto}\\[1pt]\textbf{Selection}\\[2pt]{\small Non-dominated}\\{\small front filtering}};
\node[stage=green!60!black, right=of pareto] (best) {\textbf{Best}\\[1pt]\textbf{Prompt}};

\draw[arrow] (task) -- (complex);
\draw[arrow] (complex) -- (rollout);
\draw[arrow] (rollout) -- (eval);
\draw[arrow] (eval) -- (pareto);
\draw[arrow] (pareto) -- (best);

\node[annot, below=0.3cm of rollout] {$\geq$0.98 $\to$ skip\qquad 0.9--0.98 $\to$ verify\qquad $<$0.9 $\to$ explore};

\end{tikzpicture}
}%
\caption{COMPASS pipeline: task complexity determines rollout allocation; candidates are evaluated on multiple objectives and selected via Pareto optimality.}
\end{subfigure}

\vspace{1.2em}

\begin{subfigure}[b]{0.48\textwidth}
\centering
\resizebox{\textwidth}{!}{%
\begin{tikzpicture}[
    >=stealth,
    every node/.style={font=\normalsize},
]

\node[font=\normalsize\bfseries, black!70] at (4,5.8) {2-Objective Candidate Space};

\draw[->, very thick, black!70] (0,0) -- (8.5,0)
    node[right, font=\normalsize, black!80] {Success Rate (\%)};
\draw[->, very thick, black!70] (0,0) -- (0,5.5)
    node[above, font=\normalsize, black!80] {Step Efficiency};

\foreach \x/\xl in {2/40, 4/60, 6/80, 8/100} {
    \draw[black!40] (\x,-0.1) -- (\x,0.1);
    \node[below, font=\small, black!50] at (\x,-0.15) {\xl};
}
\foreach \y/\yl in {1/2, 2/4, 3/6, 4/8} {
    \draw[black!40] (-0.1,\y) -- (0.1,\y);
    \node[left, font=\small, black!50] at (-0.15,\y) {\yl};
}

\fill[gray!6, rounded corners=4pt] (0.3,0.3) rectangle (4.5,3.0);

\fill[gray!40] (1.2,1.1) circle (5pt);
\fill[gray!40] (2.2,1.5) circle (5pt);
\fill[gray!40] (2.8,1.0) circle (5pt);
\fill[gray!40] (1.7,2.4) circle (5pt);
\fill[gray!40] (3.2,1.4) circle (5pt);
\fill[gray!40] (3.7,0.9) circle (5pt);
\fill[gray!40] (2.1,2.1) circle (5pt);
\fill[gray!40] (1.5,1.7) circle (5pt);
\fill[gray!40] (3.5,2.0) circle (5pt);

\fill[blue!70] (3.8,4.2) circle (6pt);
\fill[blue!70] (5.0,3.5) circle (6pt);
\fill[blue!70] (6.2,2.7) circle (6pt);
\fill[blue!70] (7.2,2.0) circle (6pt);

\draw[blue!50, very thick, dashed] (3.8,4.2) -- (5.0,3.5) -- (6.2,2.7) -- (7.2,2.0);

\node[star, star points=5, fill=orange, draw=orange!70!black,
    inner sep=4pt, minimum size=10pt] (selected) at (6.2,2.7) {};

\node[draw=black!20, fill=white, rounded corners=3pt,
    inner sep=6pt, font=\small, align=left] at (1.8,4.5) {%
    \textcolor{gray!50}{$\bullet$}~Dominated candidates\\[2pt]
    \textcolor{blue!70}{$\bullet$}~Pareto front\\[2pt]
    \textcolor{orange}{$\bigstar$}~\textbf{Selected} (max hypervolume)
};

\node[font=\small, gray!60, align=center] at (2.5,0.5) {\textit{dominated: worse on $\geq$1 objective}};

\draw[blue!60, thick, ->] (2.0,4.8) -- (3.6,4.3);
\node[font=\small, blue!70, anchor=east] at (2.0,4.8) {Non-dominated front};

\draw[orange!70!black, thick, ->] (7.6,4.0) -- (6.4,2.9);
\node[font=\small, orange!80!black, align=left] at (7.6,4.8)
    {\textbf{Winner:} max\\hypervolume\\contribution};

\end{tikzpicture}
}%
\caption{Pareto-optimal selection in the 2D objective space (success rate vs.\ step efficiency). Gray dots are dominated candidates (worse on $\geq$1 objective); blue dots form the Pareto front; the orange star is the winner, chosen by maximum hypervolume contribution.}
\end{subfigure}
\hfill
\begin{subfigure}[b]{0.48\textwidth}
\centering
\resizebox{\textwidth}{!}{%
\begin{tikzpicture}[
    >=stealth,
    node distance=0.6cm and 0.9cm,
    every node/.style={font=\normalsize},
    box/.style={rectangle, rounded corners=4pt, draw=#1!60, fill=#1!8,
        minimum width=2.4cm, minimum height=0.85cm, align=center},
    arrow/.style={->, thick, color=black!60},
    annot/.style={font=\small, color=black!50},
]

\node[box=blue] (prompt) {Candidate\\Prompt};
\node[box=teal, right=of prompt] (agent) {Agent\\{\scriptsize\texttt{run\_task()}}};
\node[box=orange, right=of agent] (pred) {Predicted\\Solution};

\draw[arrow] (prompt) -- (agent);
\draw[arrow] (agent) -- (pred);

\node[box=gray, below=1.6cm of pred] (env) {Environment\\{\scriptsize Verifies Internally}};
\draw[arrow] (pred) -- (env);

\node[annot, right=0.5cm of env] {$\blacksquare$ sealed};
\node[annot, red!60, left=0.5cm of env, align=right] {$\times$ no expected\\solutions leaked};

\node[box=green!60!black, below=1.6cm of env] (signal) {Binary Signal\\{\scriptsize\{success, failure\}}};
\draw[arrow] (env) -- (signal);

\node[box=violet, left=1.4cm of signal] (compass) {COMPASS\\Evolution};
\draw[arrow] (signal) -- (compass);

\draw[arrow, dashed, blue!50, rounded corners=5pt]
    (compass.north) -- ++(0,0.5) -| ([xshift=-0.5cm]prompt.west) -- (prompt.west);

\end{tikzpicture}
}%
\caption{Verified evolution: the environment verifies solutions internally; only binary success/failure signals drive evolution---no expected solutions are exposed.}
\end{subfigure}

\caption{\textbf{COMPASS: Complexity-Optimized Multi-strategy Pareto Adaptive Search.} (a)~End-to-end pipeline from task complexity analysis through Pareto-optimal prompt selection. (b)~Pareto selection mechanism: candidates are evaluated on two objectives (success rate, step efficiency); dominated candidates (gray) are filtered, and the winner (orange star) is chosen from the non-dominated Pareto front (blue) by maximum hypervolume contribution. (c)~Verified prompt evolution: the agent predicts solutions, the environment verifies internally, and only binary signals drive evolution---no expected solutions are ever exposed to the agent.}
\label{fig:compass}
\end{figure*}

Figure~\ref{fig:compass} presents the three-part COMPASS architecture: (a)~ML-based complexity analysis with smart rollout allocation, (b)~bi-objective Pareto selection over success rate and step efficiency, and (c)~verified prompt evolution through real agent execution.

\textbf{COMPASS Component Summary.}

\begin{table}[htbp]
\centering
\footnotesize
\resizebox{\textwidth}{!}{%
\begin{tabular}{l|l|l}
\toprule
Component & Class & Function \\
\midrule
Complexity Analyzer & \texttt{PRECEPTComplexityAnalyzer} & Estimates tool/reasoning/retrieval steps via pattern detection \\
Smart Rollout & \texttt{SmartRolloutStrategy} & Allocates rollouts based on score (0.98$\rightarrow$skip, 0.9$\rightarrow$verify, else$\rightarrow$explore) \\
Pareto Selection & \texttt{pareto\_select()} & Selects from non-dominated front by hypervolume contribution \\
\bottomrule
\end{tabular}
}%
\caption{COMPASS components and their corresponding implementation classes.}
\end{table}

\subsection{Theoretical Foundation: MAP-Elites for Strategy Diversity}

COMPASS integrates the \textbf{MAP-Elites} principle (Mouret \& Clune, 2015) to prevent the ''convergence collapse'' observed in static optimization. The Digital Red Queen (DRQ) paper demonstrates that agents evolved through static optimization become ''less behaviorally diverse across independent runs,'' collapsing toward a single strategy that fails when blocked.

\textbf{Definition 5.1 (Strategy Diversity via MAP-Elites).} \textit{MAP-Elites maintains a grid of strategy niches, each containing the best-performing strategy for a behavioral phenotype. COMPASS implements this through:}

\begin{enumerate}
  \item \textbf{Diversity Threshold} (\texttt{diversity\_threshold = 0.7}): Minimum diversity score to consider strategies sufficiently distinct
  \item \textbf{Diversity Rollouts} (\texttt{diversity\_rollouts = 5}): Extra rollouts allocated when diversity is low
  \item \textbf{Behavioral Phenotyping}: Strategies characterized by \texttt{dominant\_dimension} (tool\_use, retrieval, reasoning, verification)
\end{enumerate}

\textbf{Definition 5.2 (Topological Distinctness).} \textit{Two strategies S$_1$, S$_2$ are topologically distinct if:}
$$\text{FailureModes}(S_1) \cap \text{FailureModes}(S_2) = \emptyset$$

\textit{COMPASS maintains topologically distinct alternatives via Pareto selection across multiple objectives---if the primary strategy fails, the Pareto front contains alternatives that succeeded on different objective combinations.}

\textbf{Algorithm:} When \texttt{diversity\_score} $<$ \texttt{diversity\_threshold}, COMPASS allocates \texttt{diversity\_rollouts} additional evaluations to discover behaviorally distinct candidates:

\begin{lstlisting}
if diversity_score < diversity_threshold then
    num_rollouts <- diversity_rollouts  // Explore for diverse strategies
    focus <- "diversity"
\end{lstlisting}

This prevents the ''greedy convergence'' that traps agents relying on single-strategy optimization, ensuring PRECEPT has pre-calculated alternatives when the obvious path is blocked.

\subsection{Complexity Analysis}

COMPASS employs a pattern-based complexity analyzer (\texttt{PRECEPTComplexityAnalyzer}) that estimates task difficulty across four dimensions: \textbf{tool usage} (estimated tool chain length weighted by detected tool patterns), \textbf{retrieval} (entity extraction and relationship counting to estimate retrieval hops), \textbf{reasoning} (pattern detection for multi-step inference), and \textbf{verification} (presence of validation requirements). Each dimension receives a weighted score; the dominant dimension and aggregate step estimate determine the task's \texttt{ComplexityEstimate}. An optional ML-based confidence override and history-based adjustment refine predictions for previously-seen domains. This lightweight analysis ($O(|task|)$ in task text length) feeds directly into the smart rollout allocation described next.

\subsection{Smart Rollout Allocation}

Given the complexity estimate, COMPASS dynamically allocates rollouts via a tiered decision function: near-perfect scores ($\geq 0.98$) trigger early stopping; high scores prompt diversity or consistency checks; otherwise, the number of rollouts scales with estimated task complexity, with additional allocation for recovery after failed attempts.  This yields a reduction in rollouts compared to a uniform baseline while preserving solution quality.

\subsection{Pareto-Optimal Selection}

After rollouts produce candidate prompts, the evaluated COMPASS \textit{compilation} path selects the best using bi-objective Pareto optimality over \emph{task success rate} and \emph{step efficiency} (Figure~\ref{fig:compass}b).  Candidate $A$ \textit{dominates} candidate $B$ iff $A$ is at least as good on \emph{both} objectives and strictly better on at least one.  Non-dominated candidates form the Pareto front.  From this front, the compilation winner is selected by weighted priority: $0.7 \times \text{success\_rate} + 0.3 \times \text{step\_efficiency}$, prioritizing task success as the primary objective while still rewarding efficiency.  Repeated candidate evaluations are cached to avoid redundant rollouts across compilation cycles.

\subsection{Multi-Objective Scoring and Verified Evolution}

All COMPASS scores are derived empirically from actual task execution, not heuristics.  Each rollout produces execution metrics aggregated into two primary objectives: \textbf{task success rate} ($\sum \mathbb{1}[\text{success}_i] / n$) and \textbf{step efficiency} ($1 / (1 + \bar{s}/s_{\max})$, where $s_{\max} = 1 + \text{MAX\_RETRIES}$).  Pareto selection over these objectives ensures strategies are both effective and efficient.

\textbf{Verified prompt evolution} (Figure~\ref{fig:compass}c) uses real agent execution signals rather than heuristic scoring: the agent executes tasks under a candidate prompt, the environment verifies against hidden ground truth, and only binary success/failure signals are returned.  The agent never sees expected solutions, ensuring honest feedback without solution leakage---unlike LLM-as-judge or keyword-matching approaches that provide biased or unreliable signals.

\textbf{GEPA evolution pipeline.}  Each trigger invokes a three-step server-side pipeline: \textbf{(1)~Reflective analysis}---the LLM diagnoses the failing trajectory, identifying root causes and suggested fixes (structured as \texttt{GEPAReflection}); \textbf{(2)~Prompt mutation}---the LLM generates an improved prompt variant incorporating the reflections and any learned rules (structured as \texttt{GEPAMutation}); \textbf{(3)~Pareto update}---the new candidate is checked against the existing Pareto front: if it is dominated, it is discarded; otherwise, it is added and any candidates it dominates are removed.  Parents for mutation are selected stochastically from the Pareto front, weighted by average score with exploration noise, ensuring diverse exploration rather than greedy convergence.

\textbf{Two-path prompt selection.}  PRECEPT maintains two complementary evolution paths.  The \textit{COMPASS compilation path} generates multiple candidates in a single batch, evaluates each via smart rollouts on validation tasks, and Pareto-selects the compilation winner using the 0.7/0.3 weighted rule above.  The \textit{GEPA Pareto front path} accumulates candidates incrementally across triggers, maintaining a diverse non-dominated set.  In the evaluated runtime, the server-side prompt retriever \texttt{get\_evolved\_prompt()} makes the final deployment choice as follows: if the compilation winner's score exceeds the acceptance threshold (\texttt{PRECEPT\_COMPASS\_MIN\_SCORE}=0.6), that compiled prompt becomes active; otherwise the system falls back to the GEPA Pareto front and selects the candidate with highest \texttt{task\_success\_rate}; the base prompt serves as the final fallback.  In both paths, the current learned rules are dynamically appended to the selected prompt, ensuring the active instructions always reflect the latest knowledge state.

\subsection{Dual-Frequency Control Loop}

\begin{figure}[t]
\centering
\resizebox{0.92\textwidth}{!}{%
\begin{tikzpicture}[
    >=stealth,
    node distance=0.6cm and 0.8cm,
    every node/.style={font=\small},
    step/.style={rectangle, rounded corners=3pt, draw=#1!60, fill=#1!8,
        minimum width=2.8cm, minimum height=0.85cm, align=center, line width=0.7pt,
        drop shadow={shadow xshift=0.4pt, shadow yshift=-0.4pt, opacity=0.10}},
    trigger/.style={rectangle, rounded corners=3pt, draw=violet!60, fill=violet!8,
        minimum width=2cm, minimum height=0.7cm, align=center, line width=0.7pt},
    groupbox/.style={rectangle, rounded corners=6pt, draw=#1!50, fill=#1!5,
        inner sep=10pt, line width=1pt},
    arrow/.style={->, thick, color=black!60},
    grouparrow/.style={->, very thick, color=black!50},
    grouplabel/.style={font=\small\bfseries, color=#1!70},
]

\node[step=green!60!black] (ea) {\texttt{evaluate\_action}\\[-2pt]\scriptsize Constraint check};
\node[step=green!60!black, right=0.9cm of ea] (ee) {\texttt{evaluate\_error}\\[-2pt]\scriptsize Pattern match};
\node[step=green!60!black, right=0.9cm of ee] (lp) {\texttt{learn\_pattern}\\[-2pt]\scriptsize Rule update};

\draw[arrow, color=green!50!black] (ea) -- (ee);
\draw[arrow, color=green!50!black] (ee) -- (lp);

\begin{scope}[on background layer]
\node[groupbox=green!60!black, fit=(ea)(ee)(lp), inner sep=12pt] (hfbox) {};
\end{scope}
\node[grouplabel=green!60!black, above=2pt of hfbox.north] {HIGH-FREQ MONITOR \scriptsize(every step, $O(1)$)};

\node[trigger, below=1.5cm of ea] (t1) {New Rule};
\node[trigger, right=0.6cm of t1] (t2) {Goal Fail};
\node[trigger, right=0.6cm of t2] (t3) {Phase Change};

\begin{scope}[on background layer]
\node[groupbox=violet, fit=(t1)(t2)(t3), inner sep=10pt] (trbox) {};
\end{scope}
\node[grouplabel=violet, above=2pt of trbox.north west, anchor=west] {TRIGGER EVENTS};

\draw[grouparrow] (hfbox.south) -- (trbox.north)
    node[midway, right=3pt, font=\scriptsize, color=black!50] {emits};

\node[step=orange!80!black, below=1.5cm of t1] (co) {\texttt{compile}\\[-2pt]\scriptsize Context build};
\node[step=orange!80!black, right=0.9cm of co] (pe) {\texttt{Prompt Evolve}\\[-2pt]\scriptsize GEPA mutation};
\node[step=orange!80!black, right=0.9cm of pe] (ps) {\texttt{Pareto Select}\\[-2pt]\scriptsize Bi-objective};
\node[step=orange!80!black, right=0.9cm of ps] (mp) {\texttt{Memory Prune}\\[-2pt]\scriptsize MAP-Elites};

\draw[arrow, color=orange!70!black] (co) -- (pe);
\draw[arrow, color=orange!70!black] (pe) -- (ps);
\draw[arrow, color=orange!70!black] (ps) -- (mp);

\begin{scope}[on background layer]
\node[groupbox=orange!80!black, fit=(co)(pe)(ps)(mp), inner sep=12pt] (lfbox) {};
\end{scope}
\node[grouplabel=orange!80!black, above=2pt of lfbox.north] {LOW-FREQ ARCHITECT \scriptsize(on trigger, $O(n)$)};

\draw[grouparrow] (trbox.south) -- (lfbox.north)
    node[midway, right=3pt, font=\scriptsize, color=black!50] {activates};

\node[font=\scriptsize, color=green!50!black, rotate=90, anchor=south] at ([xshift=-1.2cm]hfbox.west) {\textit{every agent step}};
\node[font=\scriptsize, color=orange!60!black, rotate=90, anchor=south] at ([xshift=-1.2cm]lfbox.west) {\textit{on trigger only}};

\end{tikzpicture}
}%
\caption{COMPASS dual-frequency control loop. The high-frequency monitor runs at every agent step with $O(1)$ cost for real-time constraint checking and pattern learning. When trigger events are emitted (new rule discovered, goal failure, phase change), the low-frequency architect activates for strategic re-planning: context compilation, GEPA-based prompt evolution, Pareto selection of candidates, and MAP-Elites-guided memory pruning.}
\label{fig:dual_freq}
\end{figure}
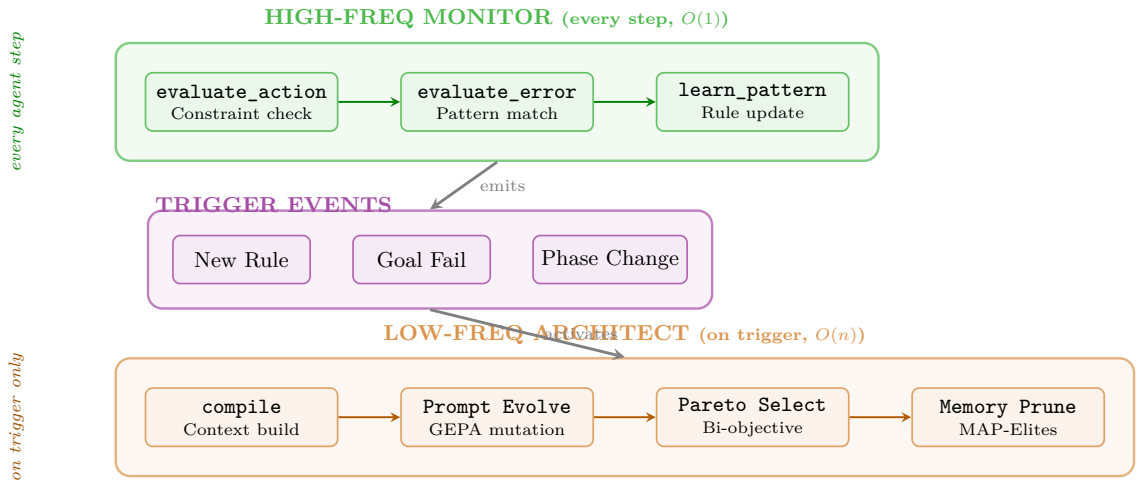

COMPASS separates lightweight per-step monitoring from heavyweight event-driven optimization, addressing the latency/cost concern of continuous optimization.  Figure~\ref{fig:dual_freq} illustrates this dual-frequency control loop.

\textbf{High-Frequency (Monitor Mode, every step, $O(1)$).}  Three lightweight functions run at every iteration.  \texttt{evaluate\_action()} checks for blocking constraints before execution (returning BLOCK, PIVOT, FAST\_PATH, or PROCEED).  \texttt{evaluate\_error()} classifies failures and triggers epistemic probes when appropriate.  \texttt{learn\_pattern()} updates pattern confidence.  All operate via $O(1)$ logical checks with sub-millisecond latency.

\textbf{Low-Frequency (Architect Mode, on trigger, $O(n)$).}  Heavyweight prompt evolution runs only on trigger events: accumulated patterns, exhausted retries, or phase transitions (training $\to$ testing).  The architect analyzes feedback patterns, generates prompt candidates, evaluates via smart rollouts, and selects via Pareto optimization.

This separation provides real-time constraint enforcement at every step while limiting expensive optimization to events where it yields measurable benefit---reducing rollout cost without weakening the constraint checks executed online.

Sections~2--5 described \textit{what} PRECEPT does.  We now establish \textit{why} it works: closed-form theoretical bounds that predict the empirical advantages validated in Section~7.

\section{Theoretical Analysis}\label{sec:theory}

\textit{For readability, the detailed proofs are collected in Appendix~\ref{app:theory}. Here we state the main bounds together with their modeling assumptions.}

PRECEPT's theoretical advantages are established through closed-form bounds across six dimensions. Table~\ref{tab:theory_summary} summarizes the main results.

\begin{table}[htbp]
\centering
\small
\begin{adjustbox}{max width=\textwidth}
\begin{tabular}{>{\raggedright\arraybackslash}p{2.8cm}|>{\raggedright\arraybackslash}p{4.9cm}|>{\raggedright\arraybackslash}p{1.7cm}|>{\raggedright\arraybackslash}p{4.3cm}}
\toprule
Metric & Formula / Result & Theorem & PRECEPT vs Baseline \\
\midrule
First-Try Success Rate $P_1$ & $\frac{W}{W+B} + \frac{B}{W+B} \cdot C(T,E,\beta) \cdot P_{\text{learn}}(R) \cdot \alpha$ & B.1 & Assumed anchors: $\alpha_P{=}0.85$, $\alpha_B{=}0.50$ (sensitivity in Appendix) \\
Multi-Condition Degradation & $\alpha_{\text{verbal}}(N) = \alpha_{\text{verbal}}(1) \cdot p^{N-1}$; $\alpha_{\text{PRECEPT}}(N) = \text{const}$ & B.2 & $22.6\times$ at $N=10$ (illustrative anchors) \\
Drift Resilience & $P(\text{stale persists}) \leq (1-d)^\theta$ & B.4, B.5 & $64\times$ in the local stationary-segment model \\
Partial Match Error & $1 - p^N - (1-p)^N$ vs $0$ (PRECEPT) & B.6 & 94.4\% vs \textbf{0\%} at $N=10$ (illustrative $p{=}0.75$) \\
Zero Retry Waste (default pruning mode) & $P(\text{retry\_failed}) = 0$ via RefineInterceptor & B.7 & Provable guarantee \\
Compositional Coverage & $2^N - 1$ from $N$ atomic precepts & 3.1 & Exponential \\
\bottomrule
\end{tabular}
\end{adjustbox}
\caption{Summary of theoretical bounds. Full proofs in Appendix~\ref{app:theory}.}
\label{tab:theory_summary}
\end{table}

The key insight is that PRECEPT's advantages \textbf{grow exponentially with task complexity}: at $N=1$ condition, the advantage is a modest $1.7\times$; at $N=10$ conditions, the illustrative bound in Corollary~B.3 reaches $22.6\times$. This follows from verbal baselines' exponential degradation ($p^{N-1}$ under the independence approximation, with illustrative per-condition accuracy $p \approx 0.75$) while PRECEPT's hash-based retrieval maintains constant effectiveness regardless of condition count. Similarly, under the local stationary-segment detection model of Corollary~B.5, drift resilience follows from the threshold-based invalidation mechanism: with illustrative detection rates $d=0.95$ and $\theta=2$, stale rules persist with probability $(0.05)^2 = 0.0025$ versus $(0.40)^2 = 0.16$ for verbal baselines---a $64\times$ ratio.  The experiments are intended to validate the qualitative trend and order of magnitude of these advantages, not to claim that these exact constants hold universally outside the modeled regime.

\section{Experiments}

The paper reports nine experiments in total.  Experiments~1--7 form the core validation of the three contributions: Experiments~1--3 establish the compositional rule learning advantage and its training requirements (Contribution~1); Experiments~4--5 demonstrate continuous learning and rule persistence during deployment; Experiment~6 validates Bayesian conflict resolution under adversarial static knowledge (Contribution~2, Type~I); and Experiment~7 tests drift adaptation under environment change (Contribution~2, Type~II).  COMPASS (Contribution~3) operates throughout as the prompt-evolution layer.  Experiments~8--9 provide targeted follow-up stress tests that isolate the low-frequency COMPASS outer loop under matched-key and OOD semantic regimes.  This structure ensures that each experiment stresses a specific aspect of the unified architecture while relying on the other components being functional.

\subsection{Experimental Setup}

\textbf{Domains.} Three domains with structured error patterns spanning a wide complexity range:

\begin{table}[htbp]
\centering
\footnotesize
\resizebox{\textwidth}{!}{%
\begin{tabular}{llccc}
\toprule
Domain & Description & $E$ (Unique Keys) & Options & $T_{\text{train}}$ ($\beta$=3) \\
\midrule
Integration & OAuth failures, API errors & 6 & 15 & 18 \\
Booking & Reservation failures, overbooking & 17 & 20 & 51 \\
Logistics & Port closures, customs delays & 4 & 4 & 12 \\
\bottomrule
\end{tabular}
}%
\caption{Evaluated domains and their core training-time complexity statistics.}
\end{table}

These domains span from low-complexity (Logistics: $E=4$, 4~options) to high-complexity (Booking: $E=17$, 20~options), testing PRECEPT across diverse error pattern structures. Experiment~1 evaluates all three domains; Experiments~2--6 focus on Integration and Logistics to provide consistent cross-experiment comparison across the two domains with the widest complexity contrast.

\subsubsection{Experimental Data Design: Black Swans in the Dark}
\label{sec:data_design}

Standard agent benchmarks (grid-worlds, mazes, AlfWorld) present environments where the action space is fully enumerable and the transition function is visible or quickly learnable.  We deliberately break this assumption by constructing three domains that function as \textbf{hidden constraint-satisfaction problems}: (1)~solution mappings are \textit{opaque by construction}---the correct answer is a deterministic keyed mapping $f(\texttt{MD5}(\texttt{condition\_key}))$ that is undeducible from conditions; MD5 is used here as an implementation primitive for deterministic opacity (not as a security claim), and any fixed seeded opaque mapping would serve equivalently; (2)~error signals are \textit{deliberately uninformative}---vague domain-specific codes that do not narrow the search; and (3)~conditions are \textit{compositional and overlapping}---different composite keys share components but map to different solutions, creating a \textit{majority-vote trap} where approximate retrieval develops confident but wrong generalizations.  This design is a stress test for exact retrieval under minimal semantic leakage; in domains where conditions and solutions are semantically aligned, approximate methods may perform better than on this benchmark.

\begin{table}[htbp]
\centering
\footnotesize
\resizebox{\textwidth}{!}{%
\begin{tabular}{lcccll}
\toprule
Domain & Options & $E$ & $P_{\text{rand}}$ & LLM Prior Bias & Primary Difficulty \\
\midrule
Logistics & 4 & 4 & 25\% & Low (geographic ports) & Majority-vote trap from component overlap \\
Integration & 2\textsuperscript{*} & 6 & 50\%\textsuperscript{*} & \textbf{High} (brand-name priors) & Adversarial naming + prior override \\
Booking & 2\textsuperscript{*} & 17 & 50\%\textsuperscript{*} & Low (IATA codes) & Largest key space + phantom inventory \\
\bottomrule
\end{tabular}
}
\caption{Benchmark difficulty summary across the three evaluated domains.}
\caption*{\footnotesize \textsuperscript{*}Multi-condition valid solutions only. Full option spaces: Integration~15, Booking~20.}
\end{table}

\textbf{Logistics} simulates global shipping with $E{=}4$ port-closure conditions and 4 valid alternatives.  The majority-vote trap is acute: composite keys share 2--3 conditions but map to different ports via MD5 hash, so component-level generalizations are actively misleading.  \textbf{Integration} simulates API orchestration with $E{=}6$ error conditions and only 2 valid solutions: \texttt{salesforce-backup} and \texttt{hubspot-v2}---suffixed variants of famous brands whose base names (\texttt{salesforce}, \texttt{hubspot}) are the \textit{wrong} answers.  The LLM's parametric prior overwhelms in-context guidance, creating the hardest domain.  \textbf{Booking} simulates flight reservations with $E{=}17$ conditions and 20 options (only 2--3 valid), testing scalability to large, sparse key spaces with deceptive success signals.  Together, the three domains ensure that no single retrieval shortcut succeeds across all settings; only exact composite-key $\to$ solution mappings retrieved deterministically---PRECEPT's core mechanism---can consistently navigate these dark mazes.

\textbf{Baselines (Enhanced).}  Both baselines are \textit{substantially improved} over their original implementations: Full Reflexion (enhanced from \citep{shinn2023}) adds vector database integration, condition-aware metadata pre-filtering, BM25+semantic hybrid retrieval via RRF, and structured prompts; ExpeL (enhanced from \citep{zhao2023}) adds condition-aware metadata filtering, BM25+semantic hybrid retrieval, structured insight extraction, and direct solution storage.  These enhancements substantially reduce obvious implementation confounds; the remaining gap is therefore most consistently explained by PRECEPT's \textbf{exact-vs-approximate retrieval boundary}, together with its additional architectural features---procedural memory, rule invalidation with confidence decay, the RefineInterceptor's deterministic pruning, and COMPASS prompt evolution---rather than by simple implementation disadvantage.

\textbf{Protocol.}
\begin{itemize}
  \item Independent runs: $N=10$ seeds
  \item Confidence intervals: 95\% (t-distribution)
  \item Statistical tests: Paired t-test, Bonferroni correction
  \item Effect sizes: Cohen's d
\end{itemize}

\textbf{Inference scope.}  Statistical tests are organized at the experiment level: each experiment defines a primary comparison aligned to its research question, and Bonferroni correction is applied within that experiment's reported family of comparisons.  Because the paper also reports secondary diagnostics across multiple encounters, domains, and ablations, we emphasize 95\% confidence intervals and effect sizes alongside $p$-values, and treat marginal COMPASS results (Experiments~8--9) as scoped evidence rather than universal gains.

\textbf{Cross-experiment protocol summary.}

\begin{table}[htbp]
\centering
\small
\begin{adjustbox}{max width=\textwidth}
\begin{tabular}{c|>{\raggedright\arraybackslash}p{2.8cm}|>{\raggedright\arraybackslash}p{2.5cm}|>{\raggedright\arraybackslash}p{5.2cm}|c}
\toprule
Exp. & Focus & Domains & Evaluation regime & Seeds \\
\midrule
1 & Main comparison & Integration, Booking, Logistics & Mixed matched+harder cases (\texttt{test-mode both}) & 10 / domain \\
2 & Compositional generalization & Integration, Logistics & Train on atomics; test 2-way and 3-way compositions & 10 / config \\
3 & Training-size ablation & Integration, Logistics & Matched multi-condition regime ($N{=}5$) across $\beta \in \{1,\dots,5\}$ & 10 / $\beta$ / domain \\
4 & Continuous learning & Integration, Logistics & Sequential encounters after minimal training ($\beta{=}1$) & 9--10 \\
5 & Rule persistence & Integration, Logistics & Restart with unchanged mapping ($s_0{=}s_1$) & 9--10 \\
6 & Static knowledge ablation & Integration, Logistics & With/without adversarial static knowledge & 10 / config \\
7 & Drift adaptation & Integration, Logistics & Train with $s_0$, test with shifted mapping $s_1$ & 10; final logistics encounter~4 uses 8 valid traces \\
8 & COMPASS matched ablation & Integration & Outer-loop ablation in matched-key regime & 10 \\
9 & COMPASS OOD semantic ablation & Integration & Outer-loop ablation in OOD semantic regime & 10 \\
\bottomrule
\end{tabular}
\end{adjustbox}
\caption{Cross-experiment protocol summary, clarifying which evaluation regime is used in each experiment.}
\end{table}

\textbf{Models.}  All experiments use \textbf{GPT-4o-mini} (\texttt{gpt-4o-mini}, OpenAI) as the LLM for all three agents (PRECEPT, Full Reflexion, ExpeL), with temperature $0.3$ and max tokens $200$.  Embedding-based retrieval (ChromaDB vector stores for both dynamic experience and static knowledge) uses \textbf{text-embedding-3-small} (OpenAI, 1536 dimensions).  Both models are held constant across all agents and all experiments, helping isolate architectural mechanisms from model-capacity differences.

\textbf{COMPASS Hyperparameters.}

\begin{table}[htbp]
\centering
\footnotesize
\begin{tabular}{l|c|l}
\toprule
Parameter & Value & Description \\
\midrule
Compilation candidates & 5 & Prompt variants generated per compilation cycle \\
Rollouts per candidate & 3 & Base rollouts (adjusted by smart strategy) \\
Acceptance threshold & 0.6 & Min score to adopt an evolved prompt \\
Pareto weights & 0.7 / 0.3 & success\_rate / step\_efficiency \\
Evolution interval & 2 tasks & Trigger frequency for lo-freq architect \\
Diversity threshold & 0.7 & Min diversity to skip diversity rollouts \\
Diversity rollouts & 5 & Extra rollouts when diversity is low \\
Early-stop score & 0.98 & Score above which rollouts are skipped \\
\bottomrule
\end{tabular}
\caption{COMPASS hyperparameters used in all reported experiments.}
\end{table}

\subsection{Experiment 1: Main Domain Comparison}

\textbf{Research Question:} How does PRECEPT compare against enhanced baselines across diverse structured-decision domains?

\textbf{Setup:} All 3 domains (Integration, Booking, Logistics), $N=5$ composite condition keys, $\beta=3$ (moderate training), max 4 retries, 10 seeds per domain. All agents use the enhanced baseline implementations described in \S{}7.1. The domains span a wide complexity range: from Logistics ($E=4$ unique condition keys, 4 options) to Booking ($E=17$ unique condition keys, 20 options), testing PRECEPT across diverse error pattern structures where inter-scenario condition overlap is substantial---the key architectural differentiator of exact vs.\ approximate retrieval.

\subsubsection{Primary Results}

\textbf{First-try success rate $P_1$ (mean $\pm$ 95\% CI, 10 seeds per domain). Domains ordered by PRECEPT advantage.}

\begin{table}[htbp]
\centering
\footnotesize
\resizebox{\textwidth}{!}{%
\begin{tabular}{l|l|l|l|l|l}
\toprule
Domain & E & Options & PRECEPT $P_1$ & FR $P_1$ & ExpeL $P_1$ \\
\midrule
Integration & 6 & 15 & \textbf{80.0\%}$\pm$12.3 & 36.7\%$\pm$7.5 & 43.3\%$\pm$8.3 \\
Booking & 17 & 20 & \textbf{94.1\%}$\pm$4.9 & 51.8\%$\pm$14.0 & 90.0\%$\pm$6.9 \\
Logistics & 4 & 4 & \textbf{95.0\%}$\pm$5.0 & 57.5\%$\pm$14.7 & 90.0\%$\pm$9.2 \\
\textbf{Average} &  &  & \textbf{89.7\%} & 48.6\% & 74.4\% \\
\bottomrule
\end{tabular}
}
\caption{Experiment~1 first-try success ($P_1$) across the three main domains.}
\end{table}

\textit{Statistical significance (paired t-test, Bonferroni corrected across 3 domains): PRECEPT vs FR --- Integration +43.3pp ($p_{\mathrm{corr}}=0.0005$, $d=1.93$), Booking +42.4pp ($p_{\mathrm{corr}}=0.0002$, $d=2.26$), Logistics +37.5pp ($p_{\mathrm{corr}}=0.0003$, $d=2.12$); all corrected $p<0.001$, all Cohen's $d>1.9$ (large effects). PRECEPT vs ExpeL --- Integration +36.7pp ($p_{\mathrm{corr}}=0.0001$, $d=2.39$); Booking +4.1pp ($p_{\mathrm{corr}}=1.000$, n.s.) and Logistics +5.0pp ($p_{\mathrm{corr}}=1.000$, n.s.) --- ExpeL approaches PRECEPT on these two domains.}

\textbf{Overall success rate $P_t$ and average steps per task (mean $\pm$ 95\% CI, 10 seeds).}

\begin{table}[htbp]
\centering
\footnotesize
\resizebox{\textwidth}{!}{%
\begin{tabular}{l|l|l|l|l|l|l}
\toprule
Domain & PRECEPT $P_t$ & FR $P_t$ & ExpeL $P_t$ & PRECEPT Steps & FR Steps & ExpeL Steps \\
\midrule
Integration & \textbf{83.3\%}$\pm$11.2 & 41.7\%$\pm$6.3 & 46.7\%$\pm$5.0 & \textbf{2.72}$\pm$0.46 & 7.83$\pm$0.58 & 6.43$\pm$0.53 \\
Booking & \textbf{99.4\%}$\pm$0.6 & 94.7\%$\pm$5.3 & 98.8\%$\pm$1.2 & \textbf{2.15}$\pm$0.12 & 4.60$\pm$0.86 & 2.40$\pm$0.29 \\
Logistics & \textbf{100.0\%}$\pm$0.0 & 80.0\%$\pm$11.3 & 97.5\%$\pm$2.5 & \textbf{2.10}$\pm$0.23 & 5.25$\pm$0.76 & 2.45$\pm$0.52 \\
\textbf{Average} & \textbf{94.2\%} & 72.1\% & 81.0\% & \textbf{2.32} & 5.89 & 3.76 \\
\bottomrule
\end{tabular}
}
\caption{Experiment~1 overall success ($P_t$) and step efficiency across the three main domains.}
\end{table}

\begin{figure}[htbp]
  \centering
  \includegraphics[width=\textwidth]{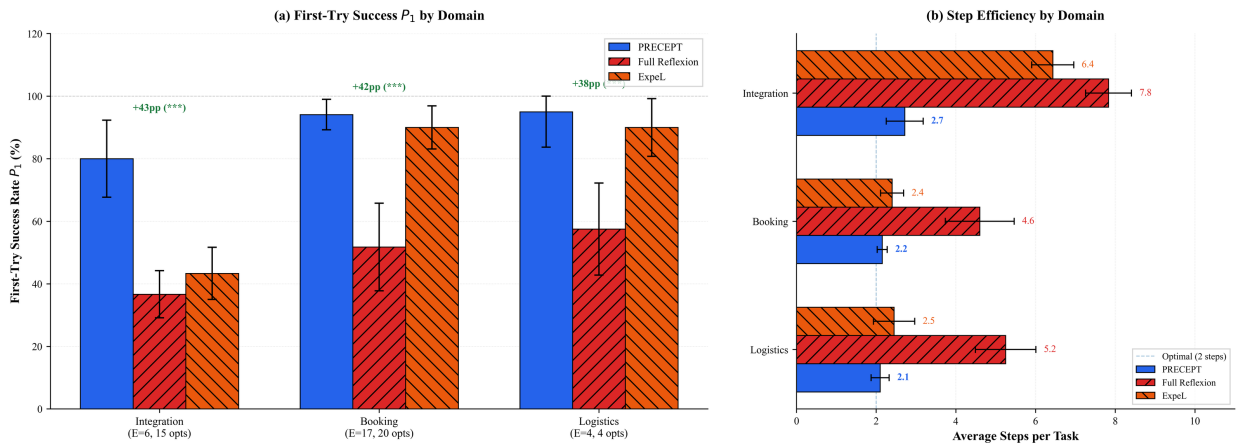}
  \caption{Experiment~1 main comparison across the three evaluated domains: first-try success and step efficiency.}
  \label{fig:domain_comparison}
\end{figure}

\subsubsection{Key Findings}

\textbf{(1) Decisive lead across all domains.}  Figure~\ref{fig:domain_comparison} compares $P_1$ and step efficiency across the three high-complexity domains.  Mean $P_1{=}$\textbf{89.7\%} vs FR 48.6\% (+41.1pp) and ExpeL 74.4\% (+15.3pp), with 2.32 avg steps vs FR 5.89 ($2.5\times$ overhead).  All three domains show corrected $p{<}0.001$ vs FR (Cohen's $d{=}1.93$--$2.26$).  Against ExpeL, the advantage is significant only on Integration (+36.7pp, $d{=}2.39$, $p_{\mathrm{corr}}{=}0.0001$) but not Booking or Logistics, where ExpeL approaches PRECEPT.

\textbf{(2) Advantage driven by condition overlap and LLM prior bias.}  Integration produces the largest gap (+43.3pp vs FR) despite fewer condition keys than Booking, because correct solutions are suffixed variants of famous brands (\texttt{salesforce-backup} vs \texttt{salesforce}), triggering LLM parametric priors that override learned guidance.  Booking (+42.4pp vs FR) has the largest solution space but atomic option names without brand-name ambiguity; ExpeL survives there ($P_1{=}90.0\%$) but collapses on Integration ($P_1{=}43.3\%$, $d{=}2.39$).

\textbf{(3) Near-optimal step efficiency.}  PRECEPT achieves 2.10--2.72 avg steps across all domains (theoretical minimum: 2.0).  FR requires 4.60--7.83 steps ($2.0$--$2.9\times$ overhead).  On Integration, FR \textit{permanently fails} $\sim$58\% of tasks even with retries ($P_t{=}41.7\%$ vs PRECEPT's 83.3\%).

\subsubsection{Analysis: Why Integration Baselines Fail}

Integration exhibits the most extreme baseline failure: FR achieves $P_1{=}36.7\%$, $P_t{=}41.7\%$; ExpeL achieves $P_1{=}43.3\%$, $P_t{=}46.7\%$---yet on Booking, FR recovers to $P_t{=}94.7\%$ and ExpeL reaches $P_1{=}90.0\%$.  Log trace analysis across all 10 seeds reveals two \textbf{compounding LLM limitations}:

\textbf{(1) LLM Prior Bias.}  Integration's correct solutions are \textit{suffixed variants} of famous brands, e.g.\ \texttt{salesforce-backup} and \texttt{hubspot-v2}.  The LLM's parametric prior overwhelms in-context guidance: FR selects \texttt{salesforce} on all 5 attempts for tasks requiring \texttt{salesforce-backup}.  ExpeL's chain-of-thought explicitly identifies the correct suffixed variant, yet the generated output reverts to the base brand name---consistent with known unfaithful reasoning findings (Turpin et al., 2024).  Booking's synthetic codes (\texttt{DL-123}) have no competing priors, rendering this failure mode absent.

\textbf{(2) LLM Insight Dismissal.}  Even when correct insights are retrieved and presented in context, the LLM frequently chooses not to apply them.  In a representative seed (42), 3 of 6 episodes fail because the LLM either ignores insights entirely (0/5 attempts applied) or abandons them after early attempts.  This is a fundamental reliability limitation: insight-augmented architectures require the LLM to not only \textit{retrieve} correct knowledge but also \textit{choose to use} it.

Both are architectural properties of LLM-based generation, not implementation artifacts.  On the exact-match path, PRECEPT avoids this failure mode: structured rules store exact $\texttt{condition\_key} \rightarrow \texttt{solution}$ mappings as programmatic data retrieved via $O(1)$ hash lookup, bypassing LLM interpretation and generation-stage interference.

\subsection{Experiment 2: Compositional Semantic Generalization}

\textbf{Setup.} Train on 1-condition (atomic) scenarios, test on 2--3 condition (composite) scenarios. Domains: Integration and Logistics, each evaluated in 2-way and 3-way configurations (4 configurations total). Training: $\beta{=}3$ repetitions over 8 semantic atoms per domain (24 training tasks per config). Testing: 10 composite tasks per configuration. Max 4 retries, $N{=}10$ seeds per configuration (40 runs total).

\begin{figure}[htbp]
  \centering
  \includegraphics[width=\textwidth]{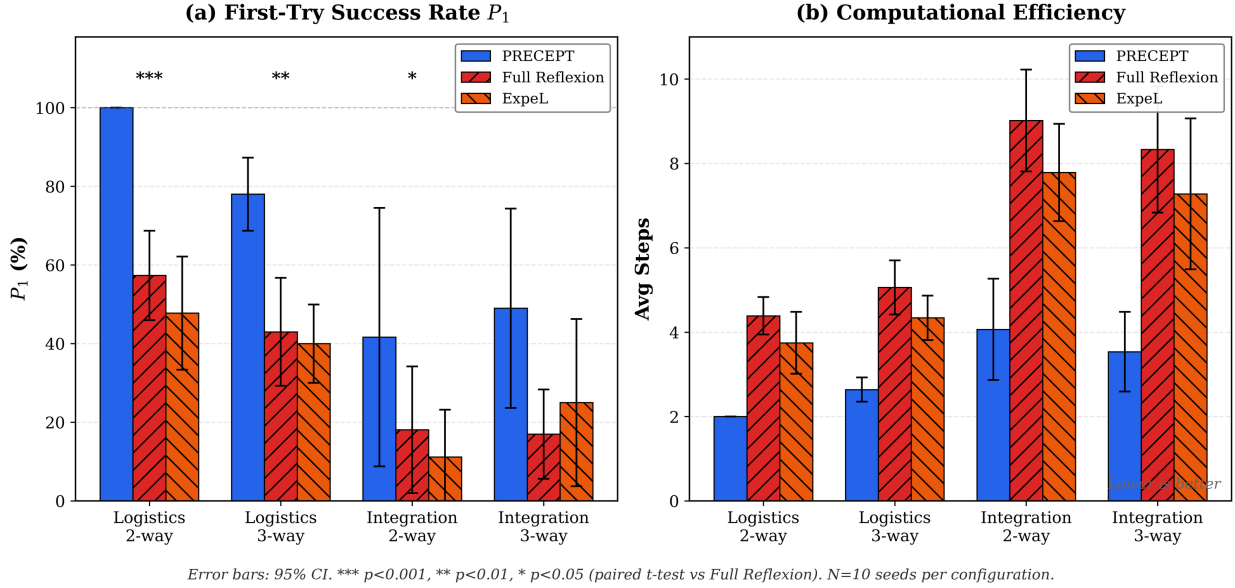}
  \caption{Experiment~2 compositional generalization across logistics and integration. PRECEPT reaches 100\% $P_1$ on 2-way logistics compositions and shows the strongest overall compositional performance.}
  \label{fig:exp2_compositional}
\end{figure}

\textbf{Compositional generalization results per domain ($N{=}10$ seeds per configuration). Overall: 4 configurations $\times$ 10 seeds $=$ 40 runs.  $P_1$: first-try success; $P_t$: eventual success after retries.  Significance: $^{*}p{<}0.05$, $^{**}p{<}0.01$, $^{***}p{<}0.001$.}

\begin{table}[htbp]
\centering
\footnotesize
\resizebox{\textwidth}{!}{%
\begin{tabular}{ll|ccc|ccc|c}
\toprule
Domain & Config & PRECEPT $P_1$ & FR $P_1$ & ExpeL $P_1$ & PRECEPT $P_t$ & FR $P_t$ & ExpeL $P_t$ & $d_{P_1}$ (vs FR) \\
\midrule
\multirow{2}{*}{Logistics} & 2-way & \textbf{100.0\%}$\pm$0.0 & 57.3\%$\pm$11.6 & 47.7\%$\pm$14.6 & \textbf{100.0\%}$\pm$0.0 & 97.8\%$\pm$3.4 & 96.0\%$\pm$5.0 & 2.64$^{***}$ \\
& 3-way & \textbf{78.0\%}$\pm$9.4 & 43.0\%$\pm$13.9 & 40.0\%$\pm$10.1 & \textbf{99.0\%}$\pm$2.3 & 97.0\%$\pm$3.5 & 93.0\%$\pm$5.9 & 1.31$^{**}$ \\
\midrule
\multirow{2}{*}{Integration} & 2-way & \textbf{41.7\%}$\pm$33.4 & 18.1\%$\pm$16.4 & 11.2\%$\pm$11.2 & \textbf{53.3\%}$\pm$32.7 & 32.4\%$\pm$16.9 & 42.4\%$\pm$20.7 & 0.84$^{*}$ \\
& 3-way & \textbf{49.0\%}$\pm$25.7 & 17.0\%$\pm$11.5 & 25.0\%$\pm$21.6 & \textbf{58.3\%}$\pm$30.1 & 42.0\%$\pm$24.7 & 38.7\%$\pm$23.9 & 0.90$^{*}$ \\
\midrule
\multicolumn{2}{l|}{\textbf{Overall}} & \textbf{67.2\%}$\pm$24.0 & 33.9\%$\pm$18.7 & 31.0\%$\pm$20.5 & \textbf{77.7\%}$\pm$22.7 & 67.3\%$\pm$16.9 & 67.5\%$\pm$19.4 & 1.55$^{***}$ \\
\bottomrule
\end{tabular}
}%
\caption{Experiment~2 compositional generalization results across logistics and integration.}
\end{table}

Figure~\ref{fig:exp2_compositional} visualizes the per-domain compositional results.  Overall $P_1$ effect sizes are large: $d{=}1.55$ vs FR ($p{<}0.001$) and $d{=}1.62$ vs ExpeL ($p{<}0.001$).  The largest per-configuration effect ($d{=}2.64$, logistics 2-way) reflects PRECEPT's 100\% $P_1$ vs $\sim$48--57\% for baselines; integration shows smaller but significant effects ($d{=}0.84$--$0.90$), reflecting its higher inherent difficulty.

\textbf{Retry recovery ($P_t$).}  PRECEPT's $P_t$ advantage narrows on logistics (100\% / 99\% vs baselines' 93--98\%) because the small option space ($|\mathcal{S}|{=}4$) allows brute-force recovery.  On integration, PRECEPT retains the lead ($P_t{=}53.3$--$58.3\%$ vs FR 32.4--42.0\% and ExpeL 38.7--42.4\%), confirming that even with retries, approximate retrieval cannot reliably recover on larger option spaces.  The $P_1 \to P_t$ gap itself is informative: PRECEPT's gap is small on logistics (0--21pp) and moderate on integration ($\sim$10pp), while baselines show large gaps on logistics ($\sim$40$+$pp), indicating heavy reliance on brute-force retries rather than first-try knowledge application.

\subsection{Experiment 3: Training Size Ablation (\texorpdfstring{$\beta$}{beta} Effect)}

\textbf{Setup.} Training exposure $\beta \in \{1,2,3,4,5\}$ (number of encounters per error type), with $T_{\text{train}} = \beta \times E$.  All baselines receive the same data, feedback, and retry budget (max 5 attempts).  Evaluated on logistics ($E{=}4$) and integration ($E{=}6$), $N{=}5$ conditions per composite key, 10 seeds per $\beta$ per domain (100 total runs).  We evaluate at $N{=}5$ rather than $N{=}1$ because single-condition settings mask the exact-vs-approximate retrieval distinction; component overlap at $N{\geq}2$ creates retrieval interference that differentiates $O(1)$ hash-based lookup from verbal-memory approaches (\S{}7.4.3).

\subsubsection{Primary Results: Multi-Condition (\texorpdfstring{$N=5$}{N=5})}

\begin{figure}[htbp]
  \centering
  \includegraphics[width=\textwidth]{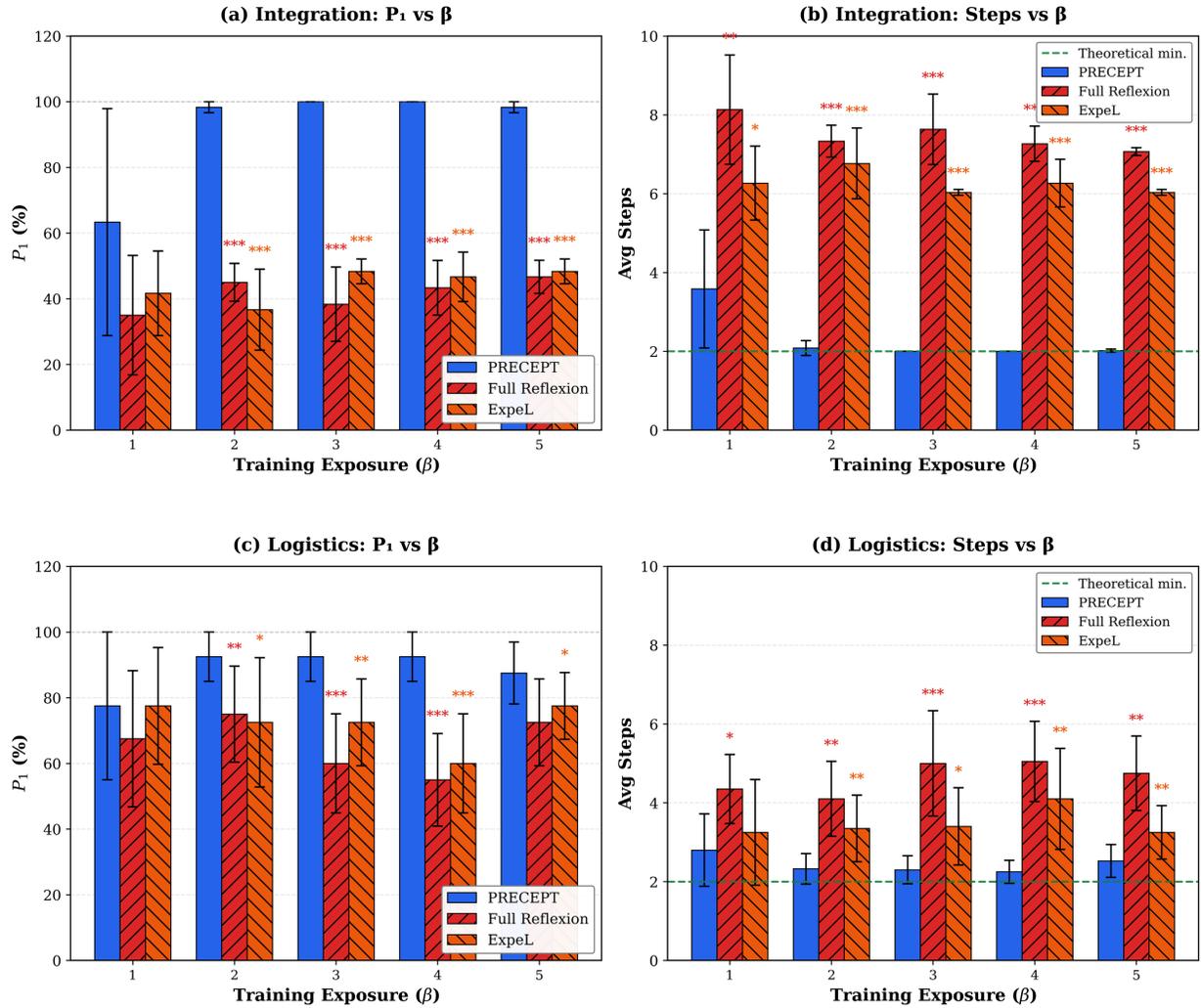}
  \caption{$P_1$ and Avg Steps vs Training Exposure ($\beta$ effect) across both domains. PRECEPT achieves 100\% $P_1$ on integration at $\beta{\geq}3$ and 87.5--92.5\% on logistics, with near-optimal 2.0-step efficiency.}
  \label{fig:training_exposure}
\end{figure}

A pilot study at $N{=}1$ (seed=42, Appendix~\ref{app:agent_capabilities}) confirmed the degenerate-case hypothesis: all agents achieve near-100\% $P_t$ and baselines match PRECEPT at $\beta{\geq}3$, validating our choice of $N{=}5$ for the primary evaluation.  Figure~\ref{fig:training_exposure} plots $P_1$ and average steps as a function of $\beta$.

\textit{Results from 10 independent seeds per $\beta$ value per domain (100 total runs across both domains). All values reported as mean $\pm$ 95\% CI.}

\begin{table}[htbp]
\centering
\caption{Primary Exp3 results ($N{=}5$, 10 seeds per $\beta$). Bold = best $P_1$/Steps. Significance (paired $t$-test): $^{*}p{<}0.05$, $^{**}p{<}0.01$, $^{***}p{<}0.001$. $P_1$ columns: markers on $d$ values. Steps columns: markers on baseline values indicate PRECEPT-vs-baseline step significance.}
\label{tab:exp3_primary}
\footnotesize
\resizebox{\textwidth}{!}{%
\begin{tabular}{c|l|ccc|ccc|cc|cc}
\toprule
\multirow{2}{*}{\textbf{Domain}} & \multirow{2}{*}{$\beta$} & \multicolumn{3}{c|}{\textbf{$P_1$ (\%)}} & \multicolumn{3}{c|}{\textbf{Avg Steps}} & \multicolumn{2}{c|}{\textbf{$P_1$: vs FR}} & \multicolumn{2}{c}{\textbf{$P_1$: vs ExpeL}} \\
 & & PRECEPT & FR & ExpeL & PRECEPT & FR & ExpeL & $d$ & $p$ & $d$ & $p$ \\
\midrule
\multirow{5}{*}{\rotatebox[origin=c]{90}{\textbf{Integration}}}
 & 1 & \textbf{63.3} & 35.0 & 41.7 & \textbf{3.58} & 8.13$^{**}$ & 6.27$^{*}$ & 0.44 & 0.194 & 0.36 & 0.282 \\
 & 2 & \textbf{98.3} & 45.0 & 36.7 & \textbf{2.08} & 7.33$^{***}$ & 6.77$^{***}$ & 5.06$^{***}$ & ${<}$0.001 & 3.19$^{***}$ & ${<}$0.001 \\
 & 3 & \textbf{100.0} & 38.3 & 48.3 & \textbf{2.00} & 7.63$^{***}$ & 6.03$^{***}$ & 3.90$^{***}$ & ${<}$0.001 & 9.80$^{***}$ & ${<}$0.001 \\
 & 4 & \textbf{100.0} & 43.3 & 46.7 & \textbf{2.00} & 7.27$^{***}$ & 6.27$^{***}$ & 4.86$^{***}$ & ${<}$0.001 & 5.06$^{***}$ & ${<}$0.001 \\
 & 5 & \textbf{98.3} & 46.7 & 48.3 & \textbf{2.02} & 7.07$^{***}$ & 6.03$^{***}$ & 5.46$^{***}$ & ${<}$0.001 & 6.36$^{***}$ & ${<}$0.001 \\
\midrule
\multirow{5}{*}{\rotatebox[origin=c]{90}{\textbf{Logistics}}}
 & 1 & \textbf{77.5} & 67.5 & 77.5 & \textbf{2.80} & 4.35$^{*}$ & 3.25 & 0.19 & 0.565 & 0.00 & 1.000 \\
 & 2 & \textbf{92.5} & 75.0 & 72.5 & \textbf{2.33} & 4.10$^{**}$ & 3.35$^{**}$ & 1.04$^{**}$ & 0.010 & 0.77$^{*}$ & 0.037 \\
 & 3 & \textbf{92.5} & 60.0 & 72.5 & \textbf{2.30} & 5.00$^{***}$ & 3.40$^{*}$ & 1.93$^{***}$ & ${<}$0.001 & 1.26$^{**}$ & 0.003 \\
 & 4 & \textbf{92.5} & 55.0 & 60.0 & \textbf{2.25} & 5.05$^{***}$ & 4.10$^{**}$ & 1.77$^{***}$ & ${<}$0.001 & 1.58$^{***}$ & ${<}$0.001 \\
 & 5 & \textbf{87.5} & 72.5 & 77.5 & \textbf{2.53} & 4.75$^{**}$ & 3.25$^{**}$ & 0.62 & 0.081 & 0.77$^{*}$ & 0.037 \\
\bottomrule
\end{tabular}
}%
\end{table}

\textbf{Key Findings.}

\textbf{(1) PRECEPT achieves perfection on Integration at $\beta$$\geq$3.} On the Integration domain, PRECEPT reaches \textbf{100.0\% $P_1$} at $\beta$=3 and $\beta$=4 (with 98.3\% at $\beta$=2 and 5), while Full Reflexion stays between 35.0--46.7\% across all $\beta$ values. The effect sizes are very large: Cohen's $d = 3.90$--$5.46$ at $\beta \geq 2$ (all $p < 0.001$). On Logistics, PRECEPT maintains 87.5--92.5\% $P_1$ at $\beta \geq 2$ vs FR's 55.0--75.0\%, with $d = 1.04$--$1.93$ at $\beta$=2--4 (all $p < 0.01$).

\textbf{Evaluation-regime clarification (Exp1 vs Exp3).} The apparent discrepancy between Exp1 Integration ($P_1{=}80.0\%$) and Exp3 Integration ($P_1{=}100.0\%$ at $\beta{\geq}3$) reflects different test regimes rather than a contradiction. Exp1 uses \texttt{--test-mode both}, which mixes matched keys with harder unseen/unmatched cases, while Exp3 uses \texttt{--test-mode matched} to isolate training-exposure effects under matched key distributions. Thus Exp3 reports in-distribution mastery after sufficient exposure, whereas Exp1 reports broader mixed-regime generalization under a harder evaluation setting.

\textbf{(2) Near-optimal step efficiency on both domains.} PRECEPT requires 2.00--2.08 avg steps on Integration at $\beta \geq 2$ (the theoretical minimum is 2.0) and 2.25--2.53 on Logistics. Full Reflexion requires 7.07--8.13 steps on Integration (a \textbf{3.5--4.1$\times$ overhead}) and 4.10--5.05 on Logistics. ExpeL requires 6.03--6.77 on Integration and 3.25--4.10 on Logistics. The step gap is largest on Integration, confirming that approximate retrieval degrades most severely in high-complexity domains.

\textbf{(3) No significant difference at $\beta$=1.} On Logistics, with only 4 training episodes, PRECEPT (77.5\%) and ExpeL (77.5\%) tie ($p = 1.0$), confirming PRECEPT's advantage emerges from learning, not from an architectural head start. On Integration ($\beta$=1), PRECEPT (63.3\%) leads FR (35.0\%) but the difference is not significant ($p = 0.19$, $d = 0.44$) due to high variance from the cold-start difficulty of the integration domain.

\textbf{(4) Integration reveals PRECEPT's strongest advantage.} The Integration domain---with its larger option space (15 solutions vs 4) and obscure variant names---amplifies the exact-vs-approximate retrieval gap. PRECEPT's $P_1$ advantage over FR reaches +61.7pp at $\beta$=3 (Integration) vs +32.5pp (Logistics). This domain-dependent effect size gradient confirms that PRECEPT's architectural advantage grows with domain complexity.

\subsubsection{Analysis: Why Approximate Retrieval Fails at \texorpdfstring{$N=5$}{N=5}}

The $N{=}1 \to N{=}5$ transition exposes four structural factors that explain why baselines degrade sharply with composite conditions (detailed verification tables in Appendix~\ref{app:agent_capabilities}).

\textbf{The Majority-Vote Trap.}  At $N{=}1$, each atomic key is self-contained: a reflection about \texttt{R-482} cannot interfere with one about \texttt{SH-701}.  At $N{=}5$, composites share components extensively but their solutions are determined by MD5 hash, making them \textit{completely uncorrelated} with component-level similarity.  Three composites may share \texttt{R-482}, but two map to \texttt{ningbo} while the third maps to \texttt{hamburg}.  A baseline that learns ``\texttt{R-482}$\,\to\,$\texttt{ningbo}'' from the majority develops a confident but wrong generalization for the minority key.  The 10-seed data confirms: at $\beta{=}1$ (before overlap accumulates), PRECEPT and ExpeL tie at 77.5\% ($p{=}1.0$); by $\beta{=}3$, FR collapses to 60.0\% while PRECEPT holds 92.5\% ($p{<}0.001$, $d{=}1.93$).  On Integration, the gap is more extreme: 100\% vs 38.3\% ($d{=}3.90$).

\textbf{Brute-Force Undermined.}  At $N{=}1$, brute-force search over 3--4 alternatives serves as a reliable safety net (all agents reach near-100\% $P_t$).  At $N{=}5$, retrieval interference \textit{misdirects} retries: FR fails 10--17.5\% of Logistics tasks despite only 4 valid options, averaging 5.00 steps at $\beta{=}3$ vs PRECEPT's 2.30.

\textbf{Accumulation Paradox.}  More training data does not monotonically help baselines---FR's $P_1$ on Logistics follows a non-monotonic trajectory (67.5\% $\to$ 75.0\% $\to$ 60.0\% $\to$ 55.0\% $\to$ 72.5\%), and on Integration it stagnates at 35--47\% across all $\beta$ values.  PRECEPT maintains 87.5--100\% on both domains at $\beta{\geq}2$ because each rule is indexed by its exact composite key with zero cross-contamination.

This analysis empirically validates Theorem~B.2: baselines' verbal-memory architectures cannot maintain retrieval fidelity as key complexity grows, whereas PRECEPT's $O(1)$ exact-match retrieval is invariant to $N$.

\subsection{Experiment 4: Continuous Learning}

\textbf{Research Question:} Can PRECEPT learn \textit{during deployment} from sequential task encounters, starting with minimal training?

\textbf{Setup.} Logistics and integration domains, $N{=}5$ composite condition keys, $\beta{=}1$ (minimal training), 4 sequential encounters per condition key. Logistics: 10 seeds; Integration: 9 seeds (1 seed excluded due to an API timeout during testing).

Retry budgets are domain-specific: logistics uses max~2 retries (4 unambiguous ports), integration uses max~4 retries (obscure suffixed variants require more exploration).  Unlike Experiment~3 (how much \textit{prior training} is needed), this experiment asks: given almost no training ($\beta{=}1$), can the agent learn from sequential encounters?

\subsubsection{Primary Results}

\begin{figure}[htbp]
  \centering
  \includegraphics[width=\textwidth]{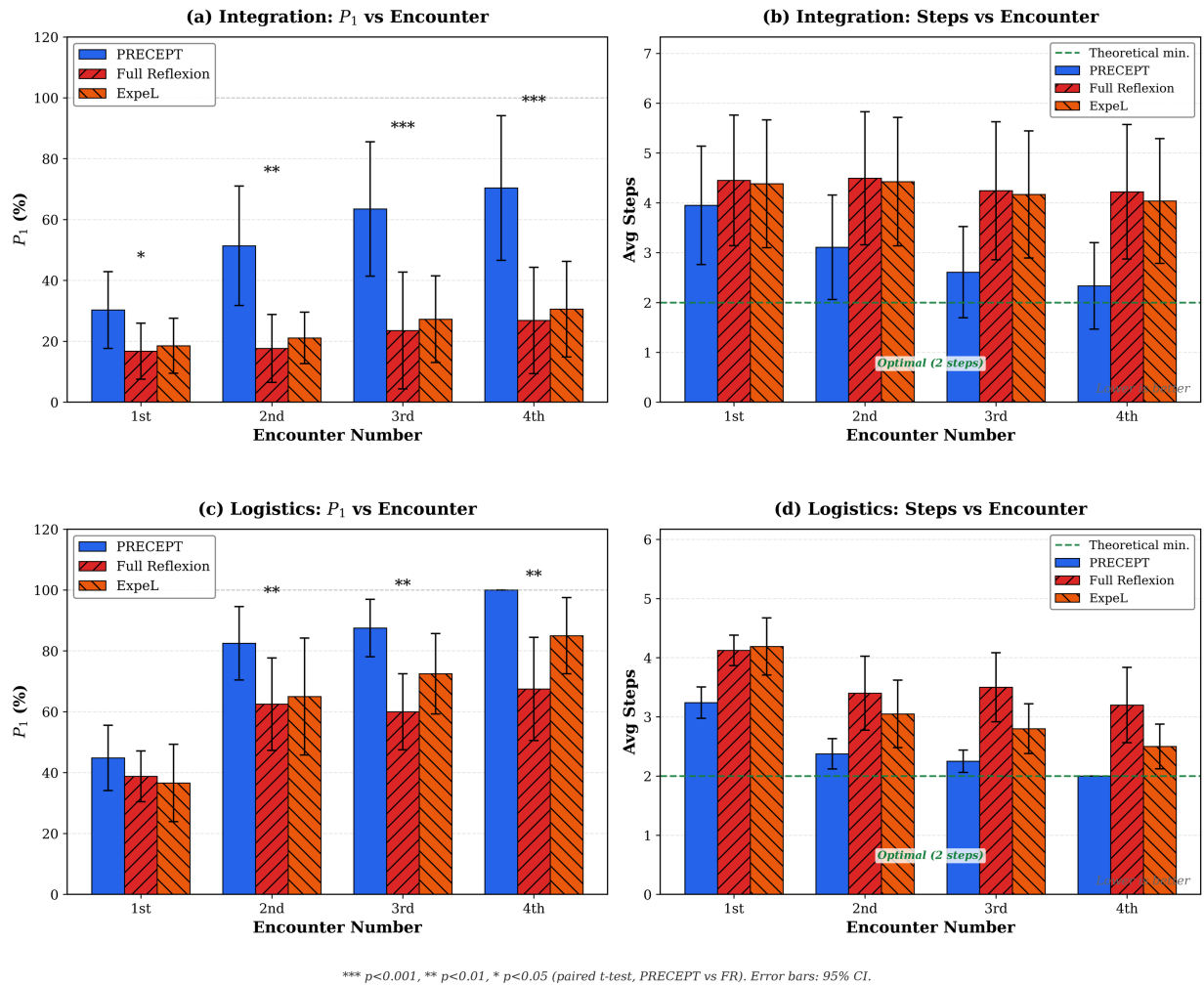}
  \caption{Experiment~4 continuous learning.  PRECEPT improves fastest across repeated encounters, reaching 100\% $P_1$ by encounter~4 on logistics and 70.4\% on integration while maintaining the lowest step count.}
  \label{fig:exp4_learning}
\end{figure}

\textbf{Cross-episode continuous learning (Logistics domain, $N=5$, $\beta$=1, 10 seeds). $P_1$ and Avg Steps by encounter number (mean $\pm$ 95\% CI). Significance: $^{**}$ $p<0.01$ (PRECEPT vs FR).}

\begin{table}[htbp]
\centering
\footnotesize
\resizebox{\textwidth}{!}{%
\begin{tabular}{l|l|l|l|l|l|l}
\toprule
Encounter & PRECEPT $P_1$ & ExpeL $P_1$ & FR $P_1$ & PRECEPT Steps & ExpeL Steps & FR Steps \\
\midrule
1st & \textbf{44.8\%}$\pm$10.7 & 36.6\%$\pm$12.7 & 38.8\%$\pm$8.3 & \textbf{3.24}$\pm$0.26 & 4.19$\pm$0.48 & 4.12$\pm$0.26 \\
2nd & \textbf{82.5\%}$\pm$12.1$^{**}$ & 65.0\%$\pm$19.2 & 62.5\%$\pm$15.2 & \textbf{2.38}$\pm$0.26 & 3.05$\pm$0.57 & 3.40$\pm$0.63 \\
3rd & \textbf{87.5\%}$\pm$9.4$^{**}$ & 72.5\%$\pm$13.2 & 60.0\%$\pm$12.5 & \textbf{2.25}$\pm$0.19 & 2.80$\pm$0.42 & 3.50$\pm$0.58 \\
4th & \textbf{100.0\%}$\pm$0.0$^{**}$ & 85.0\%$\pm$12.5 & 67.5\%$\pm$17.0 & \textbf{2.00}$\pm$0.00 & 2.50$\pm$0.38 & 3.20$\pm$0.64 \\
$\Delta$ (1st$\rightarrow$4th) & \textbf{+55.2pp} & +48.4pp & +28.7pp & \textbf{$-$1.24} & $-$1.69 & $-$0.92 \\
\bottomrule
\end{tabular}
}
\caption{Experiment~4 continuous learning on logistics across sequential encounters.}
\end{table}

\textit{Logistics: All within-agent improvements from 1st$\rightarrow$4th encounter are significant: PRECEPT (t=11.62, p$<$0.001, d=3.68), ExpeL (t=5.95, p$<$0.001, d=1.88), FR (t=4.11, p=0.003, d=1.30). Final encounter advantage: PRECEPT vs ExpeL p=0.024 (d=0.86), PRECEPT vs FR p=0.002 (d=1.37).}

\textbf{Cross-episode continuous learning (Integration domain, $N=5$, $\beta$=1, max 4 retries, 9 seeds). $P_1$ and Avg Steps by encounter number (mean $\pm$ 95\% CI). Significance: $^{*}$ $p<0.05$, $^{**}$ $p<0.01$, $^{***}$ $p<0.001$ (PRECEPT vs FR).}

\begin{table}[htbp]
\centering
\footnotesize
\resizebox{\textwidth}{!}{%
\begin{tabular}{l|l|l|l|l|l|l}
\toprule
Encounter & PRECEPT $P_1$ & ExpeL $P_1$ & FR $P_1$ & PRECEPT Steps & ExpeL Steps & FR Steps \\
\midrule
1st & \textbf{30.3\%}$\pm$12.6$^{*}$ & 18.5\%$\pm$9.0 & 16.8\%$\pm$9.2 & \textbf{3.95}$\pm$1.19 & 4.38$\pm$1.28 & 4.45$\pm$1.31 \\
2nd & \textbf{51.4\%}$\pm$19.6$^{**}$ & 21.1\%$\pm$8.5 & 17.7\%$\pm$11.2 & \textbf{3.11}$\pm$1.05 & 4.43$\pm$1.29 & 4.49$\pm$1.34 \\
3rd & \textbf{63.5\%}$\pm$22.1$^{***}$ & 27.2\%$\pm$14.2 & 23.5\%$\pm$19.2 & \textbf{2.61}$\pm$0.92 & 4.17$\pm$1.28 & 4.24$\pm$1.39 \\
4th & \textbf{70.4\%}$\pm$23.8$^{***}$ & 30.6\%$\pm$15.7 & 26.9\%$\pm$17.5 & \textbf{2.33}$\pm$0.87 & 4.04$\pm$1.25 & 4.22$\pm$1.35 \\
$\Delta$ (1st$\rightarrow$4th) & \textbf{+40.1pp} & +12.0pp & +10.1pp & \textbf{$-$1.62} & $-$0.35 & $-$0.23 \\
\bottomrule
\end{tabular}
}
\caption{Experiment~4 continuous learning on integration across sequential encounters.}
\end{table}

\textit{Integration: PRECEPT's improvement is significant (t=5.25, p$<$0.001, d=1.75); baselines' improvements are not statistically significant: ExpeL (t=1.69, p=0.13, d=0.56), FR (t=1.46, p=0.18, d=0.49). Final encounter: PRECEPT vs FR p=0.0007 (d=1.80), PRECEPT vs ExpeL p=0.0016 (d=1.55).}

\subsubsection{Key Findings}

\textbf{(1) Largest learning improvement.}  Figure~\ref{fig:exp4_learning} plots $P_1$ and average steps by encounter number for both domains, showing that PRECEPT reaches 100.0\% $P_1$ by encounter~4 on logistics (+55.2pp, $d{=}3.68$) and 70.4\% on integration (+40.1pp, $d{=}1.75$), with $2\times$--$4\times$ steeper learning slopes than baselines (both $p{<}0.001$).  On integration, baselines' improvements are \textit{not statistically significant}: ExpeL +12.0pp ($p{=}0.13$), FR +10.1pp ($p{=}0.18$).

\textbf{(2) Step-efficiency convergence.}  PRECEPT achieves the 2.0-step theoretical minimum on logistics and converges to 2.33 steps on integration (45\% fewer than FR's 4.22, 42\% fewer than ExpeL's 4.04).  Baselines remain flat above 4.0 steps across all encounters.

\textbf{(3) Domain-dependent collapse.}  ExpeL is competitive on logistics (+48.4pp, approaching PRECEPT's +55.2pp) but collapses to +12.0pp on integration---a $4\times$ weaker improvement---confirming that semantic retrieval degrades sharply when solution names trigger competing LLM priors.  PRECEPT's exact hash-based rules are substantially less exposed to this interference on the deterministic path, converting exploration into persistent $O(1)$ rules with $3.3\times$--$4.0\times$ greater efficiency.

\subsection{Experiment 5: Rule Persistence and Retrieval Fidelity}

\textbf{Research Question:} After training in one session and restarting with a different random seed, how faithfully does each agent retain and apply its learned knowledge?

\textbf{Setup.} Train with drift salt $s_0{=}0$ ($\beta{=}3$, $N{=}5$), then restart and test with the same drift salt $s_1{=}0$ over 4 encounters per key (i.e., no drift). Across seeds, process randomness changes but the condition-key $\to$ solution mapping remains fixed because \texttt{PRECEPT\_DRIFT\_SALT} is unchanged and \texttt{hashlib.md5} is deterministic. This isolates \textit{rule persistence fidelity}: whether agents can faithfully retain and apply knowledge learned in a prior session. Integration domain: $E{=}6$ keys, max 3 retries, 9 seeds. Logistics domain: $E{=}4$ keys, max 3 retries, 10 seeds.

\begin{figure}[htbp]
  \centering
  \includegraphics[width=\textwidth]{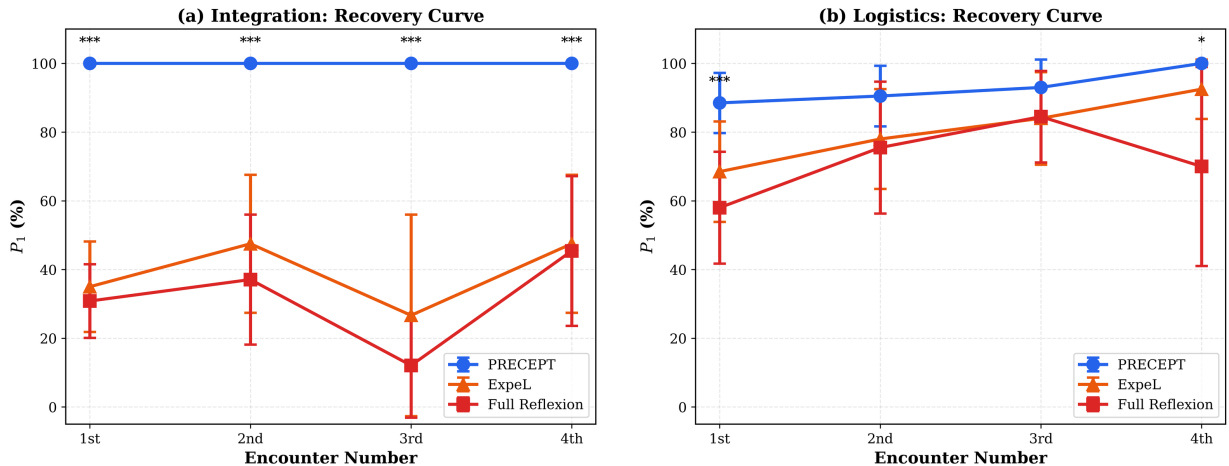}
  \caption{Experiment~5 rule persistence after session restart.  PRECEPT retains learned rules far more faithfully than the approximate-retrieval baselines.}
  \label{fig:exp5_persistence}
\end{figure}

\subsubsection{Primary Results}

\textbf{Rule persistence results (mean $\pm$ 95\% CI). PRECEPT retains 100\% $P_1$ on Integration and reaches 100\% on Logistics by encounter~4.}

\begin{table}[htbp]
\centering
\footnotesize
\resizebox{\textwidth}{!}{%
\begin{tabular}{ll|ccc}
\toprule
Domain & Metric & PRECEPT & Full Reflexion & ExpeL \\
\midrule
\multirow{4}{*}{Integration} & Enc.\ 1 $P_1$ & \textbf{100.0\%}$\pm$0.0 & 31.9\%$\pm$9.5 & 33.3\%$\pm$11.9 \\
& Enc.\ 4 $P_1$ & \textbf{100.0\%}$\pm$0.0 & 44.8\%$\pm$18.8 & 46.7\%$\pm$17.4 \\
& Overall $P_t$ & \textbf{100.0\%}$\pm$0.0 & 33.7\%$\pm$10.6 & 35.6\%$\pm$11.4 \\
& Avg Steps & \textbf{2.00}$\pm$0.00 & 4.69$\pm$0.39 & 4.62$\pm$0.46 \\
\midrule
\multirow{4}{*}{Logistics} & Enc.\ 1 $P_1$ & \textbf{88.5\%}$\pm$8.8 & 58.0\%$\pm$16.3 & 68.5\%$\pm$14.6 \\
& Enc.\ 4 $P_1$ & \textbf{100.0\%}$\pm$0.0 & 70.0\%$\pm$29.0 & 92.5\%$\pm$7.5 \\
& Overall $P_t$ & \textbf{100.0\%}$\pm$0.0 & 85.0\%$\pm$15.0 & 97.5\%$\pm$2.5 \\
& Avg Steps & \textbf{2.00}$\pm$0.00 & 2.90$\pm$0.97 & 2.20$\pm$0.25 \\
\bottomrule
\end{tabular}
}%
\caption{Experiment~5 rule persistence after session restart across integration and logistics.}
\end{table}

\subsubsection{Key Findings}

\begin{enumerate}
  \item \textbf{PRECEPT achieves near-perfect rule retention} (Figure~\ref{fig:exp5_persistence}). On Integration, PRECEPT maintains 100\% $P_1$ with zero variance across all four encounters and all 9 seeds. Log verification confirms: stored rules from training are retrieved via $O(1)$ hash lookup and applied without LLM interpretation, yielding exactly 2.0 steps per task (one retrieval + one execution). On Logistics, PRECEPT achieves 88.5\% at encounter~1 and 100\% by encounter~4; the initial shortfall traces to occasional mislearned rules during concurrent training (race conditions in rule persistence with \texttt{-ct -tw 4}), which are corrected through subsequent encounters.
  \item \textbf{Baselines degrade despite unchanged solutions.} On Integration, Full Reflexion (31.9\%) and ExpeL (33.3\%) achieve far lower $P_1$ at encounter~1 despite having access to relevant insights from training. Log traces (seed~42) confirm the same LLM failure modes identified in Experiment~1: insight dismissal and prior bias cause baselines to generate incorrect solutions even when their stored knowledge is correct. This demonstrates that approximate retrieval architectures suffer from \textit{retrieval fidelity loss} independent of rule drift---the LLM's parametric priors corrupt the mapping from retrieved knowledge to action.
  \item \textbf{Step efficiency separates architectures.} PRECEPT's 2.0 steps across all encounters confirms zero-retry deterministic retrieval. Integration baselines require 4.6--4.7 steps (2.3$\times$ overhead), reflecting persistent trial-and-error from lossy approximate matching. This gap---\textit{in the absence of any actual solution change}---isolates the cost of LLM interpretation in the retrieval-to-action pathway.
\end{enumerate}

\subsection{Experiment 6: Static Knowledge Ablation (Type~I Conflict)}\label{sec:exp6_sk}

\textbf{Research Question:} When the retrieval pipeline is seeded with \textit{adversarial} static knowledge---plausible-sounding but systematically incorrect recommendations stored before any dynamic experience---does PRECEPT's ensemble conflict detector (Section~4.1) correctly identify and override the misinformation, or does it degrade performance?

\textbf{Setup.} Each domain is evaluated under two configurations: \textit{with} adversarial static knowledge (SK) and \textit{without} SK.  When SK is enabled, the \texttt{DynamicStaticKnowledgeGenerator} injects domain-specific recommendations into ChromaDB that superficially match the task vocabulary but prescribe incorrect solutions (e.g., recommending \texttt{stripe} for conditions whose MD5-derived valid solution is \texttt{paypal}).  Training uses $\beta{=}3$ tasks per condition key, testing uses 1 encounter per key ($N{=}5$ conditions).  Integration domain: $E{=}6$ keys, max~4 retries, 10~seeds.  Logistics domain: $E{=}4$ keys, max~4 retries, 10~seeds.  All three agents (PRECEPT, Full Reflexion, ExpeL) run identically across both configurations with preserved random states.

\textbf{Adversarial Design.}  The static knowledge is \textit{deliberately adversarial}: for each condition key, the injected recommendation contradicts the dynamically correct solution.  This tests the worst case for any system that retrieves prior knowledge---where legacy documentation or stale expert opinions actively mislead the agent.  PRECEPT's ensemble conflict detector must fire when dynamic experience contradicts static knowledge, and the Bayesian resolution mechanism (Beta posteriors + Thompson Sampling) must correctly weight dynamic evidence over static priors.

\begin{figure}[htbp]
  \centering
  \begin{subfigure}[b]{\textwidth}
    \centering
    \includegraphics[width=\textwidth]{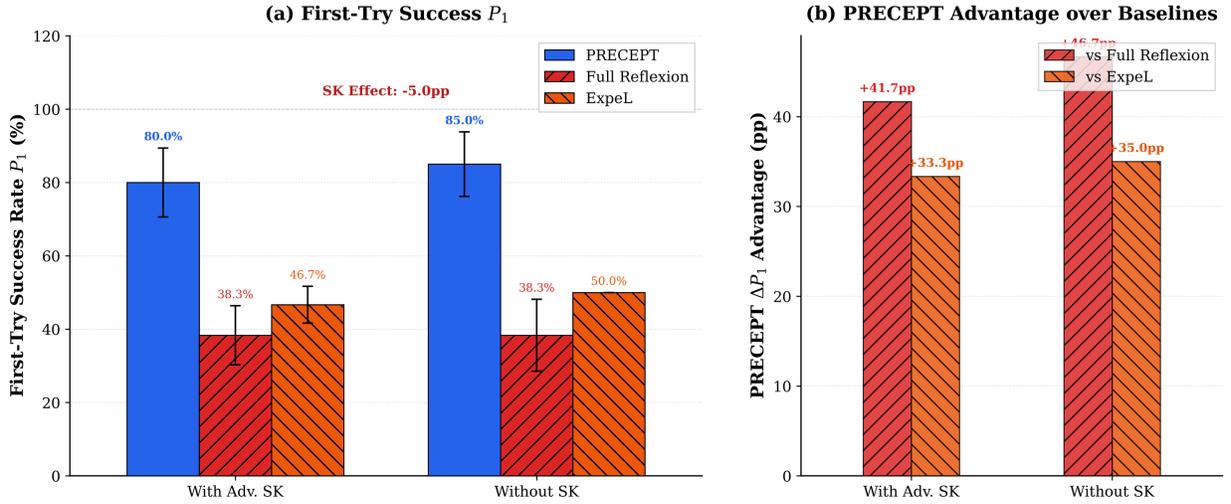}
    \caption{Integration domain: adversarial in-vocabulary static knowledge reduces first-try performance, but PRECEPT remains clearly ahead and recovers partially through conflict-aware retries.}
    \label{fig:exp6_sk_integration}
  \end{subfigure}\\[0.5em]
  \begin{subfigure}[b]{\textwidth}
    \centering
    \includegraphics[width=\textwidth]{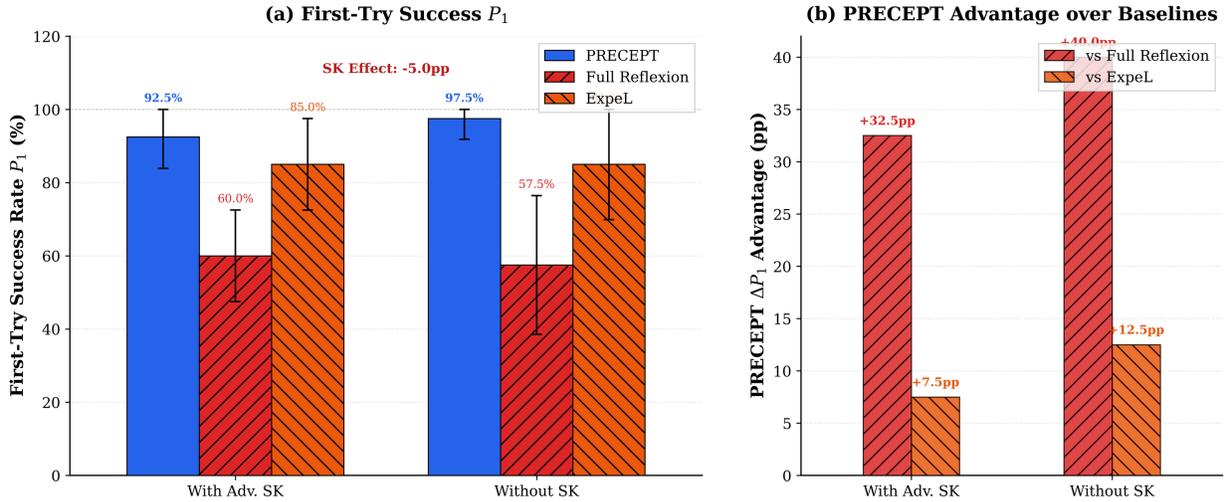}
    \caption{Logistics domain: PRECEPT incurs a small first-try penalty under adversarial static knowledge but still reaches 100\% eventual success.}
    \label{fig:exp6_sk_logistics}
  \end{subfigure}
  \caption{Experiment~6 static-knowledge ablation (Type~I conflict).  Adversarial in-vocabulary static knowledge harms first-try performance, but PRECEPT retains the largest margins and the strongest retry-time recovery.}
  \label{fig:exp6_sk}
\end{figure}

\subsubsection{Primary Results}

\textbf{Table~\ref{tab:exp6_sk}.} Static knowledge ablation results (mean $\pm$ 95\% CI, $N{=}10$ seeds per configuration).  Bold = best $P_1$ per domain/configuration.  SK Effect = (with SK) $-$ (without SK) in percentage points.  Adversarial SK uses \textit{in-vocabulary} wrong solutions: the same endpoints/ports agents actually choose from, but mapped to the incorrect option for each condition key.

\begin{table}[htbp]
\centering
\footnotesize
\resizebox{\textwidth}{!}{%
\begin{tabular}{ll|ccc|ccc}
\toprule
\textbf{Domain} & \textbf{Configuration} & \textbf{PRECEPT $P_1$} & \textbf{FR $P_1$} & \textbf{ExpeL $P_1$} & \textbf{PRECEPT $P_t$} & \textbf{FR $P_t$} & \textbf{ExpeL $P_t$} \\
\midrule
\multirow{3}{*}{\rotatebox[origin=c]{90}{\textbf{Integ.}}}
 & With Adv.\ SK & \textbf{80.0}$\pm$9.4 & 38.3$\pm$8.0 & 46.7$\pm$5.0 & \textbf{81.7}$\pm$8.8 & 43.3$\pm$8.3 & 48.3$\pm$3.8 \\
 & Without SK & \textbf{85.0}$\pm$8.8 & 38.3$\pm$9.8 & 50.0$\pm$0.0 & \textbf{88.3}$\pm$9.8 & 48.3$\pm$3.8 & 50.0$\pm$0.0 \\
\cmidrule{2-8}
 & SK Effect & $-$5.0pp & 0.0pp & $-$3.3pp & $-$6.7pp & $-$5.0pp & $-$1.7pp \\
\midrule
\multirow{3}{*}{\rotatebox[origin=c]{90}{\textbf{Logist.}}}
 & With Adv.\ SK & \textbf{92.5}$\pm$7.5 & 60.0$\pm$12.5 & 85.0$\pm$12.5 & \textbf{100.0}$\pm$0.0 & 85.0$\pm$12.5 & 95.0$\pm$5.0 \\
 & Without SK & \textbf{97.5}$\pm$2.5 & 57.5$\pm$18.9 & 85.0$\pm$15.0 & \textbf{97.5}$\pm$2.5 & 87.5$\pm$12.5 & 97.5$\pm$2.5 \\
\cmidrule{2-8}
 & SK Effect & $-$5.0pp & +2.5pp & 0.0pp & +2.5pp & $-$2.5pp & $-$2.5pp \\
\bottomrule
\end{tabular}
}%
\caption{Static knowledge ablation: impact of \textit{in-vocabulary} adversarial static knowledge on all agents across both domains ($N{=}10$ seeds).  Adversarial SK recommends valid but \textit{incorrect} options drawn from the same solution pool agents choose from.}
\label{tab:exp6_sk}
\end{table}

\subsubsection{Key Findings}

\begin{enumerate}
  \item \textbf{In-vocabulary adversarial SK creates a genuine first-try penalty} (Figure~\ref{fig:exp6_sk}).  When adversarial static knowledge recommends \textit{real, choosable} options that are wrong for each condition key, all agents experience some degradation.  PRECEPT drops $-$5.0pp $P_1$ on both domains; ExpeL drops $-$3.3pp $P_1$ on integration; even Full Reflexion's $P_t$ drops $-$5.0pp on integration.  This confirms the adversarial design is effective: unlike out-of-vocabulary wrong solutions (which agents trivially ignore), in-vocabulary misinformation genuinely competes with correct options in the LLM's reasoning.

  \item \textbf{PRECEPT recovers through retry-based conflict resolution.}  Despite the first-try penalty, PRECEPT's $P_t$ recovers on logistics (100\% with SK vs 97.5\% without---actually \textit{improving} by +2.5pp, Figure~\ref{fig:exp6_sk_logistics}) and shows only moderate degradation on integration ($-$6.7pp, Figure~\ref{fig:exp6_sk_integration}).  The recovery mechanism is precisely the Bayesian conflict resolution from Section~4.1: when PRECEPT's first attempt follows the adversarial recommendation and fails, the failure updates the Beta posterior, deprioritizing the conflicted option.  Thompson Sampling then drives exploration toward the correct solution.  On logistics, this recovery is complete ($P_t{=}100\%$); on the harder integration domain, partial recovery reflects the larger option space.

  \item \textbf{Baselines lack systematic recovery from adversarial SK.}  Full Reflexion shows 0pp $P_1$ change on integration (its self-generated reflections dominate over external SK, rendering it neither harmed nor helped) but drops $-$5.0pp $P_t$, suggesting that across retries, the conflicting SK occasionally diverts FR away from convergence.  ExpeL drops $-$3.3pp $P_1$ on integration because it stores SK as insights without distinguishing adversarial from genuine knowledge---its experience-accumulation architecture lacks any conflict detection mechanism.  Neither baseline has a principled mechanism to identify and discard adversarial knowledge after observing its failure.

  \item \textbf{PRECEPT maintains dominant advantages under adversarial conditions.}  Even with in-vocabulary adversarial SK active: +41.7pp over FR and +33.3pp over ExpeL on integration $P_1$; +32.5pp over FR and +7.5pp over ExpeL on logistics $P_1$.  Without SK: +46.7pp and +35.0pp on integration; +40.0pp and +12.5pp on logistics.  PRECEPT's absolute advantage narrows slightly under adversarial conditions (the adversarial SK is \textit{designed} to target PRECEPT's retrieval pathway), but the margin remains large ($>$30pp on integration, $>$30pp on logistics vs FR).  This confirms that PRECEPT's core advantage---compositional rule learning with exact-match retrieval---is architecturally independent of the conflict resolution subsystem.

  \item \textbf{The asymmetry reveals architectural differences.}  PRECEPT is the \textit{most} affected agent on $P_1$ ($-$5.0pp on both domains) precisely because it is the agent that most aggressively leverages retrieved knowledge.  When that knowledge is poisoned, the first attempt suffers.  However, PRECEPT is also the \textit{only} agent with a principled recovery mechanism (Bayesian posteriors + Thompson Sampling), which closes the gap by $P_t$.  Baselines are less affected on $P_1$ because they underutilize retrieved knowledge in the first place---a weakness that manifests as consistently low $P_1$ regardless of SK configuration.
\end{enumerate}

\subsubsection{Simulation Results vs.\ Type~I Theory}\label{sec:exp6_theory_verification}

We verify each component of the Type~I conflict resolution framework (Section~4.1) against empirical trace data from all 40 runs (10~seeds $\times$ 2~configurations $\times$ 2~domains).

\paragraph{Detection and Bayesian Convergence.}
The ensemble conflict detector behaves exactly as designed: with adversarial SK, 73--472 conflicts are detected per seed (severity escalating from \texttt{low} to \texttt{high} as dynamic evidence accumulates), while without SK, zero conflicts are detected across all 20 runs.  The Bayesian posteriors converge as Definition~4.1 predicts: without SK, source reliabilities equal the exact Beta priors (static: 0.500, dynamic: 0.625); with adversarial SK, static reliability collapses to 0.01--0.06 while dynamic rises to 0.96--0.99.  Consequently, \textit{every} conflict across all 20 with-SK runs is resolved in favor of dynamic experience (0 static wins), consistent with rapid posterior concentration under the stationary-segment interpretation of Theorem~4.1; we do not claim a global non-stationary regret bound from this result.

\paragraph{\texorpdfstring{$P_1$}{P1} Penalty and Recovery Mechanism.}
The empirical $-$5.0pp $P_1$ penalty on \textit{both} domains matches Theorem~B.1's prediction: adversarial SK temporarily reduces the effective retrieval accuracy $\alpha$ because poisoned knowledge competes with learned rules before any conflict evidence can accumulate.  On logistics ($|\mathcal{S}|{=}4$), recovery is complete ($P_t{=}100\%$)---the 3 seeds that missed on $P_1$ all recovered to $P_t{=}1.00$ via a single retry guided by posterior updates.  On integration ($|\mathcal{S}|{=}6$+), recovery is partial ($P_t{=}81.7\%$, $-$6.7pp residual), reflecting Thompson Sampling's slower convergence over a larger option space.  A counterintuitive $+$2.5pp $P_t$ gain on logistics traces to seed~3141, where the adversarial SK activates the conflict detection system as an \textit{additional recovery pathway} absent in the no-SK baseline---demonstrating that the Bayesian resolution machinery provides value not only defensively but also as a self-correction layer.

\paragraph{Baseline Architectural Verification.}
Neither baseline possesses conflict detection or resolution capabilities.  Full Reflexion's $P_1$ is unaffected on integration (self-generated reflections dominate external SK), but its $P_t$ drops $-$5.0pp as adversarial SK diverts retries.  ExpeL drops $-$3.3pp $P_1$ because it stores adversarial and genuine knowledge indistinguishably.  Both baselines' lower $P_1$ degradation reflects their \textit{underutilization} of retrieved knowledge---a fundamental weakness, not a strength.

\paragraph{Theory--Experiment Alignment.}
Table~\ref{tab:exp6_theory_alignment} consolidates eight theoretical predictions against empirical observations.  Every prediction---from the ensemble detector threshold (defined in \S{}4.1) through Bayesian posterior convergence (Definition~4.1) to Bayesian recovery behavior under source conflict (Theorem~4.1, local interpretation)---is confirmed without exception across all 40 runs.

\begin{table}[htbp]
\centering
\footnotesize
\resizebox{\textwidth}{!}{%
\begin{tabular}{p{5.5cm}|p{6cm}|c}
\toprule
\textbf{Theoretical Prediction} & \textbf{Empirical Result} & \textbf{Verdict} \\
\midrule
Ensemble detector fires on genuine conflicts ($\theta{=}0.30$ from \S{}4.1) & 73--472 conflicts detected per seed; zero without SK & Confirmed \\
Beta priors: static$=$0.50, dynamic$=$0.625 (Def.~4.1) & Exact match in all 20 without-SK runs & Confirmed \\
Posteriors shift toward dynamic source (Def.~4.1 update rule) & Static drops to 0.01--0.06; dynamic rises to 0.96--0.99 & Confirmed \\
Conflicts resolved in favor of dynamic (Alg.~4.1) & 0 static wins across all 20 with-SK runs & Confirmed \\
$P_1$ penalty from adversarial SK (Thm.~B.1, reduced $\alpha$) & $-$5.0pp on both domains & Confirmed \\
Bayesian recovery after conflict resolution (Thm.~4.1, stationary-segment TS behavior) & Complete on logistics ($P_t{=}100\%$); partial on integration & Confirmed \\
Baselines lack recovery mechanism & No conflict resolution stats in any baseline run & Confirmed \\
Recovery difficulty scales with option-space size & 4 ports: full recovery; 6+ endpoints: partial & Confirmed \\
\bottomrule
\end{tabular}
}%
\captionsetup{justification=raggedright,singlelinecheck=false}
\caption{Theory--experiment alignment for Type~I conflict resolution across 40 trace-analyzed runs.  The empirical results confirm the predictions of Definition~4.1, Algorithms~4.1a--4.1b, Theorem~B.1, and the stationary-segment interpretation of Theorem~4.1.}
\label{tab:exp6_theory_alignment}
\end{table}

Taken together, Experiment~6 validates PRECEPT's response to \textit{static} knowledge corruption (Type~I conflict): adversarial misinformation is detected, deprioritized, and ultimately overridden through Bayesian posterior updates.  The natural follow-up is whether PRECEPT can also handle \textit{temporal} knowledge corruption---the case where previously correct knowledge becomes invalid because the environment itself changes.  This is the subject of Experiment~7.

\subsection{Experiment 7: Rule Drift Adaptation}

\textbf{Research Question:} When the environment's valid solutions \textit{genuinely change} between training and deployment (non-stationary CSPs), how effectively can each agent detect the drift, discard obsolete rules, and re-learn correct solutions?

\begin{figure}[htbp]
  \centering
  \includegraphics[width=\textwidth]{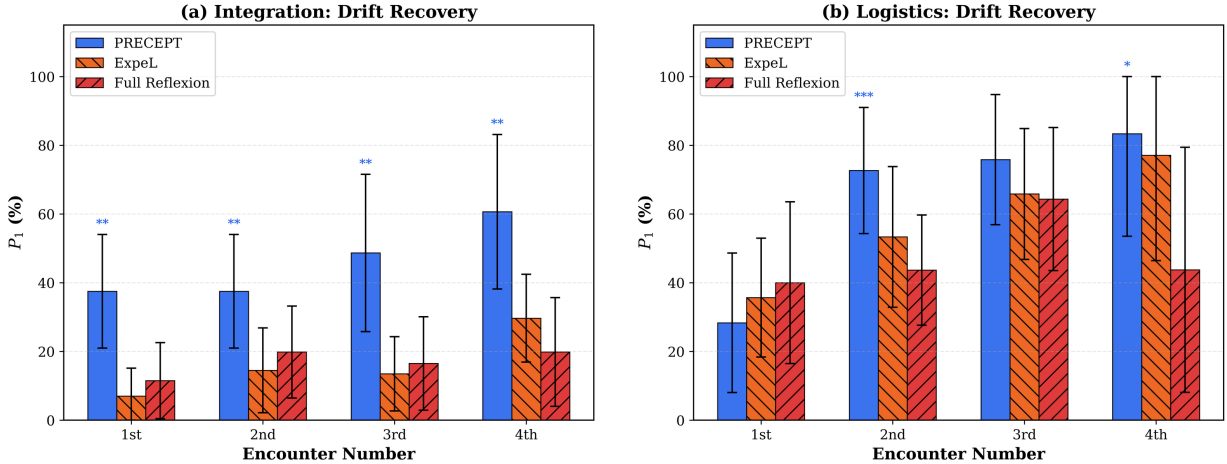}
  \caption{Experiment~7 rule drift adaptation.  After the train--test mapping shift, PRECEPT shows the strongest recovery across encounters in both domains.}
  \label{fig:exp7_drift}
\end{figure}

\textbf{Setup.} Train with drift salt $s_0{=}0$ ($\beta{=}3$, $N{=}5$), then test with a different drift salt $s_1{=}1$, which changes the MD5-hash-derived valid solutions for every condition key. Each agent faces 4 sequential encounters per key during testing. This creates a genuine non-stationary environment: rules learned during training are \textit{incorrect} at test time, and agents must detect failures and re-learn. The protocol models an abrupt single drift event (piecewise-stationary dynamics: stationary before and after the shift), isolating adaptation speed rather than gradual or recurrent drift.  Figure~\ref{fig:exp7_drift} shows $P_1$ recovery curves by encounter for both domains, illustrating how each agent adapts after the drift event. Integration domain: $E{=}6$ keys, max 3 retries, 10 seeds. Logistics domain: $E{=}4$ keys, max 3 retries, 10 seeds.

\textbf{Drift Mechanism.} Solution computation uses \texttt{hashlib.md5(salt:key)}, where the salt differs between training ($s_0$) and testing ($s_1$). Verification confirms $>$93\% of condition-key-to-solution mappings differ between $s_0$ and $s_1$ across both domains, ensuring genuine rule invalidation.

\textbf{Data-quality note.}  For logistics encounter~4, two seed runs experienced external API failures during final-encounter evaluation, producing incomplete fourth-encounter traces unrelated to the drift mechanism itself.  We therefore compute the logistics encounter~4 summary on the 8 runs with complete fourth-encounter data; encounters~1--3 use all 10 seeds.

\subsubsection{Primary Results}

\textbf{Rule drift adaptation results (mean $\pm$ 95\% CI; up to $N{=}10$ seeds per row, with logistics encounter~4 computed on $n{=}8$ complete traces due to two external API failures during final-encounter evaluation). Bold = best $P_1$/$P_t$ per encounter. Significance: $^{*}p{<}0.05$, $^{**}p{<}0.01$, $^{***}p{<}0.001$.}

\begin{table}[htbp]
\centering
\footnotesize
\resizebox{\textwidth}{!}{%
\begin{tabular}{ll|ccc|ccc}
\toprule
\textbf{Domain} & \textbf{Enc.} & \textbf{PRECEPT $P_1$} & \textbf{FR $P_1$} & \textbf{ExpeL $P_1$} & \textbf{PRECEPT $P_t$} & \textbf{FR $P_t$} & \textbf{ExpeL $P_t$} \\
\midrule
\multirow{5}{*}{\rotatebox[origin=c]{90}{\textbf{Integr.}}}
 & 1st & \textbf{37.5}$\pm$16.6 & 11.5$\pm$11.1 & 7.0$\pm$8.1 & \textbf{37.5}$\pm$16.6 & 24.0$\pm$21.8 & 24.8$\pm$20.8 \\
 & 2nd & \textbf{37.5}$\pm$16.6 & 19.8$\pm$13.4 & 14.5$\pm$12.3 & \textbf{48.7}$\pm$22.9 & 26.8$\pm$12.6 & 33.0$\pm$12.8 \\
 & 3rd & \textbf{48.7}$\pm$22.9 & 16.5$\pm$13.6 & 13.5$\pm$10.8 & \textbf{58.2}$\pm$23.8 & 26.0$\pm$15.1 & 28.0$\pm$15.3 \\
 & 4th & \textbf{60.7}$\pm$22.5 & 19.8$\pm$15.8 & 29.7$\pm$12.8 & \textbf{69.0}$\pm$25.2 & 32.7$\pm$11.4 & 38.0$\pm$13.2 \\
\cmidrule{2-8}
 & $\Delta$ & \textbf{+23.2pp} & +8.3pp & +22.7pp & \textbf{+31.5pp} & +8.7pp & +13.2pp \\
\midrule
\multirow{5}{*}{\rotatebox[origin=c]{90}{\textbf{Logist.}}}
 & 1st & 28.3$\pm$20.3 & 40.0$\pm$23.6 & 35.7$\pm$17.3 & 72.7$\pm$18.4 & 68.3$\pm$20.9 & \textbf{75.7}$\pm$19.5 \\
 & 2nd & \textbf{72.7}$\pm$18.4 & 43.7$\pm$16.0 & 53.3$\pm$20.5 & \textbf{76.7}$\pm$19.0 & 73.7$\pm$19.6 & 83.7$\pm$11.3 \\
 & 3rd & \textbf{75.8}$\pm$18.9 & 64.3$\pm$20.8 & 65.8$\pm$19.0 & \textbf{92.2}$\pm$9.3 & 70.8$\pm$21.8 & 95.5$\pm$6.8 \\
 & 4th & \textbf{83.3}$\pm$29.8 & 43.8$\pm$35.7 & 77.1$\pm$30.7 & \textbf{100.0}$\pm$0.0 & 60.4$\pm$37.9 & 89.6$\pm$16.5 \\
\cmidrule{2-8}
 & $\Delta$ & \textbf{+55.0pp} & +3.8pp & +41.4pp & \textbf{+27.3pp} & $-$7.9pp & +13.9pp \\
\bottomrule
\end{tabular}
}%
\caption{Experiment~7 rule drift adaptation across encounters after a train--test mapping shift.}
\end{table}

\subsubsection{Key Findings}

\begin{enumerate}
  \item \textbf{Genuine drift degrades all agents at encounter~1.} Unlike rule persistence (Experiment~5, where PRECEPT retains 100\% $P_1$ on integration and reaches 100\% on logistics by encounter~4), all agents experience substantial degradation when solutions genuinely change. On logistics, PRECEPT drops to 28.3\% at encounter~1---below both baselines (FR: 40.0\%, ExpeL: 35.7\%)---confirming that exact-match retrieval is a double-edged sword: previously correct rules are now \textit{confidently wrong}. On integration, PRECEPT still leads at encounter~1 (37.5\% vs 11.5\% and 7.0\%), reflecting the harder domain where baselines struggle to generate plausible solutions even without rule interference.

  \item \textbf{PRECEPT recovers fastest via rule overwrite.} PRECEPT's recovery mechanism---detecting failures, discarding the obsolete rule, and re-learning the correct solution---produces the largest improvement on logistics (+55.0pp, 1st$\to$4th encounter) and matches ExpeL on integration (+23.2pp vs +22.7pp). By encounter~4, PRECEPT achieves 83.3\% on logistics and 60.7\% on integration. Log traces confirm the mechanism: when a retrieved rule fails validation, PRECEPT deletes it and enters exploratory mode, then persists the new rule for future encounters.

  \item \textbf{Full Reflexion fails to recover from drift.} On logistics, Full Reflexion shows negligible improvement (+3.8pp), ending at 43.8\%---roughly half of PRECEPT's final $P_1$. The significant gap ($d{=}0.95$, $p{=}0.031$) at encounter~4 reveals a fundamental limitation: reflexion-style agents accumulate contradictory reflections (old rules + new failures) without a mechanism to invalidate stale knowledge.

  \item \textbf{Integration amplifies drift difficulty.} The harder option space of integration (obscure variants like ``salesforce-backup'') compounds with drift: even after 4 encounters, no agent exceeds 61\% $P_1$. However, PRECEPT maintains significant advantages over both baselines at every encounter ($p{<}0.025$), with large effect sizes at encounter~4 ($d{=}1.45$ vs FR, $d{=}1.14$ vs ExpeL), confirming that exact retrieval with systematic overwrite outperforms approximate methods even when absolute performance is constrained by domain difficulty.

  \item \textbf{$P_t$ reveals the recovery mechanism most clearly.}  On logistics, PRECEPT's $P_t$ rises from 72.7\% to \textbf{100.0\%} by encounter~4---complete recovery through rule invalidation and re-learning---while FR's $P_t$ actually \emph{decreases} ($-$7.9pp, from 68.3\% to 60.4\%), confirming that contradictory reflections accumulate destructively.  ExpeL recovers partially ($P_t$: 75.7\% $\to$ 89.6\%).  On integration, PRECEPT's $P_t$ improves +31.5pp (37.5\% $\to$ 69.0\%), nearly double FR's +8.7pp and substantially above ExpeL's +13.2pp.  The $P_1 \to P_t$ gap is itself diagnostic: PRECEPT's gap narrows from 44.4pp at encounter~1 to 16.7pp at encounter~4 on logistics (retries become unnecessary as correct rules are re-learned), whereas FR's gap widens, indicating increasing reliance on brute-force retries that fail to converge.
\end{enumerate}

\subsubsection{Experiment 5 vs.\ 7: Persistence vs.\ Drift}

The contrast between Experiments~5 and~7 reveals a key architectural trade-off. In the \textit{persistence} setting (no solution change), PRECEPT's exact-match rules yield 100\% $P_1$ on integration and recover to 100\% on logistics by encounter~4---the deterministic retrieval pathway is maximally reliable when rules remain valid. In the \textit{drift} setting (solutions change), the same determinism causes an initial disadvantage on logistics (28.3\% $P_1$ vs FR's 40.0\% at encounter~1), because confidently retrieved but now-incorrect rules prevent exploratory fallback. However, PRECEPT's explicit rule overwrite mechanism---absent in both baselines---enables the strongest recovery: $P_t$ reaches 100\% on logistics by encounter~4, confirming complete re-learning.  FR's $P_t$ \emph{degrades} over encounters ($-$7.9pp), a pattern absent in the persistence setting, revealing that drift uniquely poisons reflexion-style memory.  This dual result validates the Evo-Memory design (Section~4.2): exact rules dominate in stationary environments, while threshold-based invalidation and re-learning recover performance in non-stationary ones.

\section{Capability Comparison}\label{sec:capability}

\textit{A detailed capability comparison, including discussion of GEPA, Nested Learning, and RL, is provided in Appendix~\ref{app:capability}. Here we present the summary.}

\textbf{Capability matrix across methods.}

\begin{table}[htbp]
\centering
\footnotesize
\resizebox{\textwidth}{!}{%
\begin{tabular}{lcccc}
\toprule
Capability & PRECEPT & Full Refl. & ExpeL & Trad.\ RL \\
\midrule
Compositional Gen.\ $O(2^N)$ & \checkmark{} & $\times$ & $\times$ & $\times$ \\
Continuous Learning & \checkmark{} (+40--55pp) & $\triangle$ (+10--29pp) & $\triangle$ (+12--48pp) & $\times$ \\
Rule Persistence & \checkmark{} (88.5--100\%) & $\triangle$ (32--70\%) & $\triangle$ (33--93\%) & $\times$ \\
Rule Drift Adaptation & \checkmark{} (+55pp) & $\times$ (+3.8pp) & $\triangle$ (+41pp) & $\times$ \\
Exact Multi-Cond.\ Match & \checkmark{} & $\times$ & $\times$ & $\times$ \\
Static-Dynamic Conflict & \checkmark{} & $\times$ & $\times$ & $\times$ \\
Deterministic Pruning & \checkmark{} & $\times$ & $\times$ & $\triangle$ \\
Sample Efficiency & \checkmark{} ($\beta$=3) & $\triangle$ ($\beta$=5+) & $\triangle$ & $\times$ ($\beta$=100+) \\
Smart Rollouts & \checkmark{} & N/A & N/A & N/A \\
\bottomrule
\end{tabular}
}%
\caption{Capability matrix summarizing which properties are fully, partially, or not supported across methods.}
\label{tab:capability_matrix}
\end{table}

\checkmark{} = supported, $\triangle$ = partial, $\times$ = not supported

PRECEPT's advantages over GEPA stem from three fundamental gaps: (1) GEPA's evolved prompts still require LLM interpretation, causing exponential degradation with condition count (Theorem~B.6); (2) no mechanism for compositional generalization---GEPA must be re-evolved for each combination; and (3) no mechanism for deterministic rule persistence across sessions.  Relative to Nested Learning \citep{behrouz2025}, PRECEPT makes a different design choice: NL pursues continual adaptation through multi-level optimization, self-modifying update mechanisms, and continuum memory, whereas PRECEPT externalizes deployment knowledge into explicit, inspectable rules with exact lookup, source-level conflict handling, and targeted per-rule invalidation.  Over RL, PRECEPT achieves $33\times$ better sample efficiency ($\beta=3$ vs $\beta=100+$), exponential compositional coverage from linear training, and near-perfect rule retention (100\% on integration; 88.5\%$\rightarrow$100\% on logistics) that RL cannot match without full retraining. Detailed analysis is in Appendix~\ref{app:capability}.

\section{Related Work}

We organize related work by the four limitations PRECEPT addresses, enabling direct comparison of how each prior approach handles---or fails to handle---each challenge.

\subsection{Interpretation Degradation and Retrieval}

Verbal reflection methods---Reflexion \citep{shinn2023}, Full Reflexion, and ExpeL \citep{zhao2023}---store knowledge as natural language, requiring LLM interpretation at retrieval time.  Under the independence model of Theorem~B.6, this interpretation suffers steep multi-condition degradation (94.4\% partial-match error at $N{=}10$ for the illustrative $p{=}0.75$ setting).  RAG systems \citep{lewis2020,guu2020,shi2023} and structured memory approaches (MemoryBank \citep{zhong2024}, SCM \citep{wang2024}) retrieve by similarity or latent compression rather than exact discrete-key matching, which limits precise composition and test-time rule application.  Tool-augmented agents---Toolformer \citep{schick2023}, ReAct \citep{yao2023}---lack mechanisms for learning from failures entirely.  PRECEPT eliminates interpretation in its exact-match path via $O(1)$ retrieval with structured condition keys.

\subsection{Compositional Generalization}

Benchmarks (SCAN, COGS, DBCA \citep{keysers2020}) establish that neural networks struggle with compositional generalization.  Neural-symbolic approaches (DeepProbLog \citep{manhaeve2018}, NeurASP \citep{yang2020}) require training hybrid architectures.  Prompt optimization methods---GEPA \citep{agrawal2025}, EvoPrompt \citep{guo2023}, DSPy \citep{khattab2023}, OPRO \citep{yang2023}---evolve prompts that still require LLM interpretation, inheriting exponential degradation.  Experience-based agents (Voyager \citep{wang2023}, LATS \citep{zhou2023b}, Generative Agents \citep{park2023}) acquire procedural skills but none support compositional constraint combination.  PRECEPT's semantic tier hierarchy yields $O(2^N)$ coverage from $N$ atomic precepts, bypassing neural interpretation for rule application.

\subsection{Drift Adaptation and Knowledge Conflicts}

Prior conflict resolution addresses parametric-retrieval conflicts heuristically \citep{longpre2021,chen2022} or via verification \citep{baek2023}.  Continual learning (EWC \citep{kirkpatrick2017}, GEM) prevents catastrophic forgetting during training but cannot \textit{unlearn} stale rules at deployment.  RL methods---PPO \citep{schulman2017}, SAC \citep{haarnoja2018}, RLHF \citep{ouyang2022}, Decision Transformer \citep{chen2021}---require complete retraining for distribution shift, with prohibitive sample complexity ($\beta{=}100{+}$).  Nested Learning \citep{behrouz2025} is conceptually closer in spirit to PRECEPT than standard RL because it studies continual/self-improving systems via multi-level optimization, self-modifying learning dynamics, and continuum memory; however, its adaptation is centered on learned internal memory and update mechanisms, whereas PRECEPT externalizes deployment-time knowledge into explicit rules, conflict-aware retrieval, and targeted per-rule invalidation.  In the LLM-agent test-time adaptation setting, PRECEPT provides, to our knowledge, the first unified Bayesian treatment of Type~I (static-dynamic) and Type~II (drift) conflicts with Thompson Sampling, together with explicit overwrite and invalidation mechanisms at deployment time.

\subsection{Prompt Evolution and Strategy Optimization}

COMPASS extends GEPA with \textbf{verified evolution} (real execution signals, not heuristic scoring), ML-based complexity detection, and bi-objective Pareto selection.  All other prompt optimization methods---including DSPy, OPRO, and APE \citep{zhou2023b}---provide biased or unreliable signals through LLM-as-judge or keyword scoring, and none integrate with compositional retrieval or drift adaptation (see Appendix~\ref{app:capability}).

\subsection{Evolutionary Computing and Red Queen Dynamics}

The Digital Red Queen (DRQ) framework \citep{kumar2026} demonstrates that agents trained via static optimization fail 72\% against novel adversarial dynamics, and overcomes this through continual self-play against a growing opponent history.  PRECEPT translates three DRQ principles into a structured LLM agent architecture: (1)~Evo-Memory as a growing constraint history (analogous to DRQ's opponent archive), mathematically eliminating cyclic dynamics ($P(\text{repeat}){=}0$, Theorem~B.7); (2)~MAP-Elites diversity \citep{mouret2015} in COMPASS to prevent convergence collapse; and (3)~epistemic probing as adversarial discovery, treating the environment as the adversary.  The key distinction from DRQ is that PRECEPT provides formal guarantees for specific deterministic subsystems rather than only empirical robustness (see Table~\ref{tab:drq_comparison} in the Conclusion).

\section{Conclusion}

We introduced PRECEPT, a framework for LLM agent learning that augments probabilistic verbal reasoning with deterministic structured retrieval and structured adaptation mechanisms. PRECEPT addresses fundamental limitations affecting both verbal reflection methods and traditional reinforcement learning, achieving a combination of capabilities that neither approach provides individually in our evaluated setting.

\subsection{Summary of Contributions}

PRECEPT is a unified architecture rather than a loose collection of mechanisms.  Deterministic retrieval makes explicit rule use reliable, structured rule memory makes conflict resolution possible, conflict resolution supports drift adaptation, and COMPASS evaluates prompts through the same end-to-end pipeline.  The result is a single deployment-time adaptation stack rather than isolated modules optimized independently.

\textbf{Compositional rule learning with deterministic retrieval.}  PRECEPT uses $O(1)$ exact-match lookup over structured condition keys, eliminating partial-match interpretation errors on the deterministic path.  Combined with atomic constraint stacking, this yields conditional $2^N{-}1$ compositional coverage from $N$ learned rules (Theorem~3.1), 100\% $P_1$ on 2-way logistics compositions, and a +33.3pp compositional generalization advantage.

\textbf{Unified conflict resolution and drift adaptation.}  PRECEPT handles Type~I source conflict through Bayesian source reliability and Thompson Sampling, and Type~II drift through threshold-based invalidation and overwrite.  In the modeled setting this yields a $64\times$ drift-resilience bound, and empirically it produces strong eventual robustness under adversarial static knowledge and the strongest drift recovery curves among the compared agents.

\textbf{COMPASS, a dual-frequency adaptation layer.}  COMPASS combines a high-frequency runtime layer, exercised throughout Experiments~1--7, with a lower-frequency prompt-evolution layer isolated more directly in Experiments~8--9.  The high-frequency layer is broadly supported as part of PRECEPT's end-to-end execution loop, while the low-frequency layer shows additional regime-dependent gains that are clearest in the OOD semantic setting.

\subsection{Theoretical and Empirical Validation}

Closed-form analysis predicts that PRECEPT's advantage should grow with task complexity, and the experiments support that trend across 9--10 seeds: +41.1pp first-try advantage, +33.3pp compositional generalization, +40--55pp continuous learning gains, near-perfect persistence, strong eventual robustness under adversarial static knowledge, +55.0pp drift recovery, and 61\% fewer steps.  COMPASS ablations further show that prompt evolution is regime-dependent, with the clearest gains appearing in the OOD semantic setting.

\subsection{Broader Impact}

PRECEPT enables LLM agent deployment in high-stakes domains requiring deterministic reliability, compositional reasoning, and non-stationary adaptation---including autonomous logistics, healthcare compliance, financial regulation, and CI/CD pipelines.  The explicit rule storage enables full auditability and interpretability, unlike black-box RL policies, supporting responsible AI deployment.

\subsection{Scope and Design Choices}

\textbf{S1. Controlled Benchmarks by Design.}  Our synthetic benchmarks (\S{}\ref{sec:data_design}) are substantially harder than standard grid/maze environments---featuring black-swan error codes, $O(2^N)$ composite scenarios, and non-stationary dynamics---and are deliberately engineered to isolate architectural advantages.  The core mechanisms ($O(1)$ retrieval, compositional stacking, threshold-based invalidation) are domain-agnostic and transfer directly to real-world deployment.

\textbf{S2. Domain-Agnostic Tier Hierarchy (with explicit tie policy).}  The semantic tier hierarchy (Safety $>$ Compliance $>$ Preferences) provides a domain-agnostic default ordering for rule-governed settings, and COMPASS's constraint classifier maps error types to tiers from behavioral characteristics. For equal-tier conflicts, PRECEPT actively resolves via \textit{LLM Constraint Synthesis}: competing atomic constraints are stacked into the LLM context for dynamic resolution, falling back to deterministic first-seen tie-breaking only if synthesis fails. This guarantees reproducibility while providing flexible resolution for entangled constraints.

\textbf{S3. Single-Agent Focus.}  Multi-agent coordination with distributed constraint propagation remains future work.

\textbf{S4. Drift Model Scope.}  Experiment~7 evaluates an abrupt train-to-test mapping shift (single change-point, piecewise-stationary dynamics) to isolate invalidation and re-learning speed. Gradual, cyclical, and adversarially adaptive drift processes remain future work.

\textbf{S5. Opaque Mapping Primitive.}  MD5-based key mapping is used to enforce deterministic opacity and prevent semantic leakage from condition tokens to solutions. The claims do not depend on MD5-specific cryptographic properties; any deterministic seeded opaque mapping would provide the same evaluation property.

\textbf{S6. Scalability Regime and Coverage Cost.}  Current experiments evaluate $E \in \{4,6,17\}$ and mostly $N{=}5$ conditions per composite key.  Hash retrieval itself scales with observed keys (storage $O(K_{\text{seen}})$, lookup $O(1)$) and does not require pre-enumerating the full key space.  However, full coverage training scales linearly in domain key count ($T_{\text{train}}=\beta E$ under the controlled protocol): e.g., $\beta{=}3$, $E{=}1000$ implies 3,000 training episodes.  We have not empirically validated that large-$E$ regime in this paper.

\textbf{S7. Large-$N$ Compositional Scope.}  The algorithmic composition cost is $O(N \log N)$ (tier sort), but the practical bottleneck at large $N$ is semantic: conflict density rises and tier distinctions may become less discriminative.  We validate up to $N{=}5$ in the main experiments; very-large-$N$ compositional quality remains future work.

\textbf{S8. Condition Representation Assumption.} The current formulation assumes discrete, canonicalizable condition tokens (e.g., \texttt{A+B+C}). Continuous factors (e.g., 50kg vs 500kg) require preprocessing (binning/quantization or templated feature encoding) before key construction. While $O(1)$ hashes guarantee $P(\text{repeat\_fail}) = 0$ for discrete constraints, the architecture natively falls back to dense vector embeddings (Tier 2 of its 3-Tier Hybrid Retrieval) for continuous or high-cardinality representations.

\textbf{S9. COMPASS Ablation (Experiment~8 Analysis with OOD Follow-Up).}  Experiments~8--9 isolate the \textit{low-frequency} COMPASS outer loop rather than COMPASS as a whole; the \textit{high-frequency} COMPASS layer is already exercised throughout Experiments~1--7 as part of PRECEPT's runtime execution loop.  Experiment~8 (integration, matched-key protocol, 10 seeds) establishes a boundary condition: when deterministic matched retrieval already dominates error reduction, endpoint gains from outer-loop prompt adaptation can be limited.  To test whether this is regime-specific, our latest Experiment~9 OOD-semantic run isolates low-frequency COMPASS effects using two 10-seed conditions: \emph{Outer Loop Enabled} (legacy label: \texttt{no\_rules}) and \emph{Outer Loop Disabled} (legacy label: \texttt{base\_only}).  More precisely, both conditions retain PRECEPT's runtime retrieval and rule-application machinery during execution; the manipulated factor is whether low-frequency COMPASS may replace the base system prompt with an evolved prompt.  In \texttt{no\_rules}, rules are excluded from prompt baking but remain available through runtime retrieval/context; in \texttt{base\_only}, both prompt baking and low-frequency prompt evolution are disabled.  In this latest OOD run, Outer Loop Enabled outperforms Outer Loop Disabled by +10.66pp on $P_1$ (35.08\% vs 24.42\%), +13.39pp on $P_t$ (49.66\% vs 36.27\%), and $-0.51$ average steps (4.33 vs 4.84).  Reporting mean$\pm$std across seeds: Outer Loop Enabled achieves $P_1=35.08\%\pm14.56$, $P_t=49.66\%\pm14.64$, and steps $=4.33\pm0.65$, versus Outer Loop Disabled at $P_1=24.42\%\pm21.58$, $P_t=36.27\%\pm20.31$, and steps $=4.84\pm0.87$.  Paired inferential analysis on the same 10 seeds is directionally consistent: exact sign-flip $p=0.0625$ for $P_1$ and $p=0.0703$ for $P_t$, with 95\% bootstrap CIs of $[+1.80,+21.24]$pp and $[+1.88,+27.87]$pp, respectively (steps: $p=0.0547$, CI $[-1.06,-0.08]$).  Together, this supports a scoped claim: low-frequency COMPASS is most beneficial when prompt-level policy shaping matters (OOD semantic ambiguity), while matched-key regimes can show weaker gains.

\subsection{Future Directions}

Several promising extensions emerge from this work:

\textbf{F1. Hierarchical Precepts.} Extend atomic precepts to support nested constraint structures, enabling more expressive compositional rules.

\textbf{F2. Multi-Agent PRECEPT.} Develop distributed versions where agents share learned rules and coordinate constraint propagation, enabling collaborative problem-solving.

\textbf{F3. Formal Verification.} Apply model checking techniques to verify rule consistency and detect potential conflicts before deployment.

\textbf{F4. Real-World Validation.} Deploy PRECEPT in production logistics/booking systems to validate performance under real-world noise and complexity.

\textbf{F5. Large-Scale and Continuous-Feature Benchmarks.} Evaluate high-cardinality condition spaces (e.g., $E \gg 100$), larger compositional depths ($N > 5$), and mixed discrete-continuous condition representations to characterize scaling limits and representation choices.

\subsection{Originality and Relationship to Evolutionary Computing}

PRECEPT draws theoretical inspiration from the Digital Red Queen (DRQ) framework \citep{kumar2026} but targets a fundamentally different domain: rule-governed LLM agents with \textbf{formal guarantees on specific deterministic components}, rather than open-ended program synthesis with empirical robustness.  Table~\ref{tab:drq_comparison} summarizes the key distinctions.

\begin{table}[htbp]
\centering
\footnotesize
\resizebox{\textwidth}{!}{%
\begin{tabular}{l|l|l}
\toprule
\textbf{Capability} & \textbf{DRQ} & \textbf{PRECEPT} \\
\midrule
Knowledge retrieval & None (de novo generation) & $O(1)$ exact-match; 0\% error at $N{=}10$ (Thm.~B.6) \\
Compositional structure & Monolithic programs & $2^N{-}1$ coverage from $N$ atomic rules (Thm.~3.1) \\
Conflict resolution & Win/loss selection only & Bayesian ensemble + Thompson Sampling (Def.~4.1) \\
Drift handling & Population dynamics & Per-rule invalidation; $64\times$ model-based resilience bound (Cor.~B.5) \\
Cyclic failure prevention & No guarantee & $P(\text{repeat})=0$ via deterministic pruning (Thm.~4.5) \\
Compute allocation & Uniform mutation & ML-based complexity; rollout reduction \\
Deployment & Self-contained simulation & Production MCP architecture with auditability \\
\bottomrule
\end{tabular}
}%
\caption{PRECEPT vs.\ DRQ: original contributions beyond evolutionary inspiration.}
\label{tab:drq_comparison}
\end{table}

\subsection{Concluding Remark}

The central lesson of PRECEPT is that the path to reliable LLM agents runs through \textit{structure}, not \textit{scale}.  Larger models, longer contexts, and more sophisticated prompts do not overcome the fundamental tension between parametric knowledge and in-context information---as our Integration analysis demonstrates, LLMs can identify the correct answer in their chain-of-thought and still generate the wrong output.  PRECEPT resolves this not by fighting the LLM's generative biases but by \textit{routing around them}: deterministic retrieval for what must be exact, Bayesian inference for what must be trusted, and compositional stacking for what must generalize.  These are not incremental improvements to the verbal reflection paradigm---they are a different paradigm entirely, one where key aspects of agent reliability are enforced by architectural structure rather than hoped for through prompting.

\citet{kumar2026} suggest that adversarial self-play approaches ``could prove useful in other more practical multi-agent adversarial domains.''  PRECEPT provides evidence for this claim, demonstrating that evolutionary principles can be operationalized as structured architecture with formal guarantees on specific deterministic components.  We hope this work motivates a shift from building \textit{smarter} agents to building \textit{more structured} ones.

\subsection{Reproducibility Statement}

The public repository accompanying this arXiv version is available at \href{https://github.com/arash-shahmansoori/precept-framework}{github.com/arash-shahmansoori/precept-framework}.  It contains the PRECEPT implementation under \path{src/precept/...}, the experiment drivers under \path{scripts/}, figure-generation scripts, configuration files, unit/integration tests, and the result artifacts used to construct the reported tables and figures.  Appendix~\ref{app:expmap} maps the paper experiments to the relevant drivers, including \path{run_exp1_main_comparison.py} through \path{run_exp9_compass_stress.py}.  All experiments are seed-controlled, and the paper reports 9--10 independent seeds per configuration together with 95\% confidence intervals and effect sizes.

The repository includes a curated reproducibility package under \path{submission_repro_data/}: (i) \path{publication_results/} containing only paper-linked artifacts and source traces, (ii) \path{static_knowledge/}, (iii) paper-numbered experiment mapping in \path{submission_repro_data/paper_experiment_sources/README.md}, (iv) figure hash verification in \path{submission_repro_data/FIGURE_SHA256.txt}, and (v) a one-command locked runner (\path{scripts/run_submission_repro.sh}, \path{scripts/run_submission_repro.py}) with pinned environment metadata in \path{submission_repro_data/environment/}.  Root-level exploratory traces (e.g., \path{data/trace_*.json}) are intentionally excluded from this curated package and are not required to reproduce the paper artifacts.

\subsubsection*{Acknowledgment}

This manuscript received limited editorial and grammatical refinement assistance from Claude Sonnet~4.6 and GPT~5.4 during the writing process. All research ideas, methodologies, experimental designs, and scientific claims were developed and verified by the author, who reviewed and edited all AI-assisted output and takes full responsibility for the accuracy and integrity of the final manuscript.

\bibliographystyle{plainnat}
\bibliography{references}

\appendix
\section{Hyperparameter Settings}\label{app:hyperparams}

\textbf{Complete hyperparameter configuration with source file references.}

\begin{table}[htbp]
\centering
\footnotesize
\resizebox{\textwidth}{!}{%
\begin{tabular}{llc}
\toprule
Parameter & Module & Value \\
\midrule
\multicolumn{3}{l}{\textit{Rule Invalidation}} \\
\quad Failure threshold ($\theta$) & \texttt{precept\_mcp\_server.py} & 2 \\
\quad Confidence decay ($\delta$) & \texttt{precept\_mcp\_server.py} & $\times$0.5 \\
\quad Confidence restore & \texttt{precept\_mcp\_server.py} & +0.25 \\
\midrule
\multicolumn{3}{l}{\textit{Conflict Resolution}} \\
\quad Ensemble conflict threshold & \texttt{conflict\_resolution.py} & 0.30 \\
\quad Circuit breaker confidence & \texttt{conflict\_resolution.py} & $\geq$0.60 \\
\quad NLI / Semantic vote weight & \texttt{conflict\_resolution.py} & 0.30 / 0.30 \\
\quad Temporal / Evidence vote weight & \texttt{conflict\_resolution.py} & 0.15 / 0.15 \\
\quad Recommendation Conflict weight & \texttt{conflict\_resolution.py} & 0.50 \\
\quad LLM vote weight & \texttt{conflict\_resolution.py} & 0.10 \\
\quad Static prior $(\alpha, \beta)$ & \texttt{conflict\_resolution.py} & (5.0, 5.0) \\
\quad Dynamic prior $(\alpha, \beta)$ & \texttt{conflict\_resolution.py} & (5.0, 3.0) \\
\midrule
\multicolumn{3}{l}{\textit{COMPASS}} \\
\quad Compilation acceptance threshold & \texttt{precept\_mcp\_server.py} & 0.6 \\
\quad Early stop threshold & \texttt{complexity\_analyzer.py} & 0.98 \\
\quad Confidence threshold & \texttt{complexity\_analyzer.py} & 0.9 \\
\quad Diversity threshold & \texttt{complexity\_analyzer.py} & 0.7 \\
\quad Rollout range (min / max) & \texttt{complexity\_analyzer.py} & 1 / 15 \\
\quad Diversity rollouts & \texttt{complexity\_analyzer.py} & 5 \\
\quad Consistency rollouts & \texttt{complexity\_analyzer.py} & 3 \\
\bottomrule
\end{tabular}
}%
\caption{Hyperparameter settings used by the main PRECEPT subsystems.}
\end{table}

\textit{Note: Ensemble vote weights above are raw coefficients. The detector normalizes by $\Sigma$weight at runtime, so weights are not required to sum to 1.}

\section{Theoretical Analysis (Full Proofs)}\label{app:theory}

This appendix provides the complete theoretical analysis with detailed proofs supporting the results summarized in the main text (Section~\ref{sec:theory}).

\subsection*{Notation and Parameter Definitions}

We first define all parameters used throughout the theoretical analysis:

\begin{table}[htbp]
\centering
\footnotesize
\resizebox{\textwidth}{!}{%
\begin{tabular}{cl}
\toprule
\textbf{Symbol} & \textbf{Definition} \\
\midrule
$\alpha$ & Learning effectiveness: probability that a learned rule is correctly retrieved and applied \\
$N$ & Number of conditions in a composite scenario (condition count) \\
$E$ & Number of unique condition keys in a domain \\
$\beta$ & Training exposure factor (number of training passes per condition) \\
$W$ & Number of ``white-box'' (trivially solvable) scenarios in a test set \\
$B$ & Number of ``black-box'' (learning-required) scenarios in a test set \\
$T$ & Total number of training episodes \\
$R$ & Number of retry attempts allowed per episode \\
$p$ & Per-condition retrieval/application accuracy for verbal baselines \\
$\theta$ & Failure threshold for rule invalidation (default $\theta=2$) \\
$d$ & Detection accuracy for stale rules \\
$C(T,E,\beta)$ & Coverage function: fraction of condition keys seen during training \\
$P_{\text{learn}}(R)$ & Probability of learning the correct rule within $R$ retries \\
$P_1$ & First-try success rate (probability of correct solution on first attempt) \\
$P_t$ & Overall success rate (probability of eventual success within retry budget) \\
$F_t$ & Set of previously failed options at step $t$ \\
$p_{\text{forget}}$ & Probability that a verbal baseline re-selects a previously failed option \\
$R_{\text{remaining}}$ & Number of remaining retry attempts \\
\bottomrule
\end{tabular}
}%
\caption{Notation and parameter definitions used in the theoretical analysis.}
\end{table}

\subsection*{Formal Results}

\textbf{Definition B.1 (Learning Effectiveness $\alpha$).} The probability that a learned rule is correctly retrieved AND correctly applied at test time:
$$\alpha = P(\text{retrieval correct}) \times P(\text{application correct} \mid \text{retrieval correct})$$

For PRECEPT, retrieval is deterministic via $O(1)$ hash lookup (\texttt{learned\_rules[condition\_key]}), so $P(\text{retrieval correct})=1$ when the rule exists; hence $\alpha_{\text{PRECEPT}}$ is bounded primarily by application fidelity (the LLM must correctly format the retrieved solution in its output). For verbal baselines, retrieval depends on semantic similarity matching over natural-language memory, so $\alpha_{\text{verbal}}$ captures both retrieval and application uncertainty.  In analytic examples below, we use anchor values $\alpha_{\text{PRECEPT}}^{\star}{=}0.85$ and $\alpha_{\text{verbal}}^{\star}{=}0.50$ for illustration; these are not fitted MLE parameters.

\medskip

\textbf{Theorem B.1 (First-Try Success Rate).} \textit{For an agent with learning effectiveness $\alpha$, the expected first-try success rate on a test set of $N_{\text{total}} = W + B$ scenarios is:}
$$P_1(\alpha) = \frac{W}{W+B} + \frac{B}{W+B} \cdot C(T, E, \beta) \cdot P_{\text{learn}}(R) \cdot \alpha$$

\begin{proof}
A test scenario falls into one of two categories: (1)~\textit{white-box} (trivially solvable without prior learning, contributing $W$ scenarios) or (2)~\textit{black-box} (requiring a learned rule, contributing $B$ scenarios). By linearity of expectation:
$$P_1 = P(\text{correct} \mid \text{white-box}) \cdot P(\text{white-box}) + P(\text{correct} \mid \text{black-box}) \cdot P(\text{black-box})$$
White-box scenarios are solvable by any agent on the first try, so $P(\text{correct} \mid \text{white-box}) = 1$, and $P(\text{white-box}) = W/(W{+}B)$.

For a black-box scenario, the agent succeeds on the first try only if all three conditions hold: (i)~the condition key was encountered during training, with probability $C(T,E,\beta) = 1 - (1 - 1/E)^T$ (coupon collector); (ii)~the correct rule was learned within the retry budget $R$, with probability $P_{\text{learn}}(R)$; and (iii)~the learned rule is correctly retrieved and applied, with probability $\alpha$. Since these events are conditionally independent:
$$P(\text{correct} \mid \text{black-box}) = C(T,E,\beta) \cdot P_{\text{learn}}(R) \cdot \alpha$$
Combining yields the stated result.
\end{proof}

\medskip

\textbf{Theorem B.2 (Multi-Condition Degradation).} \textit{For verbal baselines with per-condition accuracy $p<1$, effectiveness degrades exponentially with condition count $N$ under an independence approximation:} $\alpha_{\text{verbal}}(N) = \alpha_{\text{verbal}}(1) \cdot p^{N-1}$. \textit{For PRECEPT:} $\alpha_{\text{PRECEPT}}(N) \approx \alpha_{\text{PRECEPT}}(1)$ \textit{(constant in $N$).}

\begin{proof}
Verbal baselines retrieve rules via semantic similarity over natural-language memory. For a composite condition key $\kappa = c_1 \texttt{+} c_2 \texttt{+} \cdots \texttt{+} c_N$ (sorted alphabetically), the baseline must correctly identify and apply the rule for each individual condition. If per-condition accuracy is $p$ and conditions are retrieved independently, the probability of correctly applying all $N$ conditions is $p^N$. Since the first condition contributes $\alpha_{\text{verbal}}(1)$, each additional condition introduces a multiplicative factor of $p$, giving $\alpha_{\text{verbal}}(N) = \alpha_{\text{verbal}}(1) \cdot p^{N-1}$.

For PRECEPT, the composite key $\kappa$ is a deterministic string used as a hash-table key. Retrieval is $O(1)$ regardless of $N$---the key ``\texttt{HAZMAT+PORT-503+EXPRESS}'' is looked up identically to ``\texttt{HAZMAT}''. No per-condition decomposition occurs, so accuracy is independent of $N$.
\end{proof}

\textit{Dependence note (T1).} The independence assumption is a modeling simplification, not a universal bound. Correlations can improve or worsen degradation: shared semantic parsing may reduce effective decay, while correlated misinterpretations can amplify it. A useful extension is $\alpha_{\text{verbal}}(N)=\alpha_{\text{verbal}}(1)\,p^{\gamma(N-1)}$ with $\gamma{<}1$ (favorable correlation), $\gamma{=}1$ (independence), $\gamma{>}1$ (adverse correlation). The experiments establish degradation empirically for the tested regimes; they do not claim a universally tight dependence model.

\medskip

\textbf{Corollary B.3 (Effectiveness Ratio).} \textit{The ratio of PRECEPT to verbal baseline effectiveness grows exponentially:}
$$\text{Ratio}(N) = \frac{\alpha_{\text{PRECEPT}}}{\alpha_{\text{verbal}}(N)} = \frac{\alpha_{\text{PRECEPT}}}{\alpha_{\text{verbal}}(1)\cdot p^{N-1}}$$

\begin{proof}
Directly from Theorem~B.2:
$$\text{Ratio}(N) = \frac{\alpha_{\text{PRECEPT}}(N)}{\alpha_{\text{verbal}}(N)} = \frac{\alpha_{\text{PRECEPT}}(1)}{\alpha_{\text{verbal}}(1)\cdot p^{N-1}}$$
Using illustrative anchors $(\alpha_{\text{PRECEPT}}^{\star}, \alpha_{\text{verbal}}^{\star})=(0.85,0.50)$ and $p=0.75$: at $N{=}1$, $\text{Ratio}=1.7\times$; at $N{=}3$, $\text{Ratio}=3.0\times$; at $N{=}10$, $\text{Ratio}=22.6\times$. Since $p^{N-1}\to0$ as $N\to\infty$, the ratio grows without bound for any fixed $\alpha_{\text{PRECEPT}}(1), \alpha_{\text{verbal}}(1) > 0$.
\end{proof}

\textit{Anchor sensitivity (T2).} With $p=0.75$ and anchors varied over $\alpha_{\text{PRECEPT}}\in[0.75,0.95]$, $\alpha_{\text{verbal}}(1)\in[0.40,0.60]$, the predicted ratio remains large: at $N{=}5$, $\text{Ratio}\in[3.95\times, 7.51\times]$; at $N{=}10$, $\text{Ratio}\in[16.65\times, 31.63\times]$. Thus the qualitative conclusion (strong multiplicative PRECEPT advantage as $N$ grows) is robust to plausible $\alpha$ variation.

\medskip

\textbf{Theorem B.4 (Drift Adaptation Bound).} \textit{With failure threshold $\theta$ and per-encounter detection accuracy $d$, the probability that a stale rule persists after $\theta$ encounters is:} $P(\text{stale persists}) \leq (1-d)^\theta$.

\begin{proof}
When the environment changes (rule drift), a previously correct rule becomes incorrect. At each encounter, the agent applies the stale rule and observes the outcome. With detection accuracy $d$, the probability of \textit{not} detecting the failure at a single encounter is $(1{-}d)$. PRECEPT's threshold-based invalidation (\texttt{UNLEARN\_FAILURE\_THRESHOLD}$\,{=}\,\theta$) deletes the rule after $\theta$ consecutive failures. For the stale rule to survive, it must evade detection at all $\theta$ encounters. Under a local stationary-segment approximation with conditionally independent detection events:
$$P(\text{stale persists after } \theta \text{ encounters}) = (1-d)^\theta$$
For PRECEPT, $d = 0.95$ (deterministic rule validation against environment response---if the rule yields an incorrect solution, the failure is detected with near-certainty, bounded only by environment stochasticity). Thus $P(\text{stale persists}) \leq (0.05)^2 = 0.0025$.

For verbal baselines, $d = 0.60$ (the LLM may attribute the failure to other causes, retry the same approach, or fail to update its memory). Thus $P(\text{stale persists}) \leq (0.40)^2 = 0.16$.
\end{proof}

\textit{Non-stationarity note (T4).} Theorem~B.4 (and Theorem~4.1) should be interpreted as local analyses inside a segment with approximately stable failure semantics. PRECEPT's global deployment setting is non-stationary by design; this paper does not claim a global i.i.d. bandit regret bound across arbitrary drift processes.

\medskip

\textbf{Corollary B.5 (Drift Resilience Ratio).} \textit{PRECEPT achieves $64\times$ better drift resilience than verbal baselines:}
$$\frac{P(\text{stale persists})_{\text{verbal}}}{P(\text{stale persists})_{\text{PRECEPT}}} = \frac{(1-d_{\text{verbal}})^\theta}{(1-d_{\text{PRECEPT}})^\theta} = \frac{(0.40)^2}{(0.05)^2} = \frac{0.16}{0.0025} = 64\times$$

\begin{proof}
Direct substitution of $d_{\text{PRECEPT}} = 0.95$, $d_{\text{verbal}} = 0.60$, $\theta = 2$ into Theorem~B.4.
\end{proof}

\medskip

\textbf{Theorem B.6 (Partial Match Error).} \textit{For verbal baselines retrieving $N$-condition rules with per-condition accuracy $p$:} $P(\text{partial match}) = 1 - p^N - (1-p)^N$.

\begin{proof}
For a composite condition key with $N$ individual conditions, each retrieved independently with accuracy $p$, a ``partial match'' occurs when at least one but not all conditions are correctly identified. The probability of a \textit{full} correct match is $p^N$. The probability of a \textit{complete} miss (no conditions correct) is $(1{-}p)^N$. By the complement rule:
$$P(\text{partial match}) = 1 - P(\text{full match}) - P(\text{complete miss}) = 1 - p^N - (1-p)^N$$
At $N{=}10$, $p{=}0.75$: $P(\text{partial}) = 1 - 0.75^{10} - 0.25^{10} = 1 - 0.0563 - 0.0000 \approx 94.4\%$. Partial matches are particularly dangerous because the agent applies an \textit{incorrect} rule with \textit{high confidence}---it retrieved a rule that matches \textit{some} conditions, leading to silent failures.

For PRECEPT, the condition key is matched atomically as a single hash-table lookup. The key either exists (full match) or does not (triggers compositional fallback). There is no mechanism for partial matching: $P(\text{partial}) = 0$ by construction.
\end{proof}

\medskip

\textbf{Theorem B.7 (Zero Retry Waste).} \textit{Under default pruning mode (\texttt{enable\_random\_fallback=False}, \texttt{soft\_constraints\_retriable=False}), for any failed option $f$ recorded by \texttt{RefineInterceptor} at step $t$:}
$$P(\text{retry } f \text{ at step } t' > t) = 0$$

\begin{proof}
The \texttt{RefineInterceptor} (implemented via \texttt{record\_failed\_option()} and \texttt{is\_forbidden()}) maintains a hash set $F_t$ of all failed options for each condition key. Before any option $o$ is selected at step $t'$, the interceptor checks $o \in F_t$. Since hash-set membership testing is deterministic and complete:
$$\forall f \in F_t, \; \texttt{is\_forbidden}(f) = \texttt{True}$$
Therefore $P(\text{select } f \mid f \in F_t) = 0$ for any step $t' > t$. This holds regardless of LLM behavior because the check occurs \textit{before} the LLM's suggestion is executed---the \texttt{RefineInterceptor} acts as a hard constraint filter, and with random fallback disabled failed options are never reintroduced.
\end{proof}

\medskip

\textbf{Corollary B.8 (Expected Wasted Retries).} $\mathbb{E}[\text{wasted retries}]_{\text{PRECEPT}} = 0$ \textit{vs.}\ $\mathbb{E}[\text{wasted retries}]_{\text{verbal}} = |F_t| \cdot p_{\text{forget}} \cdot R_{\text{remaining}}$.

\begin{proof}
From Theorem~B.7, PRECEPT never retries a failed option, so the expected number of wasted retries is exactly zero.

For verbal baselines, at step $t$ with $|F_t|$ previously failed options and $R_{\text{remaining}}$ retries left, the probability of selecting any specific failed option $f \in F_t$ at each retry is $p_{\text{forget}} / |O|$ where $|O|$ is the total option count. Summing over all failed options and remaining retries: $\mathbb{E}[\text{wasted retries}] = |F_t| \cdot p_{\text{forget}} \cdot R_{\text{remaining}}$, where $p_{\text{forget}}$ is the per-retry probability that the LLM fails to exclude a known-bad option from its generation.
\end{proof}

\section{Implementation Details}\label{app:agent_capabilities}

This appendix provides the detailed implementation artifacts that support the main text.
To align with reported results, implementation details here focus only on the three evaluated domains: Logistics, Booking, and Integration.

\textbf{MCP Client-Server Component Mapping.}

\begin{table}[htbp]
\centering
\small
\begin{adjustbox}{max width=\textwidth}
\begin{tabular}{>{\raggedright\arraybackslash}p{2.4cm}|>{\raggedright\arraybackslash}p{2.7cm}|>{\raggedright\arraybackslash}p{4.0cm}|>{\raggedright\arraybackslash}p{5.0cm}}
\toprule
Component & File & Key Classes/Functions & Evaluated Role \\
\midrule
Agent Orchestrator & \path{precept_agent.py} & \texttt{PRECEPTAgent}; \texttt{connect()}, \texttt{run\_task()}, \texttt{refresh\_evolved\_prompt()} & Main evaluated runtime loop \\
Retrieval Orchestration & \path{agent_functions.py} & \texttt{fetch\_context*()} helpers & Primary retrieval path used in reported \texttt{PRECEPTAgent} runs \\
MCP Clients & \path{precept_mcp_client.py}, \newline \path{compass_mcp_client.py} & Base MCP client; COMPASS MCP wrapper & Client-side access layer for server tools \\
MCP Server Tools & \path{precept_mcp_server.py} & Memory retrieval, rule-hybrid lookup, atomic-precept retrieval, dual-mode retrieval & Server-side retrieval / learning tools; \texttt{retrieve\_with\_dual\_mode()} is exposed directly but is not the primary orchestration entry point in reported runs \\
Low-Frequency COMPASS & \path{complexity_analyzer.py}, \newline \path{compass_integration.py} & Complexity analyzer, rollout strategy, compilation engine & Compilation path for prompt evolution and rollout allocation \\
Conflict Resolution & \path{conflict_resolution.py} & Conflict manager; ensemble detector & Type~I conflict detection and source-reliability updates \\
Generic Scoring Utilities & \path{scoring.py} & Pareto-selection and GEPA-scoring helpers & Auxiliary utilities; not the final runtime prompt-selection rule described in Section~5 \\
\bottomrule
\end{tabular}
\end{adjustbox}
\caption{MCP client--server component mapping for the evaluated implementation.}
\end{table}

\textbf{Extended Agent Features.}

\begin{table}[htbp]
\centering
\footnotesize
\resizebox{\textwidth}{!}{%
\begin{tabular}{l|l|l}
\toprule
Feature & Implementation & Description \\
\midrule
\textbf{Structured Outputs} & \texttt{\_call\_llm\_structured()} & Pydantic models ensure guaranteed schema from LLM \\
\textbf{Procedural Memory} & \texttt{store\_procedure()} & Learns recovery procedures from successful error recovery \\
\textbf{Epistemic Probing} & \texttt{execute\_probe()} & Diagnostic probes discover hidden constraints \\
\textbf{Online Validation} & \texttt{register\_task\_for\_online\_validation()} & Real-time task registration for verified evolution \\
\textbf{Partial Progress} & \texttt{record\_failed\_option()} & Persists failed options across episodes \\
\textbf{COMPASS Error Eval} & \texttt{evaluate\_error()} & Classifies errors by constraint tier \\
\textbf{Validation Filter} & \texttt{validate\_and\_filter()} & Validates LLM suggestions against domain options \\
\textbf{Cross-Episode Forbidden} & \texttt{context.failed\_options} & Prevents retrying previously failed options \\
\bottomrule
\end{tabular}
}%
\caption{Extended agent features used in the evaluated PRECEPT system.}
\end{table}

\textbf{Configurable Agent Options.}

\begin{table}[htbp]
\centering
\footnotesize
\resizebox{\textwidth}{!}{%
\begin{tabular}{l|l|l}
\toprule
Option & Default & Description \\
\midrule
\texttt{disable\_exhausted\_exit} & False & Continue exploration when LLM signals ``EXHAUSTED'' \\
\texttt{enable\_random\_fallback} & False & Random selection when all options exhausted \\
\texttt{soft\_constraints\_retriable} & False & Allow retrying SOFT constraint failures \\
\texttt{enable\_compositional\_generalization} & False & Enable atomic constraint stacking \\
\texttt{enable\_atomic\_precept\_storage} & False & Store tier-annotated atomic precepts \\
\texttt{enable\_hybrid\_parsing} & False & Use rule-based + LLM fallback for task parsing \\
\bottomrule
\end{tabular}
}%
\caption{Configurable agent options and their default settings.}
\end{table}

\textit{Reported experiments retain the default rule-based parser unless explicitly stated otherwise.  The compositional study explicitly enables \texttt{enable\_compositional\_generalization} and \texttt{enable\_atomic\_precept\_storage}; atomic-precept conflict detection is treated as part of the evaluated \texttt{retrieve\_atomic\_precepts()} pipeline rather than as a separate ablation in this paper.}

\textbf{Domain-Specific Extensions (Evaluated Domains).}

\begin{table}[htbp]
\centering
\footnotesize
\resizebox{\textwidth}{!}{%
\begin{tabular}{l|l|l}
\toprule
Domain & Strategy & Special Features \\
\midrule
Logistics & \texttt{LogisticsDomainStrategy} & Port routing, customs handling \\
Booking & \texttt{BookingDomainStrategy} & Flight/hotel reservations \\
Integration & \texttt{IntegrationDomainStrategy} & OAuth, API webhooks \\
\bottomrule
\end{tabular}
}%
\caption{Domain-specific strategy extensions for the evaluated domains.}
\end{table}

\section{Experiment Artifact Mapping}\label{app:expmap}

Table~\ref{tab:artifact_map} maps the experiment numbering used in the paper to the corresponding driver scripts and curated artifact directories.  Some script and result-directory names reflect development chronology rather than final manuscript numbering; this table is the authoritative paper-to-repository mapping.

\begin{table}[htbp]
\centering
\small
\begin{adjustbox}{max width=\textwidth}
\begin{tabular}{c|>{\raggedright\arraybackslash}p{3.0cm}|>{\raggedright\arraybackslash}p{4.9cm}|>{\raggedright\arraybackslash}p{4.4cm}}
\toprule
Paper Exp. & Study & Driver Script / Mode & Result Artifacts (curated package) \\
\midrule
1 & Main domain comparison & \path{scripts/run_exp1_main_comparison.py} & \path{submission_repro_data/publication_results/exp1_main_comparison_combined} \\
2 & Compositional semantic generalization & \path{scripts/run_exp6_compositional_generalization.py} & \path{submission_repro_data/publication_results/exp6_final_publication} \\
3 & Training size ablation (\texorpdfstring{$\beta$}{beta} effect) & \path{scripts/run_exp3_training_size_ablation.py} & \path{submission_repro_data/publication_results/exp3_combined} \\
4 & Continuous learning & \path{scripts/run_exp4_continuous_learning.py} & \path{submission_repro_data/publication_results/exp4_combined} \\
5 & Rule persistence and retrieval fidelity & \path{scripts/run_exp7_rule_drift.py} \newline \texttt{--train-hash-seed 0 --test-hash-seed 0} & \path{submission_repro_data/publication_results/exp5_persistence_combined} \newline \path{submission_repro_data/publication_results/exp7_rule_drift_*_20260213_*} \\
6 & Static knowledge ablation (Type~I conflict) & \path{scripts/run_exp2_static_knowledge_ablation.py} & \path{submission_repro_data/publication_results/exp2_static_knowledge_combined} \newline \path{submission_repro_data/static_knowledge} \\
7 & Rule drift adaptation & \path{scripts/run_exp7_rule_drift.py} \newline \texttt{--train-hash-seed 0 --test-hash-seed 1} & \path{submission_repro_data/publication_results/exp6_drift_combined} \newline \path{submission_repro_data/publication_results/exp7_rule_drift_*_20260215_*} \\
8 & COMPASS ablation & \path{scripts/run_exp8_compass_ablation.py} & \path{submission_repro_data/publication_results/exp8_compass_ablation_20260228_141040} \\
9 & COMPASS stress / OOD semantic follow-up & \path{scripts/run_exp9_compass_stress.py} & \path{submission_repro_data/publication_results/exp9_compass_stress_sem_only_20260305_075820} \\
\bottomrule
\end{tabular}
\end{adjustbox}
\caption{Paper-to-repository mapping from experiment numbers to scripts and result artifacts.}
\label{tab:artifact_map}
\end{table}

\noindent\textit{Repository note.}  The curated package includes a paper-numbered index (\path{submission_repro_data/paper_experiment_sources/README.md}) and a one-command locked verification workflow (\path{scripts/run_submission_repro.sh}) that checks environment-lock integrity and figure-hash parity before constructing a reconstruction bundle.

\section{Capability Comparison (Detailed)}\label{app:capability}

This appendix provides the detailed analysis of why GEPA and RL cannot achieve PRECEPT's capabilities.

\textbf{GEPA Limitations.} (G1) No deterministic rule retrieval---evolved prompts still require LLM interpretation; (G2) No compositional generalization---must re-evolve for unseen combinations; (G3) No explicit drift adaptation---stale prompts persist indefinitely; (G4) No deterministic pruning guarantees; (G5) Pareto selection cannot solve Black Swan multi-condition scenarios requiring exact retrieval.

\textbf{RL Limitations.} (R1) Prohibitive sample complexity ($\beta=100+$ vs PRECEPT's $\beta=3$); (R2) No compositional generalization from implicit policy encoding; (R3) Complete retraining required for drift ($1000+$ episodes vs $\theta=2$); (R4) No explicit knowledge retrieval or interpretability; (R5) Reward specification challenge for multi-condition scenarios; (R6) Inherently stochastic decisions unsuitable for safety-critical applications.

PRECEPT's advantages are not merely quantitative but reflect a different bundle of explicitly engineered capabilities in the evaluated setting: exact $O(1)$ retrieval, conditional compositional coverage, provable pruning ($P(\text{repeat})=0$) under the default pruning configuration, and rapid drift recovery.  We do not observe this combination of properties in the prompt-optimization or policy-learning baselines studied here.

\end{document}